\documentclass[letterpaper,twocolumn,10pt]{article}
\usepackage{zhanggroup}

\PassOptionsToPackage{hyphens}{url}
\usepackage{xurl}
\usepackage{hyperref}
\usepackage{microtype}
\usepackage{tikz}
\usepackage{listings}
\usepackage{amsfonts}
\usepackage{xspace} 
\usepackage{mathrsfs}
\usepackage{amsmath}
\usepackage{amsthm}
\usepackage{subcaption}
\usepackage{microtype}
\usepackage{booktabs}
\usepackage{ragged2e}
\usepackage{multirow}
\usepackage{graphicx} 
\usepackage{caption,subcaption}
\usepackage{makecell}
\usepackage[table]{xcolor}
\usepackage{bbm}
\usepackage{listings}
\usepackage{xcolor}
\usepackage[linesnumbered,ruled]{algorithm2e}
\hypersetup{
  colorlinks,
  linkcolor={blue!70!green},
  citecolor={green!70!blue},
  urlcolor={orange!70!red}
}
\newcommand{\refappendix}[1]{\hyperref[#1]{Appendix~\ref*{#1}}}

\usepackage{enumitem}
\setlist[itemize]{leftmargin=*}
\DeclareCaptionFormat{listing}{\textbf{#3}}
\newcommand{\mypara}[1]{\noindent{\bf {#1}.} \xspace}
\newcommand{\geocloak}{\texttt{GeoCloak}\xspace}
\newcommand{\editagent}{\texttt{EditAgent}\xspace}
\newcommand{\clueagent}{\texttt{ClueAgent}\xspace}

\date{}

\begin{document}

\title{\Large \bf GeoCloak: An Agentic Framework for Semantics-Preserving Geolocation Privacy via Peripheral Clue Manipulation} 

\author{
Yixin Wu\textsuperscript{1,2}\ \ \
Mingjie Li\textsuperscript{3}\ \ \
Yun Shen\textsuperscript{4}\ \ \
Michael Backes\textsuperscript{3}\ \ \
Yang Zhang\textsuperscript{3}\thanks{Yang Zhang is the corresponding author.}
\\
\\
\textsuperscript{1}\textit{The Hong Kong University of Science and Technology (Guangzhou)} \ \ \ 
\\
\textsuperscript{2}\textit{The Hong Kong University of Science and Technology} \ \ \
\\
\textsuperscript{3}\textit{CISPA Helmholtz Center for Information Security} \ \ \
\textsuperscript{4}\textit{HPE} \ \ \
}

\maketitle

\begin{abstract}
The rapid advancement of large vision language models (LVLMs) has made geolocation inference from images highly accurate and accessible.
This poses serious privacy risks: a passive adversary with black-box LVLM access can infer a user's location from a single photo with no metadata.
Existing protection methods rely almost exclusively on adversarial perturbation, which introduces perceptible visual artifacts and utility degradation.
We identify the \textit{visual-geolocation asymmetry}: geolocation-decisive clues, such as street signs, tend to be visually peripheral but dominate location inference.
Exploiting this asymmetry, we propose \texttt{GeoCloak}, an agentic framework for semantics-preserving geolocation privacy protection.
It acts as a clue-aware planner that analyzes each image to identify geolocation-relevant clues, dispatches targeted semantic edits to directly suppress peripheral clues, and autonomously configures adversarial perturbation to further suppress residual geolocation features, all under iterative utility-aware feedback.
We evaluate \texttt{GeoCloak} on four diverse datasets spanning 108 countries and 576 cities against five LVLMs and three inference methods.
\texttt{GeoCloak} reduces overall city-level accuracy from 41.2\% to 9.9\%, approximately half of the best-performing baseline, while achieving substantially superior utility preservation both qualitatively and quantitatively.
We hope this work takes a meaningful step toward practical, semantics-preserving geolocation privacy protection, giving users greater control over the location information inadvertently embedded in their shared images.
\end{abstract}

\section{Introduction}
\label{section:intro}

The rapid advancement of large vision language models (LVLMs) has made image geolocation inference unprecedentedly accessible and accurate.
Where geolocation once required specialized expertise and hours of manual analysis, LVLMs can infer precise geographic locations from a single image in minutes~\cite{LDDLZSZGL24,MCHDXR24}.
This capability has been further extended through inference-time approaches including prompting strategies~\cite{LYHLDZJ25,SYHTZCFGC25,LZLLZXX25}, retrieval augmentation~\cite{ZZGHLMLM24,JLLZWDHWWY24}, as well as tool-augmented agentic systems~\cite{ZWZLSBZ25,LXWMYL26,WLWHLR25} and post-training with geographic reward signals~\cite{LYJZ24,JPGZL25,XYBKTZXLDLBBPKFMKX25,LZLTW25}.
Consequently, an attacker with access to an LVLM can upload a photo and infer its precise geolocation, posing an unprecedented threat to individual location privacy.

\begin{figure}[!t]
\centering
\includegraphics[width=0.9\linewidth]{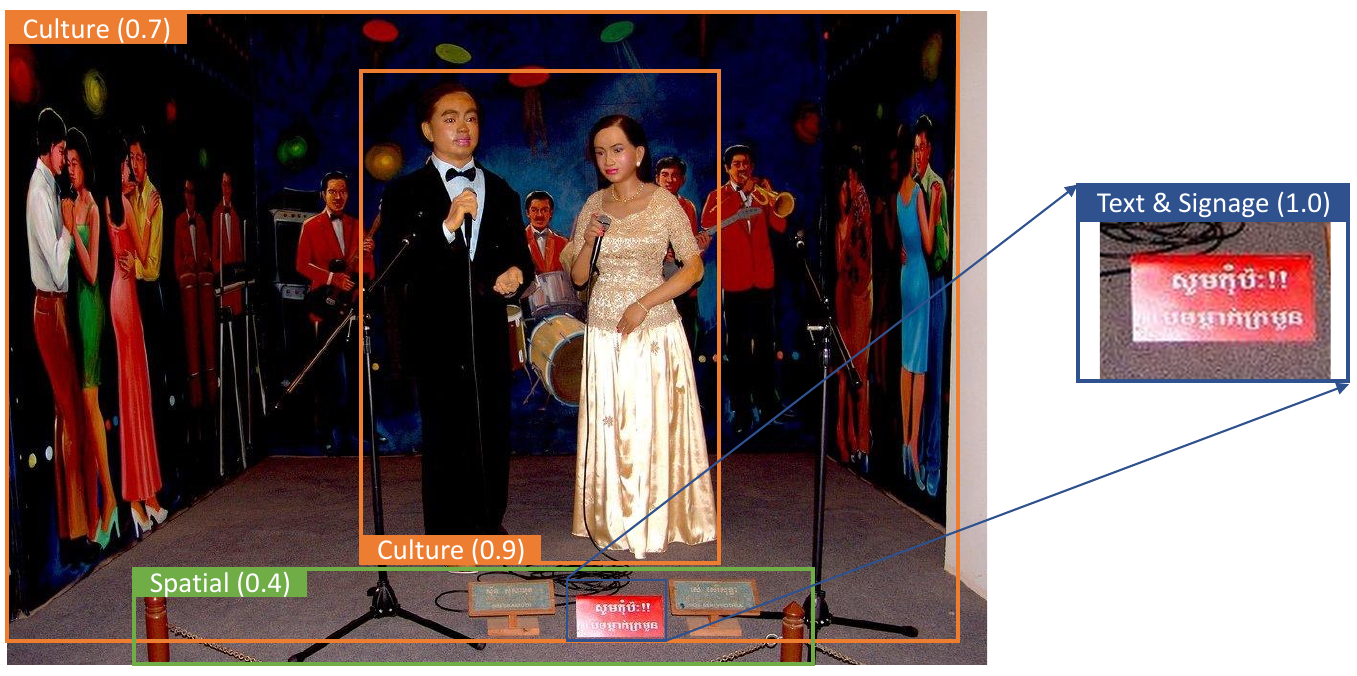}
\caption{An illustration of the visual-geolocation asymmetry.
The most decisive geolocation clue (ranking score 1.0) is a small, visually peripheral sign in Khmer script.}
\label{figure:clue_example}
\end{figure}

To counter this threat, existing work mainly proposes defenses at two levels.
Model-level approaches such as output moderation~\cite{MCHDXR24} and prompt-based refusal~\cite{LZLLZXX25} restrict location disclosure at inference time.
However, these defenses rely on LVLM providers to enforce them, and an attacker can simply turn to an unguarded model to bypass them entirely.
At the image level, the dominant approach is adversarial perturbation.
Before sharing, carefully crafted signals are injected into the image to disrupt LVLM inference.
Existing methods fall into two categories.
Targeted approaches, such as M-Attack~\cite{LZWCS25} and AnyAttack~\cite{ZYMLYCSY25}, steer predictions toward a specific wrong location.
Untargeted approaches, such as GeoShield~\cite{LJXQC26} and ReasonBreak~\cite{ZWCHL25}, concentrate perturbations on geo-revealing regions.
However, these methods apply a fixed perturbation pipeline uniformly to each image without adapting to its content, relying on post-hoc utility evaluation rather than actively incorporating utility constraints into the protection process,  introducing perceptible visual artifacts and degrading image utility (\autoref{section:exp_utility}).

In this paper, we propose \geocloak, an agentic framework for semantics-preserving geolocation privacy protection, motivated by a key observation.
As illustrated in \autoref{figure:clue_example}, a small floor sign in Khmer script definitively places this image in Cambodia, yet it occupies only a tiny fraction of the visual scene.
We term this the \textit{visual-geolocation asymmetry}.
That is, the clues most decisive for LVLM geolocation inference, e.g., street signs, are often visually inconspicuous and easy for humans to overlook.
This asymmetry further characterizes the images that genuinely require protection.
In such images, the location can be inferred from subtle, inconspicuous clues that users never intended to share.

Exploiting this asymmetry, \geocloak introduces semantic editing as a novel protection mechanism that directly targets these peripheral clues, achieving strong privacy protection with minimal impact on visual quality.
Adversarial perturbation is reserved for residual scene-level features.
By integrating adversarial perturbation within an iterative feedback loop, \geocloak actively adjusts hyperparameters and loss configuration based on utility feedback, keeping visual degradation manageable.
Since every image captures a unique scene, the geolocation clues it contains are inherently image-specific, and so are the protection strategies it demands.
A static pipeline is fundamentally ill-suited to this problem.
We address this with an \emph{agentic protection framework}, that acts as a clue-aware planner.
It first \textbf{identifies} the geolocation-relevant elements present in each image.
It then \textbf{dispatches} appropriate manipulation tools: semantic editing, adversarial perturbation, or both.
Finally, it \textbf{iteratively evaluates} the resulting privacy-utility tradeoff via metrics such as CLIP similarity, converging on a protected image that is both geolocationally misleading and semantically faithful.

We evaluate \geocloak on four diverse datasets spanning 108 countries and 576 cities.
Location sharing is not always undesirable~\cite{geo_protect_motivation,open_bench}, and we therefore adopt an evaluation scope, prioritizing protection for images where location can be inferred from subtle, inconspicuous clues that users did not intend to reveal.
\autoref{section:threat_model} and \autoref{section:exp_landmark} details the evaluation scope and exclusion criteria, respectively.
Our main evaluation covers five LVLMs from diverse model families and three distinct inference paradigms, zero-shot prompting, two-stage clue-guided reasoning (GeoMiner~\cite{LZLLZXX25}), and post-trained tool-augmented agentic inference (GeoVista~\cite{WLWHLR25}), yielding 11 attacker configurations.
To assess transfer beyond LVLM-based inference, we further test GeoCLIP~\cite{CNS23}, an embedding-based retrieval method, and GeoSeer~\cite{geoseer}, a commercial black-box tool.
We consider two state-of-the-art image-level defenses, M-Attack~\cite{LZWCS25} (targeted)  and GeoShield~\cite{LJXQC26} (untargeted), as baselines.
\geocloak significantly outperforms both baselines across all settings.
Starting from an initial city-level accuracy of 41.2\%, \geocloak reduces it to 9.9\%, compared to 24.5\% for M-Attack and 19.7\% for GeoShield.
Despite being powered by Gemini-2.5-Pro, \geocloak generalizes beyond its backbone, defending effectively against stronger models.
Against the strongest attacker in our evaluation, Gemini-3.1-Pro under zero-shot inference, \geocloak reduces city-level accuracy from 51.3\% to 13.8\%, while GeoShield and M-Attack only achieve 33.2\% and 37.3\%, respectively.
\geocloak remains effective across zero-shot, clue-guided, and tool-augmented inference.
It also transfers to non-LVLM attackers, including the retrieval-based GeoCLIP and the commercial black-box GeoSeer.
We further conduct a human perception study, in which \geocloak receives an overall quality rating of 4.43/5, substantially higher than M-Attack (2.63) and GeoShield (2.53).
Automatic metrics further confirm its utility preservation.
\geocloak achieves a CLIP similarity of 0.9197, compared with 0.7719 for M-Attack and 0.4209 for GeoShield.
More detailed results are provided in~\autoref{section:exp_utility} and~\refappendix{appendix:human_perception}.
Further analysis confirms that semantic editing alone already achieves strong protection, while combining it with adversarial perturbation further improves results.
\geocloak also generalizes well to both proprietary and open-sourced backbones (\autoref{section:design_choice}).
Targeting the strongest country-level protection, \geocloak requires a median cost of \$0.32 per image (approximately \$0.16 with the Gemini Batch API) and 8.3 minutes, making it practically deployable for real-world geolocation privacy protection (\autoref{section:exp_cost}).
We demonstrate that \geocloak remains effective against adaptive attacks based on conflict-aware prompting, image preprocessing, and fine-tuning on protected images (\autoref{section:adaptive_attacks}).
\noindent Overall, our main contributions are:
\begin{itemize}
\item We identify a \textit{visual-geolocation asymmetry}: geolocation-decisive clues are visually peripheral yet highly informative.

\item We propose \geocloak, an agentic framework that autonomously exploits this asymmetry through targeted semantic editing and adversarial perturbation under iterative utility-aware feedback.
We further release a user-centered interactive version of \geocloak that empowers users to directly control the protection process according to their own privacy preferences.

\item We evaluate \geocloak on four datasets spanning 108 countries and 576 cities across 11 configurations from diverse LVLM families and inference paradigms, along with non-LVLM-based, commercial, and adaptive attackers, demonstrating superior privacy protection and utility preservation.
\end{itemize}

\section{Related Work}
\label{section:related_work}

\mypara{Geolocation Inference}
Recent work has increasingly leveraged LVLMs for geolocation inference, harnessing their world knowledge accumulated during pre-training and, more importantly, their ability to reason over visual clues such as landmarks, signage, vegetation, and architectural styles~\cite{LDDLZSZGL24,ZWZLSBZ25,LYHLDZJ25,LZLLZXX25}.
A line of work exploits prompting strategies, from zero-shot queries~\cite{MCHDXR24,LDDLZSZGL24} to structured chain-of-thought frameworks~\cite{LYHLDZJ25,SYHTZCFGC25,LZLLZXX25}, to elicit hierarchical geographic reasoning from LVLMs.
Beyond prompting, some work augments LVLMs with retrieval-augmented generation to ground predictions with external geo-tagged references~\cite{ZZGHLMLM24,JLLZWDHWWY24}, while others build agentic systems that equip LVLMs with customized tools, such as web search, image zoom-in, and street view navigation, enabling iterative, multi-step evidence gathering~\cite{ZWZLSBZ25,LXWMYL26}.
A further line of work improves geolocation capability through post-training, via supervised fine-tuning or reinforcement learning with geographic reward signals~\cite{LYJZ24,JPGZL25,XYBKTZXLDLBBPKFMKX25,LZLTW25}.
More recently, agentic and post-training approaches have converged: several works train LVLMs end-to-end to act as geolocation agents, instilling both tool-calling capabilities and geographic reasoning through multi-stage post-training pipelines~\cite{WLWHLR25,JDDZHJL26}.
Prior retrieval-based~\cite{VJH17,CNS23} and classification-based methods~\cite{WKP16,HSAF24} have been empirically shown to underperform LVLMs despite requiring millions of labeled training images~\cite{LDDLZSZGL24,MCHDXR24,SYHTZCFGC25}.

\mypara{Protection Methods}
Traditional defenses such as metadata stripping and image blurring have been shown to offer limited effectiveness against LVLMs, which infer location directly from visual content~\cite{LDDLZSZGL24,LZLLZXX25}.
More recently, protection methods targeting LVLM-based geolocation inference have emerged, operating either at the model level~\cite{MCHDXR24,LZLLZXX25} or the image level~\cite{LZWCS25,ZYMLYCSY25}.
At the model level, output moderation~\cite{MCHDXR24} and prompt-based refusal~\cite{LZLLZXX25} restrict location disclosure at inference time.
They require deployment on the service side, leaving individual image owners with no direct control over how their shared images are processed.
At the image level, the dominant approach is adversarial perturbation, adding imperceptible noise to images before sharing to disrupt LVLM geolocation inference.
Existing approaches fall into two categories.
Targeted methods, such as M-Attack~\cite{LZWCS25} and AnyAttack~\cite{ZYMLYCSY25}, steer predictions toward a specific incorrect location by aligning image features with a target image.
Untargeted methods, such as GeoShield~\cite{LJXQC26} and ReasonBreak~\cite{ZWCHL25}, instead aim to suppress geolocation capability without specifying a target.
These approaches, however, rely on post-hoc utility evaluation rather than actively incorporating utility constraints into the protection process.
This inevitably introduces perceptible visual artifacts and utility degradation, as demonstrated in \autoref{figure:qualitative_result}, \autoref{table:utility_numbers}, and \refappendix{appendix:visual_comparison}.

\section{Threat Model}
\label{section:threat_model}

\mypara{Scenario}
Users frequently share images that inadvertently contain visual clues, such as street signs and background architectural details, that are highly informative of geographic location.
As LVLMs become increasingly capable of exploiting such clues, these shared images may unintentionally disclose location information that users never intended to reveal, posing serious privacy risks.

\mypara{Attacker}
The attacker is a malicious individual who seeks to infer the city-, state-, or country-level geolocation of images shared by users on public platforms.
The attacker is assumed to have access to the protected image only, without access to the original unprotected version.
To perform geolocation inference, the attacker leverages state-of-the-art LVLMs and various geolocation inference methods.

\mypara{Defender}
The user is an individual who wishes to share an image on social media or other public platforms, but wants to prevent their location from being inferred.
Before sharing, the user provides the image and its ground-truth geolocation to \geocloak, which analyzes the image and applies targeted modifications.
The ground-truth geolocation is required, as it serves as the termination condition for the iterative protection loop, protection is declared successful only when \geocloak can no longer correctly infer the ground-truth location from the modified image.
When external model APIs are used, the trust assumption extends to the providers that process the images and the ground-truth location.
Alternatively, users or trusted organizations can host both \geocloak and its underlying open-source models within their own infrastructure, keeping these inputs within that environment.
Our open-source configuration demonstrates that effective protection does not require proprietary model APIs (\autoref{table:design_choice}).

\mypara{Protection Goal}
\geocloak aims to achieve two objectives: (1) \emph{privacy goal}: prevent the attacker from correctly inferring the geolocation of a protected image;
(2) \emph{utility goal}: preserve the semantic content and visual quality of the original image so that the user's intended sharing experience is not compromised.

\mypara{Scope}
Several studies~\cite{WLWHLR25,CWWHL25} argue that easily localizable landmark images inflate reported accuracy and fail to reflect the true geolocation reasoning capability of LVLMs, as such locations are often recognized through memorization rather than visual reasoning.
Therefore, these images are excluded during evaluation.
However, we argue that this issue should not be considered solely from an evaluation perspective, but also from the standpoint of user sharing intention.
Location sharing is not always undesirable, e.g., platforms such as OpenBenches~\cite{open_bench} rely on users intentionally sharing photos alongside their geographic locations.
While major platforms and operating systems have begun stripping location metadata by default to protect user privacy, this one-size-fits-all approach has been criticized for overriding the intentional sharing preferences of users who actively want their location to be known~\cite{geo_protect_motivation}.
From both an evaluation and a user intent perspective, we scope our study to exclude images containing globally recognizable landmarks that are visually dominant, as such images are more likely to reflect an intention to share location information rather than concealing it.
For instance, a user who captures a photo with the Eiffel Tower prominently occupying the frame is most likely intending to share that they are in Paris at that moment.
Prior work~\cite{WLWHLR25,CWWHL25} does not specify how to systematically identify and exclude such images; we address this gap by introducing a practical procedure for filtering them (see \autoref{section:exp_landmark}).

\section{\geocloak}
\label{section:system}

\subsection{System Overview}
\label{section:system_overview}

\begin{figure}[!t]
\centering
\includegraphics[width=0.9\linewidth]{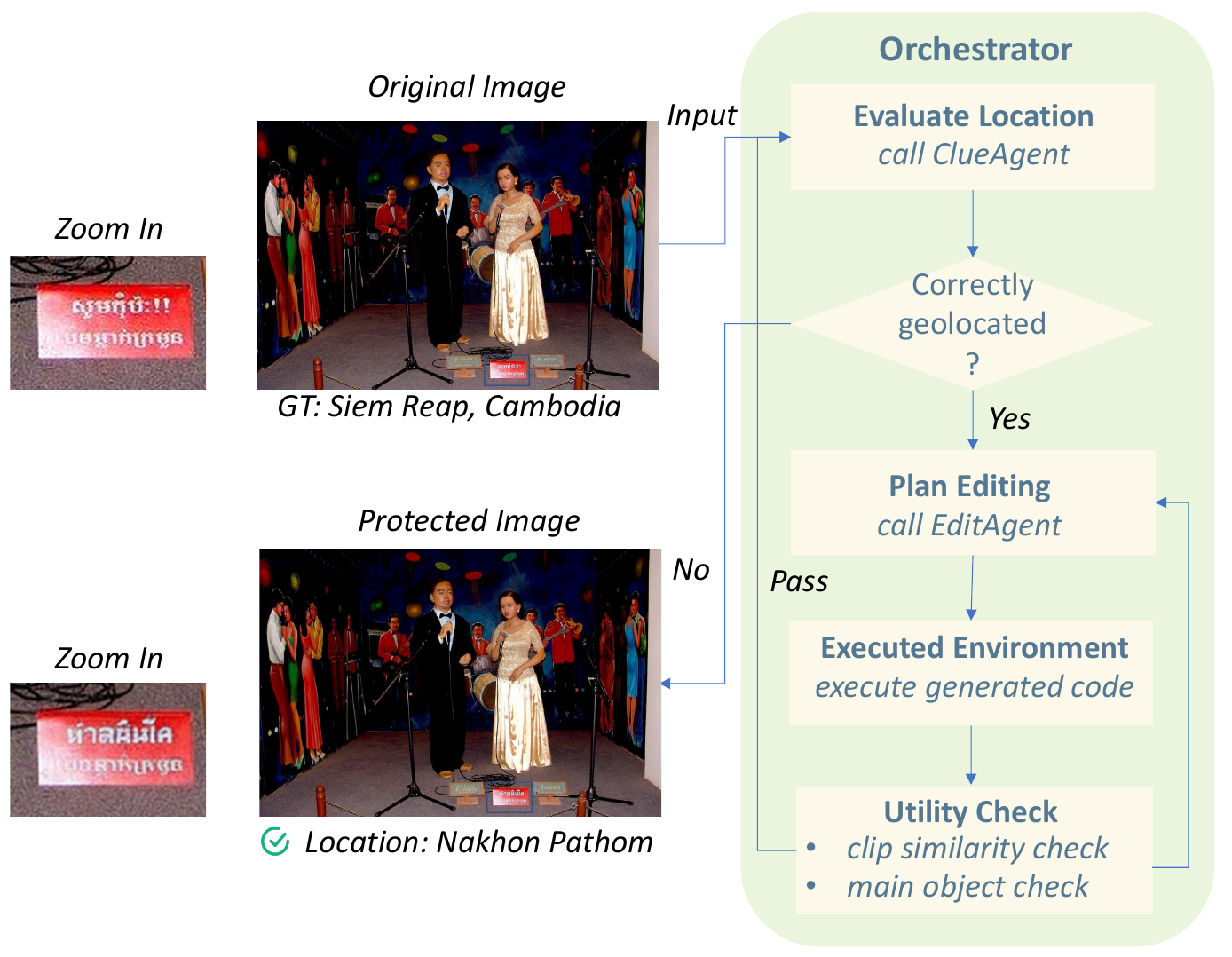}
\caption{
Overview of \geocloak.
A utility check ensures visual quality, and \clueagent re-evaluates the image as a protection check, terminating when the correct location can no longer be inferred.
See multi-turn examples in~\refappendix{appendix:protection_example}.}
\label{figure:basic_framework}
\end{figure}

As illustrated in~\autoref{figure:basic_framework}, \geocloak is an agentic framework comprising two specialized agents, \clueagent and \editagent, coordinated by a central \texttt{Orchestrator}.
Given an input image and its ground-truth geolocation, the system iteratively applies targeted modifications until the image can no longer be correctly geolocated while visual utility is preserved, or until a maximum number of steps is reached.
\clueagent identifies geolocation-revealing clues and predicts the current location, while \editagent formulates a protection plan and executes targeted edits; details are given in \autoref{section:system_clue} and \autoref{section:system_editing}.
The \texttt{Orchestrator} coordinates two checks: a \textit{utility check} to ensure visual quality is preserved, and a \textit{protection check} to determine whether geolocation has been successfully disrupted.
Formally, we define the \textit{utility check} of a protected image $\mathcal{I}^{(t)}$ at iteration $t$ as:
\begin{align}
    \mathcal{U}(\mathcal{I}^{(t)}) &=
    \mathbf{1}\!\left[\mathrm{obj}(\mathcal{I}^{(t)}) = \mathrm{obj}(\mathcal{I}^{(0)})\right] \notag \\
    &\cdot \cos\!\left(\phi^I(\mathcal{I}^{(t)}),\ \phi^I(\mathcal{I}^{(0)})\right),
    \label{eq:utility}
\end{align}
where $\mathrm{obj}(\cdot)$ returns the set of primary objects detected by the LVLM,
$\phi^I(\cdot)$ is the CLIP image encoder, and $\mathcal{I}^{(0)}$ is the original image.
$\mathbf{1}(\cdot)$ acts as a hard constraint: any significant modification that alters the primary objects immediately yields $\mathcal{U} = 0$, regardless of visual similarity.
The cosine term measures residual visual drift.
We further define a multi-granularity \textit{protection check} as:
\begin{equation}
    \mathcal{J}(\hat{\ell},\,\ell^*;\,g) = \textsc{GeoMatch}(\hat{\ell},\,\ell^*;\,g),
    \label{eq:judge}
\end{equation}
where $\textsc{GeoMatch}(\cdot,\cdot;\,g) \in \{0,1\}$ jointly determines whether $\hat{\ell}$ and $\ell^*$ agree at granularity $g$ and $\quad g \in \{\texttt{city},\,\texttt{state},\,\texttt{country}\}$.
The loop terminates at the earliest iteration $t^{*}$:
\begin{equation}
    t^{*} = \min\Bigl\{\,t \;\Big|\;
    \mathcal{J}(\hat{\ell}^{(t)},\ell^*;\,g_{\mathrm{tgt}}) = 0
    \;\wedge\; \mathcal{U}(\mathcal{I}^{(t)}) \geq \tau
    \;\;\text{or}\;\; t = N \,\Bigr\},
    \label{eq:termination}
\end{equation}
where $N$ is the maximum number of allowed iterations, $\tau \in [0,1]$ is the utility threshold, and $g_{\mathrm{tgt}} \in \{\texttt{city},\,\texttt{state},\,\texttt{country}\}$ is the target protection granularity.
When $t^{*} = N$, we report the strongest protection level $g^{*}$ achieved: country-level if $\mathcal{J}(\hat{\ell}^{(N)},\ell^*;\,\texttt{country})=0$, state-level if only $\mathcal{J}(\hat{\ell}^{(N)},\ell^*;\,\texttt{state})=0$, and city-level if only $\mathcal{J}(\hat{\ell}^{(N)},\ell^*;\,\texttt{city})=0$.
If no protection is achieved (i.e., the image remains unmodified), a fallback adversarial perturbation is applied, yielding $\mathcal{I}^{\mathrm{fb}}$, constrained to $\cos\!\left(\phi(\mathcal{I}^{(fb)}),\,\phi(\mathcal{I}^{(0)})\right) \geq \tau$.
At $t=0$, only the \textit{protection check} is performed: \clueagent and the \texttt{Orchestrator} analyze $\mathcal{I}^{(0)}$ to check whether the image is currently geolocatable, i.e., whether $\mathcal{J}(\hat{\ell}^{(0)},\ell^*;\,g_{\mathrm{tgt}}) = 1$.
At each subsequent iteration $t$, both agents and both checks are involved: \editagent formulates a plan and generates executable code, which the \texttt{Orchestrator} executes to yield a candidate $\mathcal{I}^{(t+1)}$.
The \texttt{Orchestrator} first performs the \textit{utility check} (\autoref{eq:utility}): if $\mathcal{U}(\mathcal{I}^{(t+1)}) < \tau$, the modification is rolled back and \editagent revises its plan.
If it passes, \clueagent and \texttt{Orchestrator} re-evaluate the image (\textit{protection check}) via \autoref{eq:judge}: if $\mathcal{J}(\hat{\ell}^{(t+1)},\ell^*;\,g_{\mathrm{tgt}}) = 0$, the loop terminates; otherwise the next iteration begins.
\editagent employs two complementary skills, detailed in \autoref{section:system_editing}.
We further exhibit a successful example of multi-turn \geocloak protection in~\refappendix{appendix:protection_example}.

\subsection{\clueagent}
\label{section:system_clue}

\mypara{Overview}
\clueagent serves a dual role in \geocloak: it acts as both a \textit{geolocation analyzer} and an \textit{attacker}.
At the start of each iteration $t$, \clueagent is prompted to analyze the current image $\mathcal{I}^{(t)}$ and identify all geolocation-revealing clues, producing a structured clue set:
\begin{equation}
    \mathcal{C}^{(t)} = \textsc{Analyze}\!\left(\mathcal{I}^{(t)}\right).
\end{equation}
Each clue $c_i \in \mathcal{C}^{(t)}$ is characterized by a four-tuple:
\begin{equation}
    c_i = \bigl(\tau_i,\;\; d_i,\;\; r_i,\;\; s_i\bigr),
\end{equation}
comprising its \textit{type} $\tau_i$, \textit{description} $d_i$, \textit{region} $r_i \subset \mathbb{R}^4$ (bounding box in pixel coordinates), and \textit{ranking score} $s_i \in [0,1]$.
The type $\tau_i$ has four values: \texttt{landmark}, \texttt{text\&signage}, \texttt{culture}, and \texttt{spatial}, following the categories proposed by Li et al.~\cite{LYHLDZJ25}.
The description $d_i$ captures the semantic content of each clue and the reasoning behind its geolocation relevance, serving as the primary input for \editagent to formulate targeted modification strategies.
Clues are assigned a ranking score $s_i$ reflecting how precisely they localize the scene, from uniquely identifying a specific location down to weak or ambiguous signals, where a higher rank indicates the clue more precisely identifies the location.
In addition to producing $\mathcal{C}^{(t)}$, \clueagent acts as an attacker by predicting a location estimate from the identified clues.
Concretely, \clueagent is an LVLM that is prompted with the full clue set $\mathcal{C}^{(t)}$ and asked to output its best geolocation prediction:
\begin{equation}
    \hat{\ell}^{(t)} = \textsc{GeoPredict}\!\left(\mathcal{C}^{(t)}\right),
    \label{eq:geopredict}
\end{equation}
where $\hat{\ell}^{(t)}$ is a location name (e.g., ``Warsaw, Poland''), which serves as the termination condition for the iteration loop.
That is, protection is declared successful when \clueagent can no longer correctly infer the ground truth location from the modified image, i.e., $\hat{\ell}^{(t)} \neq \ell^{*}$, where $\ell^{*}$ denotes the ground-truth location.
This criterion determines when to stop editing based on \clueagent's prediction and does not imply that all location-revealing clues have been removed.
Protection beyond \clueagent is assessed through the external attacker evaluation described in~\autoref{section:exp_setup}.

\mypara{Adaptive Strategy}
We further consider the case where \clueagent fails to correctly geolocate the image even on the unmodified image, i.e., $\hat{\ell}^{(0)} \not\sim \ell^{*}$, which occurs when geolocation clues are subtle or ambiguous.
In this case, since a stronger attacker in practice may succeed where \clueagent fails, we propose an \textit{adaptive strategy}.
Our system switches to a GT-guided mode in which the ground truth $\ell^{*}$ is explicitly provided as additional context to guide the clue identification process.
Crucially, this mode is determined \emph{once} at $t=0$ based on whether \clueagent succeeds on the original image, and remains fixed for the entire run:
\begin{equation}
    \text{mode} =
    \begin{cases}
        \texttt{standard}  & \text{if } \hat{\ell}^{(0)} \sim \ell^{*}, \\
        \texttt{GT-guided} & \text{if } \hat{\ell}^{(0)} \not\sim \ell^{*}.
    \end{cases}
    \label{eq:mode}
\end{equation}
Under GT-guided mode, each iteration uses an oracle-augmented clue analysis:
\begin{equation}
    \mathcal{C}^{(t)}_{\mathrm{GT}} = \textsc{Analyze}\!\left(\mathcal{I}^{(t)},\; \ell^{*}\right),
\end{equation}
which is then passed to \editagent in place of $\mathcal{C}^{(t)}$.
The ground-truth guidance helps \clueagent identify residual clues that a more capable attacker might exploit, even when its own geolocation inference fails.
The loop stops when \clueagent can no longer identify clues relevant to the ground-truth location, replacing the geolocation-mismatch condition in~\autoref{eq:termination}.
The utility constraint and iteration limit remain unchanged.

\subsection{\editagent}
\label{section:system_editing}

\mypara{Overview}
\editagent aims to formulate targeted modifications to suppress geolocation-revealing clues identified by \clueagent.
Upon receiving the active clue set $\widetilde{\mathcal{C}}^{(t)}$ (equal to $\mathcal{C}^{(t)}$ in standard mode or $\mathcal{C}^{(t)}_\mathrm{GT}$ in GT-guided mode, as determined by~\autoref{eq:mode}), \editagent produces a structured modification plan $\mathcal{P}^{(t)}$, selects and executes appropriate skills, and yields an updated image $\mathcal{I}^{(t+1)}$:
\begin{equation}
    \mathcal{P}^{(t)} = \textsc{Plan}\!\left(\widetilde{\mathcal{C}}^{(t)}\right), \qquad
    \mathcal{I}^{(t+1)} = \textsc{Execute}\!\left(\mathcal{I}^{(t)},\;\mathcal{P}^{(t)}\right).
\end{equation}
\editagent is equipped with two complementary skills: \textit{semantic editing}, which applies targeted modifications to discrete, fine-grained clues; and \textit{adversarial perturbation}, which applies gradient-based noise to disrupt large or structural clues that resist semantic editing.
For each iteration, \editagent selects the appropriate skill based on the remaining clues and generates a structured plan specifying the skill, its corresponding parameters, and a reasoning field that details the rationale behind the chosen strategy.
For \textit{semantic editing}, this includes the editing strategy and a natural language prompt.
For \textit{adversarial perturbation}, this includes configuring the optimization objectives (e.g., loss function) and hyperparameters (e.g., privacy budget).
Each plan is accompanied by executable Python code passed to the \texttt{Orchestrator}.
 
\mypara{Semantic Editing Skill}
This skill targets discrete, localizable clues by applying one of five strategies: \texttt{ReplaceText}, \texttt{AddDecoy},
\texttt{ReplaceObject}, \texttt{EraseObject}, and \texttt{StyleTransfer}.
It is implemented as a prompted LVLM call that receives the clue set $\widetilde{\mathcal{C}}^{(t)}$, the current image $\mathcal{I}^{(t)}$, and the strategy $\Sigma$, and returns a structured plan specifying $\sigma$ and $\rho$.
Due to space limitation, we provide examples of each strategy in~\refappendix{appendix:editing}.
The full prompt template is provided in~\refappendix{appendix:prompt_templates}.
Given the clue with high ranking score $s_i$ relative to their region area $r_i$:

\begin{equation}
    c^{*} = \operatorname*{arg\,max}_{c_i \,\in\, \widetilde{\mathcal{C}}^{(t)}} s_i/|r_i|,
\end{equation}
\editagent selects a strategy $\sigma \in \Sigma$ and constructs a natural-language prompt $\rho$ to drive the edit:
\begin{equation}
    (\sigma,\;\rho) = \textsc{StrategySelect}\!\left(c^{*},\;\mathcal{I}^{(t)}\right),
\end{equation}
where $\Sigma$ denotes the five editing strategies.
\begin{itemize}
\item \texttt{ReplaceText} substitutes geolocation-revealing text, such as street signs, shop names, or license plates, with text from a visually similar but geographically distinct region.
\item \texttt{AddDecoy} is applied when dominant clues are too large or structural to edit directly; it inserts a nearby misleading element to recontextualize existing clues.
\item \texttt{ReplaceObject} swaps geolocation-specific objects, such as post boxes, tram models, or traffic signs, with plausible alternatives from a nearby country.
\item \texttt{EraseObject} removes clues that are too distinctive to replace but small enough to inpaint cleanly.
\item \texttt{StyleTransfer} shifts the visual style of peripheral background elements, such as distant buildings or vegetation, toward a geographically distinct but visually similar region, while leaving the primary subject untouched.
\end{itemize}
Note that we do not distinguish between text and objects for erasure, as both are simply removed and inpainted; the distinction becomes necessary for replacement, where the substitution strategy differs fundamentally between linguistic clues and physical objects.
Strategy selection is implemented via the LVLM following five ordered principles, which together enforce a privacy-utility trade-off: (1)~\textit{start small}, where we prioritize clues by $c^* = \arg\max_{c_i} s_i / |r_i|$, favoring those with a high ranking score $s_i$ relative to their region area $r_i$;
(2)~\textit{blend naturally}, where all modifications specified by the generated prompt $\rho$ must be consistent with the scene's lighting, perspective, and style;
(3)~\textit{stay culturally close}, where replacements must be drawn from a geographically adjacent region to remain visually plausible;
(4)~\textit{preserve the main subject}, where $r^*$ must not overlap with the primary subject region detected by the LVLM;
(5)~\textit{persist and escalate}, where upon failure, retry within $\sigma$ with a revised $\rho$, then switches to an alternative strategy $\sigma' \in \Sigma \setminus \{\sigma\}$.
The region $r^{*}$ of the image is updated as:
\begin{equation}
\mathcal{I}^{(t+1)}\!\left[r^{*}\right] = \textsc{SemanticEdit}\!\left(\mathcal{I}^{(t)}\!\left[r^{*}\right],\;\sigma,\;\rho\right).
\end{equation}

\mypara{Adversarial Perturbation Skill}
This skill applies gradient-based noise to disrupt large or structural clues that resist semantic editing, such as scene-level architectural features.
It requires \editagent to make several autonomous decisions when constructing and refining a perturbation.
For the optimization reference, it selects either \textit{region\_focus}, which builds a bounding-box pool $\mathcal{B} = \{r_i \mid c_i \in \widetilde{\mathcal{C}}^{(t)}\} \cup \{\mathcal{I}^{(t)}\}$ including all clue regions and the full image, or \textit{full\_image}, which uses only the full image as the reference target.
For the \textit{loss function}, it selects $\lambda \in \{0, 1\}$, where $\lambda = 0$ corresponds to \texttt{feature\_only\_loss} and $\lambda = 1$ to \texttt{dual\_loss}.
At each optimization step $k$, the adversarial image $\mathcal{I}^{(t)} + \delta^{(k)}$ is cropped into a local and global view,
\begin{equation}
    \bigl(\mathcal{I}^{\mathrm{loc}}_k,\; \mathcal{I}^{\mathrm{glob}}_k\bigr)
    \;=\; \textsc{GetCrops}\!\left(\mathcal{I}^{(t)} + \delta^{(k)}\right).
    \label{eq:crops}
\end{equation}
A reference crop is sampled uniformly from the bounding box pool:
\begin{equation}
    \tilde{\mathcal{I}}^{(t)}_k \;\sim\; \mathrm{Uniform}(\mathcal{B}).
    \label{eq:ref_sample}
\end{equation}
Let $\phi^I(\cdot)$ denote the image encoder and $\phi^T(\cdot)$ the text encoder of an ensemble of CLIP models.
The per-step loss is accumulated over both adversarial crops and all ensemble members $i \in [M]$:
\begin{equation}
    \begin{aligned}
    \mathcal{L}_{\lambda,k} =
    \frac{1}{M}\sum_{i=1}^{M}
    \sum_{v \,\in\, \{\mathrm{loc},\, \mathrm{glob}\}}
    \Bigl[
        &-\cos \bigl(\phi_i^I(\mathcal{I}^{v}_k),\;\phi_i^I(\tilde{\mathcal{I}}^{(t)}_k)\bigr) \\
        &+ \lambda \cdot \mu \cdot \cos \bigl(\phi_i^I(\mathcal{I}^{v}_k),\;\phi_i^T(\rho_{\mathrm{neg}})\bigr)
    \Bigr],
    \end{aligned}
    \label{eq:loss}
\end{equation}
where $\mu > 0$ is the text-pull weight and $\rho_{\mathrm{neg}}$ is a non-geolocation description generated by \editagent that preserves the scene's visual content while stripping all location-identifying elements.
The perturbation is updated via sign stochastic gradient ascent with step size $\epsilon / 10$ and projected onto the $\ell_\infty$ ball of radius $\epsilon$, where $\epsilon$ denotes the perturbation budget:
\begin{equation}
    \begin{aligned}
    \delta^{(k+1)} &\leftarrow
        \Pi_{\|\cdot\|_\infty \leq \epsilon}\!\left(
            \delta^{(k)} + \tfrac{\epsilon}{10}\,
            \mathrm{sign}\!\left(\nabla_{\delta}\,\mathcal{L}_{\lambda,k}\right)
        \right), \\[4pt]
    \mathcal{I}^{(t+1)} &= \mathcal{I}^{(t)} + \delta^{(K)},
    \end{aligned}
    \label{eq:adv}
\end{equation}
and is run for $K$ steps.
Under $\lambda = 0$ (\texttt{feature\_only}), maximizing $\mathcal{L}_{\lambda,k}$ minimizes cosine similarity between the adversarial crops and the reference, pushing the image away from its  geolocation-revealing CLIP representation.
Under $\lambda = 1$ (\texttt{dual\_loss}), the second term pulls the adversarial crops toward $\phi_T(\rho_{\mathrm{neg}})$, constraining the perturbation direction and reducing visual degradation.

Once \editagent selects adversarial perturbation as the protection skill for the first time, it inspects the available clues and initially selects the optimization reference (\textit{region\_focus} if clues carry bounding boxes, otherwise \textit{full\_image}), loss function (\textit{dual\_loss} when the scene has clear visual content, otherwise \textit{feature\_only}), and sets an initial $K$ and $\epsilon$.
When \textit{dual\_loss} is chosen, \editagent additionally composes a $\rho_{\mathrm{neg}}$, and sets an initial $\mu$.
When a previous attempt has been made, \editagent iteratively adapts its configuration based on the observed privacy-utility tradeoff.
If $\mathcal{U}(\mathcal{I}^{(t)}) < \tau$ (e.g., primary objects have been altered or visual drift is excessive, \editagent rolls back the current perturbation and tightens the configuration: it can switch 
to \textit{dual\_loss}, raise $\mu$, make the description more descriptive, reduce $\epsilon$, or reduce $K$.
If privacy protection is insufficient while $\mathcal{U}(\mathcal{I}^{(t)}) \geq \tau$, it can switch to \textit{region\_focus}, make the description more generic, lower $\mu$, switch to \textit{feature\_only}, increase $\epsilon$, or increase $K$.
The full prompt template is provided in~\refappendix{appendix:prompt_templates}.

\section{Evaluation}
\label{section:evaluation}

\subsection{Experimental Setup}
\label{section:exp_setup}

\mypara{Datasets}
We evaluate \geocloak on four datasets that support level-wise geolocation evaluation to comprehensively assess protection effectiveness across diverse geographic and visual contexts: Im2GPS3k~\cite{VJH17}, GeoBench~\cite{WLWHLR25}, Dataset-GSS~\cite{LYHLDZJ25}, and Dataset-PCW~\cite{LYHLDZJ25}.
We use all available data from GeoBench and Dataset-PCW.
Because Im2GPS3k and Dataset-GSS are substantially larger, we randomly sample 500 images from each dataset.
This yields a benchmark of 1,779 images spanning 108 unique countries and 576 unique cities across all four datasets.
This scale exceeds the 500-image DoxBench evaluation used in the prior GeoMiner study~\cite{LZLLZXX25}.
Across the available images, our full evaluation design comprises over 75,000 evaluation runs, covering 11 attacker configurations and four conditions (Initial, M-Attack, GeoShield, and \geocloak).
Further details on dataset statistics and composition are provided in~\refappendix{appendix:datasets}.

\begin{table}[!t]
\centering
\caption{Proportion of images labeled as \textbf{Exempt} (containing intentional landmarks) across the four evaluation datasets.}
\label{table:exempt_rate}
\scalebox{0.6}{
\begin{tabular}{lrrrr}
\toprule
\textbf{} & \textbf{GeoBench} & \textbf{Im2GPS3k} & \textbf{Dataset-GSS} & \textbf{Dataset-PCW}  \\
\midrule
Exempt Rate & 37.5\% & 53.4\% & 21.0\% & 55.3\%  \\
\bottomrule
\end{tabular}}
\end{table}

\mypara{Agent Configurations}
Both \clueagent and \editagent are powered by large vision-language models.
\clueagent uses Gemini-2.5-Pro~\cite{gemini_pro} for geolocation analysis and clue identification.
\editagent requires two models: one for planning and code generation, and one for image editing.
For the former, we evaluate Gemini-2.5-Pro~\cite{gemini_pro}, o3~\cite{openai_o3}, and Qwen3.6-35B-A3B~\cite{qwen36_35b_a3b}; for the latter, we evaluate Gemini-2.5-Flash-Image~\cite{nano_banana}, i.e., Nano Banana, Seedream-4.5~\cite{seedream}, and Qwen-Image-Edit-2511~\cite{Qwen-Image-Edit-2511}.
The combination of Gemini-2.5-Pro + Gemini-2.5-Flash-Image serves as our primary configuration in the main evaluation, while o3 + Seedream-4.5 is used to verify that \geocloak generalizes across different model choices.
We additionally pair Qwen3.6-35B-A3B with Qwen-Image-Edit-2511 to evaluate a fully open-source configuration.
For all agents, the maximum number of turns is set to 20 and the temperature to 0.5.
For the utility check (\autoref{eq:utility}), we set $\tau = 0.85$, determined empirically to filter out modifications that cause perceptible visual degradation while preserving overall image fidelity.
\textbf{We set the target protection granularity $g_{\mathrm{tgt}} = \texttt{country}$ for the protection check (\autoref{eq:judge}), as country-level mismatch represents the strongest protection level}.

\mypara{Evaluation LVLMs}
We examine against five LVLMs spanning multiple model families and capability levels.
In addition to Gemini-2.5-Pro and o3, which serve as the backbone of \geocloak, we include Gemini-3.1-Pro-Preview (abbreviated as Gemini-3.1-Pro)~\cite{gemini_3_pro}, GPT-5.4~\cite{gpt_5}, and Qwen-3.5-Plus~\cite{qwen_3_plus}.
This selection allows us to assess whether protection optimized against one model generalizes to stronger attackers and to models from entirely different families.
We initially considered Claude-Sonnet-4.6~\cite{claude_sonnet_4} via OpenRouter~\cite{openrouter} as an evaluation LVLM, but excluded it due to a high rate of API errors caused by image size limitations.
In our initial run, 40.0\% of samples in Dataset-PCW exceeded the 5MB input limit, making reliable evaluation infeasible.

\mypara{Geolocation Inference Methods}
Prior retrieval-based methods~\cite{VJH17,CNS23} and classification-based methods~\cite{WKP16,HSAF24} have been empirically shown to underperform LVLMs despite requiring millions of labeled training images~\cite{LDDLZSZGL24,MCHDXR24,SYHTZCFGC25}.
We therefore leverage three geolocation inference methods of increasing sophistication, all built upon the visual reasoning capabilities and broad world knowledge of LVLMs: zero-shot prompting, a two-stage clue-guided framework (GeoMiner~\cite{LZLLZXX25}), and a post-trained tool-augmented agentic system (GeoVista~\cite{WLWHLR25}).
Further details about these inference methods are in~\refappendix{appendix:inference_method}.
Zero-Shot and GeoMiner are instantiated with each of the five evaluation LVLMs described above, while GeoVista is powered by its own trained model~\cite{geovista}.
Overall, combined with multiple LVLMs, these methods yield a total of 11 distinct geolocation attackers.

\begin{table}[!t]
\centering
\caption{Geolocation prediction performance measured by level-wise accuracy (\%) and distance accuracy (\%) on exempt images and non-exempt images.
We leverage zero-shot inference and Gemini-2.5-Pro.}
\label{table:exempt_performance}
\scalebox{0.6}{%
\begin{tabular}{ll rrr rrr}
\toprule
\multicolumn{2}{l}{} & \multicolumn{3}{c}{Level-Wise Acc.} & \multicolumn{3}{c}{Distance-Based Acc.} \\
\cmidrule(lr){3-5}\cmidrule(lr){6-8}
\textbf{Dataset} & \textbf{Type} & \textbf{City} & \textbf{State} & \textbf{Country} & \textbf{@1km} & \textbf{@25km} & \textbf{@100km} \\
\midrule
\textbf{Overall} & Exempt & 73.0\% & 84.0\% & 93.9\% & 30.8\% & 82.3\% & 86.0\% \\
 & Non-Exempt & 48.3\% & 58.9\% & 87.9\% & 16.1\% & 55.2\% & 61.6\% \\
\bottomrule
\end{tabular}%
}
\end{table}

\mypara{Effectiveness Metrics}
A wide variety of metrics have been proposed to evaluate geolocation performance.
They broadly fall into two categories.
The first is hierarchical classification metrics, which assess whether the predicted location matches the ground truth at the city, state, or country level~\cite{ZWZLSBZ25,LXWMYL26,XYBKTZXLDLBBPKFMKX25,CWWHL25,WLWHLR25,LYHLDZJ25}.
The second is distance-based metrics, which measure the geographic error between the predicted and ground-truth coordinates, such as mean or median Haversine distance~\cite{LYHLDZJ25,XYBKTZXLDLBBPKFMKX25}, or the percentage of predictions falling within a given radius threshold (e.g., within 1km accuracy)~\cite{LXWMYL26, GZAS25,WLWHLR25,LZLLZXX25}.
We leverage both types of metrics, with city-level accuracy as the primary metric.
The other metrics are presented in~\refappendix{appendix:other_metrics}.

\mypara{Utility Metrics}
We provide qualitative examples (e.g., \autoref{figure:qualitative_result}) to visually demonstrate the semantic preservation achieved by each defense method.
We quantitatively evaluate image utility from two complementary perspectives.
For visual semantic preservation, we report \textit{CLIP Similarity}, which measures the cosine similarity between the CLIP embeddings of the original and protected images, capturing high-level semantic preservation.
Following GeoShield~\cite{LJXQC26}, we further evaluate text-based semantic preservation by using an LVLM to generate textual descriptions of both images and measuring their similarity via \textit{BERT Score}, \textit{BLEU}, and \textit{ROUGE-L}.

\begin{table*}[t]
\centering
\caption{Geolocation inference performance measured by city-level accuracy (\%) under no defense (Initial) and three defenses (GeoShield~\cite{LJXQC26}, M-Attack~\cite{LZWCS25}, and \geocloak).
\textbf{Bold} indicates the lowest (best) value within each inference method, model, and dataset group.
$\downarrow$ lower is better.
\geocloak uses Gemini-2.5-Pro and Gemini-2.5-Flash-Image.
Results for additional metrics are provided in~\autoref{table:main_eval_full_metric}.}
\label{table:main_evaluation}
\resizebox{\linewidth}{!}{%
\begin{tabular}{ll | rrrrr | r | rrrrr | r}
\toprule
\multicolumn{2}{l}{} & \multicolumn{5}{c}{Zero-Shot} & \multicolumn{1}{c}{GeoVista} & \multicolumn{5}{c}{GeoMiner} & \\
\cmidrule(lr){3-7}\cmidrule(lr){8-8}\cmidrule(lr){9-13}
Dataset & Defense & \textit{Gemini-2.5-Pro} & \textit{Gemini-3.1-Pro} & \textit{GPT-5.4} & \textit{o3} & \textit{Qwen3.5-Plus} & \textit{GeoVista} & \textit{Gemini-2.5-Pro} & \textit{Gemini-3.1-Pro} & \textit{GPT-5.4} & \textit{o3} & \textit{Qwen3.5-Plus} & Overall \\
\midrule
\multirow{4}{*}{GeoBench} & Initial & 74.1\% & 81.9\% & 63.1\% & 71.6\% & 54.7\% & 63.7\% & 70.6\% & 77.2\% & 58.8\% & 62.5\% & 53.1\% & 66.5\% \\
 & M-Attack & 54.4\% & 62.8\% & 37.2\% & 47.2\% & 33.1\% & 42.8\% & 50.9\% & 60.6\% & 38.8\% & 36.9\% & 28.4\% & 44.8\% \\
 & GeoShield & 53.1\% & 64.1\% & 36.2\% & 40.3\% & 29.4\% & 35.9\% & 50.0\% & 60.3\% & 35.0\% & 30.3\% & 28.4\% & 42.1\% \\
 & \geocloak & \textbf{17.5\%} & \textbf{22.2\%} & \textbf{18.1\%} & \textbf{25.3\%} & \textbf{11.2\%} & \textbf{21.6\%} & \textbf{19.1\%} & \textbf{22.5\%} & \textbf{12.2\%} & \textbf{16.9\%} & \textbf{7.5\%} & \textbf{17.6\%} \\
\midrule
\multirow{4}{*}{Im2GPS3k} & Initial & 30.0\% & 30.9\% & 24.0\% & 29.6\% & 16.3\% & 18.9\% & 30.0\% & 26.6\% & 21.5\% & 19.7\% & 17.2\% & 24.1\% \\
 & M-Attack & 18.5\% & 17.6\% & 10.7\% & 11.2\% & 8.2\% & 7.3\% & 15.0\% & 17.2\% & 10.3\% & 5.6\% & 7.7\% & 11.7\% \\
 & GeoShield & 12.9\% & 14.2\% & 7.7\% & \textbf{6.0\%} & 5.6\% & 6.0\% & 12.0\% & 11.2\% & 5.2\% & 3.0\% & 5.2\% & 8.1\% \\
 & \geocloak & \textbf{4.7\%} & \textbf{5.2\%} & \textbf{4.3\%} & \textbf{6.0\%} & \textbf{3.9\%} & \textbf{3.9\%} & \textbf{3.9\%} & \textbf{7.3\%} & \textbf{1.7\%} & \textbf{1.3\%} & \textbf{1.7\%} & \textbf{4.0\%} \\
\midrule
\multirow{4}{*}{Dataset-GSS} & Initial & 41.0\% & 40.8\% & 31.4\% & 37.7\% & 23.5\% & 25.3\% & 41.0\% & 38.2\% & 32.4\% & 29.9\% & 20.8\% & 32.9\% \\
 & M-Attack & 24.8\% & 30.4\% & 15.9\% & 17.0\% & 7.3\% & 15.2\% & 23.8\% & 24.3\% & 15.9\% & 10.9\% & 7.6\% & 17.6\% \\
 & GeoShield & 14.7\% & 22.5\% & \textbf{9.1\%} & \textbf{8.4\%} & \textbf{3.8\%} & \textbf{7.3\%} & 15.7\% & 19.2\% & 10.1\% & \textbf{4.1\%} & 3.5\% & 10.8\% \\
 & \geocloak & \textbf{7.1\%} & \textbf{12.4\%} & 12.9\% & 8.6\% & 5.3\% & 8.9\% & \textbf{7.1\%} & \textbf{9.9\%} & \textbf{7.8\%} & 5.6\% & \textbf{3.0\%} & \textbf{8.1\%} \\
\midrule
\multirow{4}{*}{Dataset-PCW} & Initial & 38.6\% & 43.9\% & 32.5\% & 44.7\% & 21.1\% & 22.8\% & 35.1\% & 51.8\% & 26.3\% & 36.0\% & 20.2\% & 33.9\% \\
 & M-Attack & 24.6\% & 29.8\% & 18.4\% & 21.1\% & 7.0\% & 11.4\% & 21.1\% & 26.3\% & 14.9\% & 17.5\% & 5.3\% & 17.9\% \\
 & GeoShield & 15.8\% & 22.8\% & 10.5\% & \textbf{10.5\%} & \textbf{2.6\%} & 7.0\% & 16.7\% & 20.2\% & 8.8\% & 8.8\% & \textbf{4.4\%} & 11.6\% \\
 & \geocloak & \textbf{5.3\%} & \textbf{13.2\%} & \textbf{9.6\%} & \textbf{10.5\%} & \textbf{2.6\%} & \textbf{4.4\%} & \textbf{1.8\%} & \textbf{11.4\%} & \textbf{3.5\%} & \textbf{3.5\%} & 6.1\% & \textbf{6.5\%} \\
\midrule
\multirow{4}{*}{\textbf{Overall}} & Initial & 48.3\% & 51.3\% & 39.5\% & 46.9\% & 31.1\% & 35.2\% & 46.9\% & 48.9\% & 37.3\% & 38.1\% & 29.7\% & 41.2\% \\
 & M-Attack & 32.3\% & 37.3\% & 21.5\% & 25.2\% & 15.3\% & 21.4\% & 29.8\% & 33.9\% & 21.5\% & 18.3\% & 13.7\% & 24.5\% \\
 & GeoShield & 26.0\% & 33.2\% & 17.1\% & 17.7\% & 11.8\% & 15.6\% & 25.3\% & 29.9\% & 16.4\% & 12.2\% & 11.5\% & 19.7\% \\
 & \geocloak & \textbf{9.5\%} & \textbf{13.8\%} & \textbf{12.2\%} & \textbf{13.3\%} & \textbf{6.5\%} & \textbf{11.1\%} & \textbf{9.4\%} & \textbf{13.3\%} & \textbf{7.3\%} & \textbf{7.8\%} & \textbf{4.4\%} & \textbf{9.9\%} \\
\bottomrule
\end{tabular}%
}
\end{table*}

\mypara{Baseline Protection Methods}
We compare against two representative adversarial perturbation baselines: M-Attack~\cite{LZWCS25} (targeted) and GeoShield~\cite{LJXQC26} (untargeted), covering both categories identified in~\autoref{section:related_work}.
Both are reproduced based on the released code~\cite{geoshield}, following all original hyperparameters.
See details in \autoref{appendix:comparison_methods}.

\subsection{Evaluation Scoping}
\label{section:exp_landmark}

\mypara{Automatic Landmark Intention Analysis}
As discussed in~\autoref{section:threat_model}, we scope \geocloak to images with genuine privacy intent, excluding those that prominently feature globally recognizable landmarks, as users photographing such landmarks are more likely intending to share their location.
To automatically identify and filter out such images, we develop a three-stage landmark intention analysis pipeline.First, we apply $U^2$-Net~\cite{QZHDZJ20} to generate a saliency map and compute a Saliency Dominance Ratio (SDR) for each candidate region, defined as the ratio of the mean saliency within the candidate bounding box to the global image mean saliency.
A high SDR indicates that the region stands out visually from its surroundings.
Second, we employ Florence-2~\cite{XWXDHLZLY24} for phrase grounding, querying it with landmark-related terms (e.g., building, tower, statue, monument) to localize candidate objects and their bounding boxes.
For each candidate, we compute three physical metrics: relative size, centrality, and SDR.
Finally, the top-3 candidates ranked by these metrics are passed to Gemini-2.5-Pro, which classifies each object as \textit{intentional} or \textit{incidental} based on these physical metrics, and further categorizes each object into tier-A (globally recognized landmarks whose location is already public knowledge), tier-B (locally known landmarks), or tier-C (generic structures) based on the world knowledge of the LVLM.
An image is labeled \textbf{exempt} if at least one tier-A object is classified as intentional; otherwise it is labeled \textbf{non-exempt}.
The full prompt is provided in~\refappendix{appendix:prompt_templates}.
\autoref{table:exempt_rate} shows the exempt rate across datasets, ranging from 21.0\% to 55.3\%.
\autoref{table:exempt_performance} further confirms that exempt images are substantially easier to geolocate than non-exempt images across all datasets and metrics; for instance, overall city-level accuracy drops from 73.0\% to 48.3\% when landmark images are excluded, showing that the non-exempt subset presents a more challenging geolocation setting.

\mypara{Human Validation}
To assess the reliability of our automated pipeline, we conduct a human validation study.
Specifically, two independent annotators manually examined exempt samples on GeoBench.
Each sample was evaluated against four criteria: (1) whether the photo intentionally depicts the landmark, (2) whether the landmark is identifiable via Google Search, (3) whether the provided reasoning is supported by verifiable references, and (4) if the reasoning directly identifies the geographic location through the landmark, whether that location matches the ground truth.
The validation yields 86.7\% precision and 93.3\% inter-annotator agreement.
Most misclassified samples coincide with cases of annotator disagreement, highlighting the inherently subjective nature of borderline cases.
We therefore consider 86.7\% precision sufficient to support a reliable large-scale automated assessment.

\mypara{User-Centered System}
We acknowledge that user intent is inherently subjective and cannot be fully captured by automated analysis.
To address this and facilitate real-world deployment, we further release a user-centered interactive \geocloak, letting users directly specify which images they wish to protect.
An example screenshot is provided in \refappendix{appendix:interactive_geocloak}, and the system is publicly accessible.

\begin{table}[!t]
\centering
\caption{City-level accuracy of extra geolocation attackers.}
\label{table:extra_attackers}
\scalebox{0.8}{
\begin{tabular}{llrr}
\toprule
\textbf{Attacker} & \textbf{Mechanism} & \textbf{Initial} & \textbf{\geocloak} \\
\midrule
GeoCLIP~\cite{CNS23} & Embedding-based retrieval & 12.1 & 5.0 \\
GeoSeer~\cite{geoseer} & Commercial black-box tool & 30.7 & 5.7 \\
\bottomrule
\end{tabular}}
\end{table}

\subsection{Effectiveness}
\label{section:exp_main}

We report the geolocation inference performance for the initial setting (without any defense) and after applying each of the three defense methods.
For \geocloak, we use Gemini-2.5-Pro to generate clues, propose editing plans, and produce the corresponding code, while Gemini-2.5-Flash-Image is used to execute the image editing.
All other level-wise metrics and distance accuracy can be found in~\autoref{table:main_eval_full_metric}, from which we draw similar conclusions.

\mypara{Gemini-3.1-Pro is the strongest attacker model}
As shown in \autoref{table:main_evaluation}, it achieves the highest overall city-level accuracy of 51.3\% under zero-shot inference in the initial setting.
From a model perspective, Gemini-3.1-Pro consistently outperforms other models regardless of the inference method used.
From a method perspective, even the zero-shot prompting already achieves strong geolocation performance across all models, highlighting the severity of the privacy threat posed by LVLM without any specialized inference strategy.
We note that o3 exhibits a notable performance gap between zero-shot and GeoMiner, largely due to safety refusals when paired with GeoMiner.
As such refusals are consistently treated as failures across all defense methods, they introduce no systematic bias toward any particular one.
We provide detailed analysis in~\refappendix{appendix:invalid_response}.
GeoVista performs comparatively weaker, which is expected given that it is built on Qwen2.5-VL-7B-Instruct trained via reinforcement learning, whereas the other evaluation models are among the most competitive models from their providers.

\begin{figure}[!t]
\centering
\includegraphics[width=\linewidth]{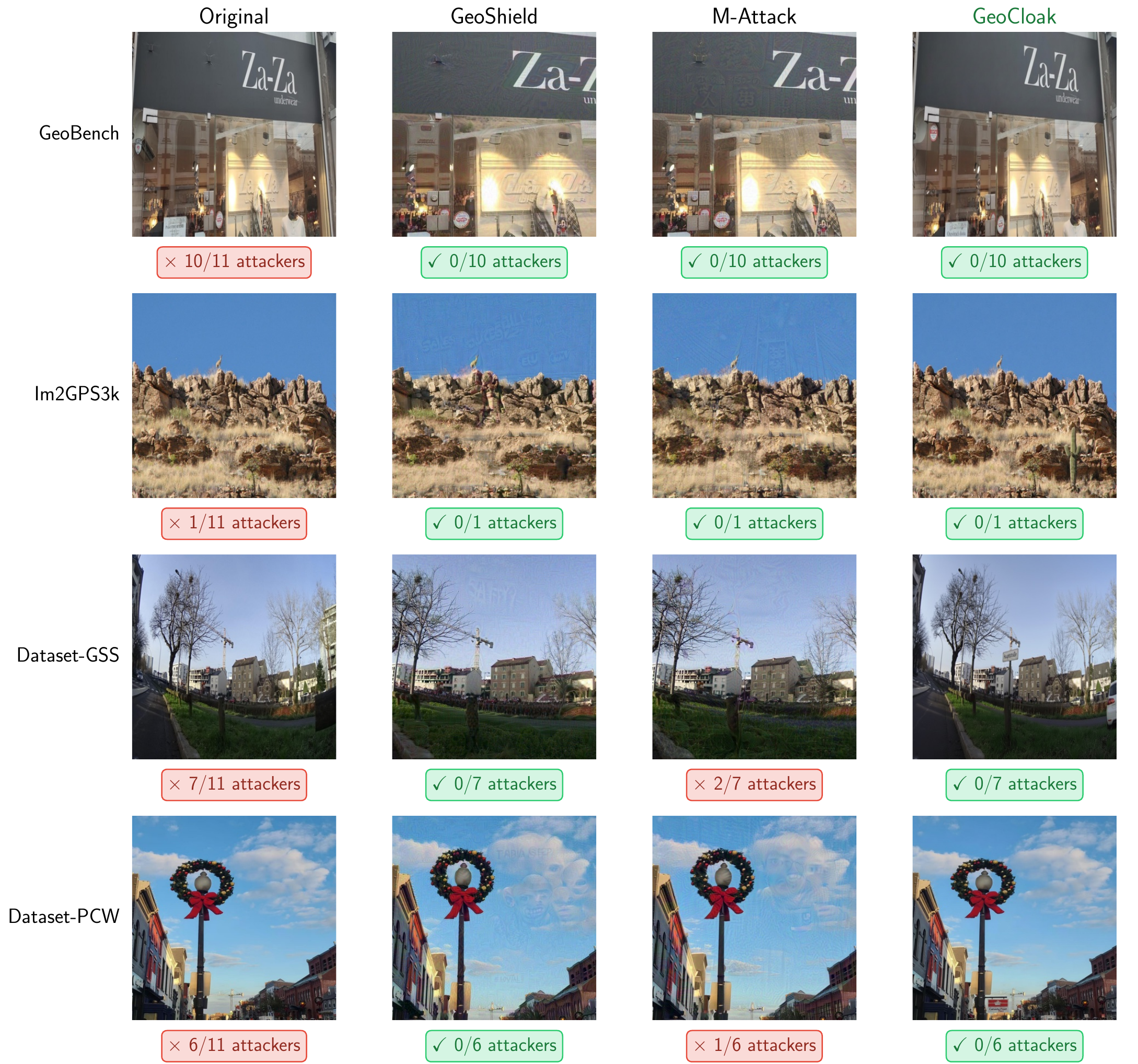}
\caption{
Qualitative comparison of defense methods.
Numbers indicate how many of the 11 attackers, each corresponding to a unique (model, inference method) pair, successfully geolocate the image.
We present cases where all defenses largely succeed.
Best viewed (\refappendix{appendix:visual_comparison}) with zoom to observe severe visual artifacts, such as floating text, superimposed objects, spatial distortions, and unnatural cropping.
}
\label{figure:qualitative_result}
\end{figure}

\mypara{\geocloak achieves the best overall protection across all data-sets and attackers}
Specifically, it reduces city-level accuracy to 9.9\%, approximately half of GeoShield (19.7\%) and substantially below M-Attack (24.5\%).
The impact is especially significant on specific datasets.
For instance, on GeoBench, \geocloak reduces city-level accuracy from 66.5\% to 17.6\%, whereas GeoShield only lowers it to 42.1\%.

\mypara{\geocloak transfers effectively to stronger attackers}
\geocloak powered by Gemini-2.5-Pro reduces the strongest attacker in our evaluation (Gemini-3.1-Pro under zero-shot inference)'s city-level accuracy from 51.3\% to 13.8\%, while GeoShield and M-Attack only achieve 33.2\% and 37.3\%, respectively.
Achieving such strong transferability is notable and indicates that \geocloak can effectively handle even stronger attackers.

\mypara{\geocloak transfers across diverse attacker families and geolocation paradigms}
Across different model families, \geocloak similarly transfers effectively, reducing accuracy to 12.2\% on GPT-5.4, 13.3\% on o3, and 6.5\% on Qwen3.5-Plus under zero-shot inference.
Its protection also transfers across substantially different inference pipelines, from zero-shot prompting and GeoMiner's structured clue-guided reasoning to GeoVista's post-trained, tool-augmented agentic inference.
We further extend the evaluation beyond LVLM-based attackers using GeoCLIP~\cite{CNS23}, an embedding-based retrieval method, and GeoSeer~\cite{geoseer}, a commercial black-box geolocation tool on Im2GPS3k.
As shown in~\autoref{table:extra_attackers}, \geocloak reduces GeoCLIP's city-level accuracy from 12.1\% to 5.0\% and GeoSeer's from 30.7\% to 5.7\%.
Together, these results provide additional evidence that \geocloak transfers across not only LVLM families and inference pipelines, but also substantially different geolocation mechanisms.
We further conduct a fine-grained analysis based on each image's initial geo-identifiability to compare the effectiveness of different protection methods and examine failure cases; see~\refappendix{appendix:geo_identifiability}.

\subsection{Utility}
\label{section:exp_utility}

\begin{table}[!t]
\centering
\caption{Human perception ratings on a five-point Likert scale.
Higher is better ($\uparrow$).}
\label{table:human_perception}
\scalebox{0.6}{
\begin{tabular}{lrrr}
\toprule
Defense & Naturalness$\uparrow$ & Semantic Preservation$\uparrow$ & Overall Quality$\uparrow$ \\
\midrule
GeoShield & 2.23 & 3.68 & 2.53 \\
M-Attack & 2.31 & 3.77 & 2.63 \\
\geocloak & \textbf{4.57} & \textbf{4.45} & \textbf{4.43} \\
\bottomrule
\end{tabular}}
\end{table}

\mypara{Human perception analysis shows that \geocloak preserves visual quality while others introduce severe artifacts}
To assess the perceptibility of visual differences, we evaluate 100 source images sampled uniformly across the four datasets.
Each source image is paired separately with the outputs of \geocloak, GeoShield, and M-Attack, yielding 300 rating tasks.
Two independent annotators rate naturalness, semantic preservation, and overall quality on a five-point Likert scale by comparing each protected image with its original, without knowing the defense method.
The annotators achieve substantial agreement, with a quadratic-weighted Cohen's $\kappa$ of 0.709~\cite{LK77}.
Full experimental details are provided in~\refappendix{appendix:human_perception}.
As shown in~\autoref{table:human_perception}, \geocloak achieves substantially higher overall quality (4.43) than M-Attack (2.63) and GeoShield (2.53), with the same trend in naturalness and semantic preservation.
Consistent with these human ratings, \autoref{figure:qualitative_result} shows that, in cases where all defenses largely achieve successful protection, GeoShield and M-Attack exhibit significant visual degradation, including cropping of non-square images, floating text, and disembodied heads.
See more details in~\refappendix{appendix:visual_comparison}.
These artifacts are clearly perceptible upon close inspection and render the protected images visually implausible.
In contrast, \geocloak achieves successful protection through semantics-preserving targeted edits, such as replacing a text sign, adding contextually plausible objects, or substituting a street sign, all of which blend naturally into the scene while effectively misleading all attackers.

\begin{table}[!t]
\centering
\caption{
Utility comparison of defense methods across four datasets.
Higher is better ($\uparrow$ higher is better).
Bold indicates the best value per dataset.}
\label{table:utility_numbers}
\scalebox{0.65}{
\begin{tabular}{ll | rrrr}
\toprule
Dataset & Defense & CLIP Sim$\uparrow$ & BERT Score$\uparrow$ & BLEU$\uparrow$ & ROUGE-L$\uparrow$ \\
\midrule
\multirow{3}{*}{\textbf{Overall}} & M-Attack & 0.7719 & 0.9124 & 0.0968 & 0.3839 \\
 & GeoShield & 0.4209 & 0.9047 & 0.0766 & 0.3480 \\
 & \geocloak & \textbf{0.9197} & \textbf{0.9158} & \textbf{0.1136} & \textbf{0.4054} \\
\bottomrule
\end{tabular}}
\end{table}

\mypara{Quantitative results further confirm the qualitative observations}
As reported in~\autoref{table:utility_numbers}, \geocloak achieves the highest utility, with an overall CLIP similarity of 0.9197, substantially higher than M-Attack (0.7719) and GeoShield (0.4209).
The particularly low CLIP similarity of GeoShield suggests that its perturbations fundamentally alter the visual semantics of the protected image, consistent with the severe artifacts observed qualitatively.
Across text-based metrics, \geocloak similarly leads on BERT score (0.9158), BLEU (0.1136), and ROUGE-L (0.4054), demonstrating that its targeted edits preserve not only visual appearance but also the semantic content describable from the image.

\mypara{User-Centered Interactive \geocloak}
While our automated utility checks work well in practice, direct user involvement provides the strongest guarantee of utility preservation, as users themselves are the ultimate judges of whether a protected image remains acceptable for sharing.
We therefore release a user-centered interactive version of \geocloak, where users upload the image they wish to protect and specify regions that must remain unmodified.
The system iteratively proposes edits, pausing after each turn for the user to accept, roll back, or stop.
An example screenshot is provided in \refappendix{appendix:interactive_geocloak}, and the system is publicly accessible.

\begin{figure}[!t]
\centering
\includegraphics[width=\linewidth]{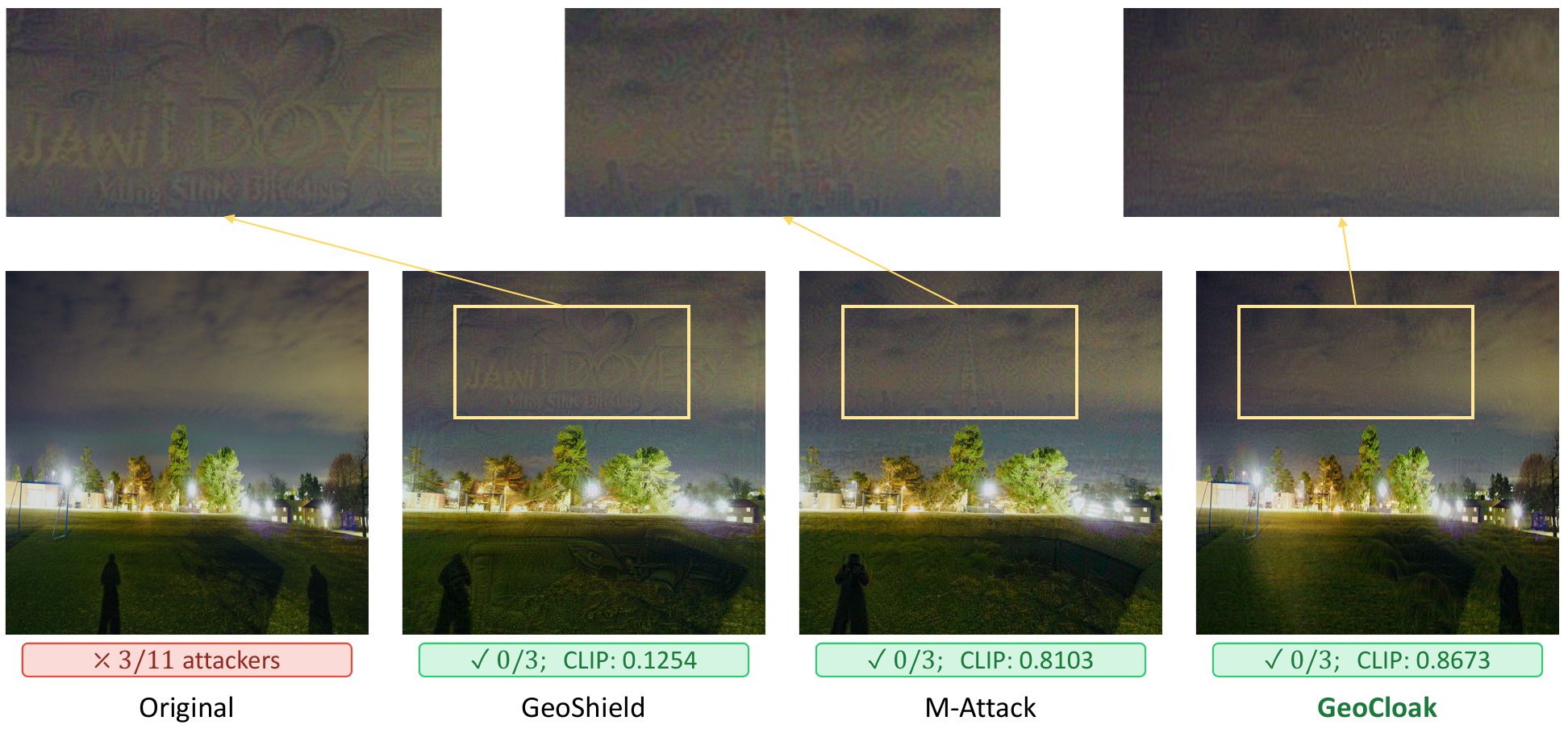}
\caption{Visual quality comparison with zoomed-in regions on a Dataset-PCW sample, where adversarial perturbation is the key skill driving \geocloak's successful protection.
Zoomed-in crops (top) reveal that \geocloak introduces imperceptible perturbations compared to GeoShield and M-Attack.
Comparison on other datasets are in~\refappendix{appendix:visual_comparison}.
}
\label{figure:zoom_in_artifact}
\end{figure}

\begin{figure*}[!t]
\centering
\begin{subfigure}{0.9\columnwidth}
\includegraphics[width=\columnwidth]{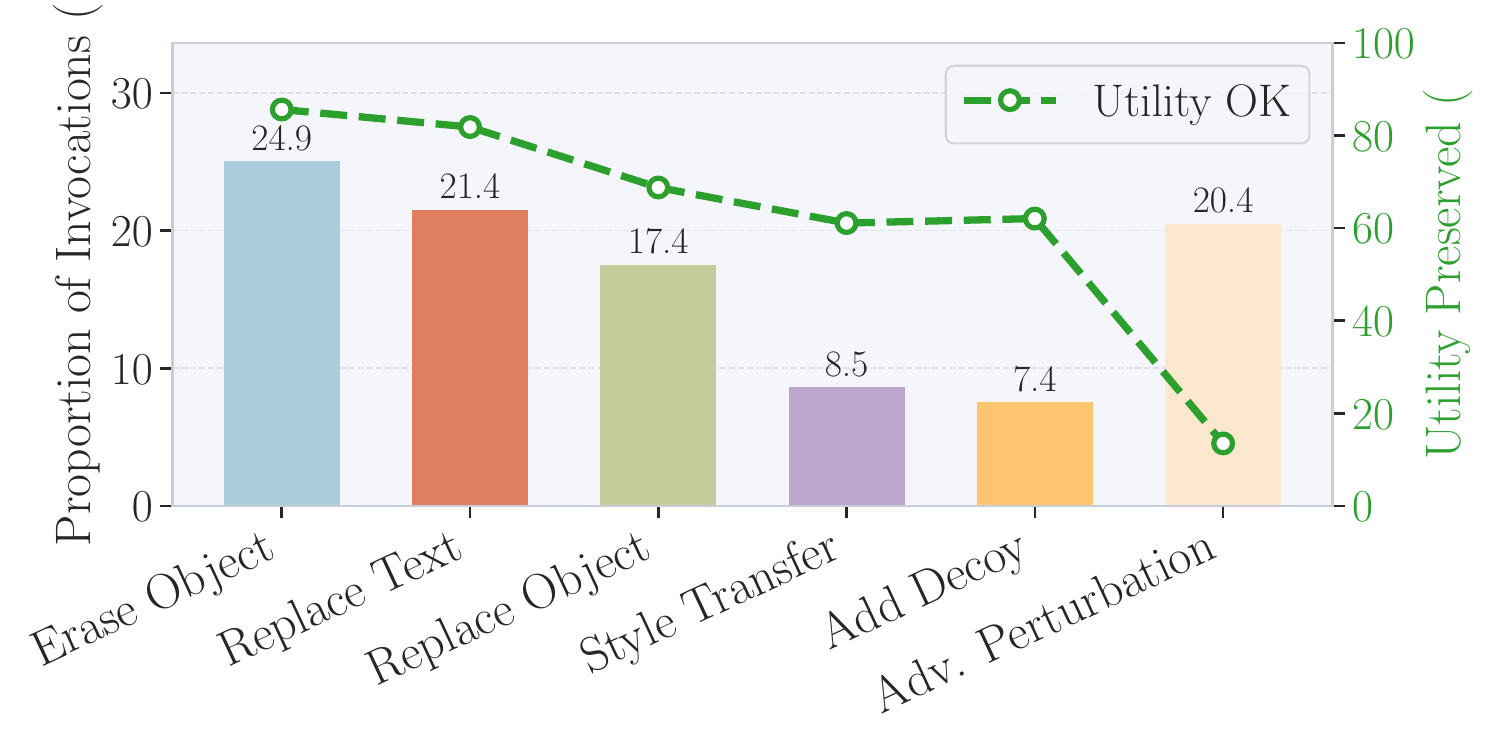}
\caption{Distribution of skill invocations across all turns.}
\label{figure:skill_distribution}
\end{subfigure}
\begin{subfigure}{1.1\columnwidth}
\includegraphics[width=\columnwidth]{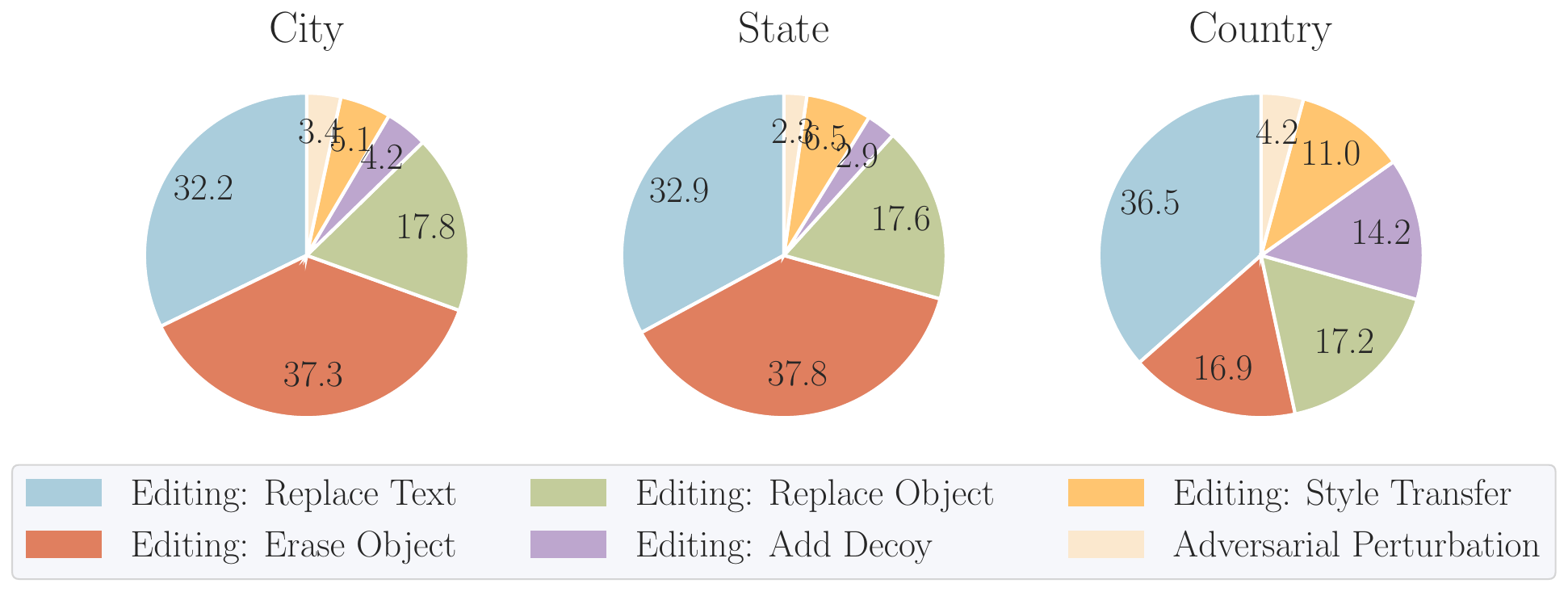}
\caption{Distribution of skills used in the successful protection turn.}
\label{figure:key_strategy}
\end{subfigure}
\caption{Skill usage analysis of \geocloak.
(a) Distribution of all skill invocations and the utility preservation rate for each skill, where Utility OK indicates that the CLIP similarity remains above a pre-defined threshold and the main object is unmodified.
(b) Breakdown of skills used in the first successful turn recorded at each exact protection level (city, state, country); runs that directly reach a stronger level are not back-counted at lower levels.
}
\label{figure:skill_usage}
\end{figure*}

\subsection{Skill Usage}
\label{section:skill_usage}

\mypara{Semantic editing dominates skill usage and yields superior utility preservation}
\autoref{figure:skill_usage} analyzes how \geocloak invokes its skills across all iterations.
As shown in~\autoref{figure:skill_distribution}, Erase Object is the most frequently invoked skill (24.9\%), followed by Replace Text (21.4\%) and Adversarial Perturbation (20.4\%).
This reflects \geocloak's preference for fine-grained, targeted modifications that prioritize high-importance but inconspicuous clues, consistent with the \textit{start small} principle defined in our agent (\autoref{section:system_editing}).
Erase Object also achieves the highest utility preservation rate (85.6\%), followed by Replace Text (81.8\%), as modifying a small, localized element is less likely to disrupt the overall visual semantics of the image.
In contrast, Adversarial Perturbation exhibits the lowest utility preservation rate (13.5\%), as gradient-based noise is inherently less controllable and more prone to causing perceptible visual degradation, a limitation consistent with prior perturbation-based methods such as GeoShield and M-Attack.
Nevertheless, \geocloak's agentic design substantially mitigates this limitation.
\editagent iteratively adjusts hyperparameters, loss composition, and optimization target based on utility feedback, and can further combine perturbation with semantic editing skills to handle discrete clues, allowing perturbation to operate at reduced intensity with less visual degradation.
Across all cases, \geocloak combines semantic editing and adversarial perturbation in 58.4\% of cases and uses semantic editing alone in 39.9\%.
As illustrated in~\autoref{figure:zoom_in_artifact}, the visual artifacts introduced by \geocloak's adversarail perturbation skill are substantially less pronounced compared to those of GeoShield and M-Attack.

\mypara{Removal suffices for city-level clues; country-level clues demand recontextualization}
\autoref{figure:key_strategy} exhibits the skill distribution at the successful protection turn across granularity levels.
We observe that a clear shift emerges from the city to the country level.
At the city level, Erase Object dominates (37.3\%), reflecting that city-level geolocation often hinges on a single highly specific clue, such as a street sign or address number, that can be simply removed.
As granularity coarsens to the country level, Erase Object's contribution drops to 16.9\%.
Instead, Replace Text becomes the dominant successful strategy (36.5\%), while Style Transfer and Add Decoy also contribute by introducing conflicting geographic evidence that redirects the attacker's inference toward a different country.
This shift suggests that country-level geolocation clues are more widely distributed and therefore harder to eliminate, as they often involve broader cultural or architectural patterns that cannot be addressed through simple removal.

\begin{table}[!t]
\centering
\caption{Ablation study on \geocloak design choices.
Overall city-level attack success rate (\%) is averaged across 11 attacker configurations on the Im2GPS3k.}
\label{table:design_choice}
\scalebox{0.7}{
\begin{tabular}{lr}
\toprule
Design Choice & Overall $\downarrow$ \\
\midrule
Initial & 24.1\% \\
\geocloak w/o Semantic Editing & 15.4\% \\
\geocloak w/o Perturbation & 5.1\% \\
Closed-source alternative\textsuperscript{\dag} & 6.3\% \\
Open-source alternative\textsuperscript{\ddag} & 7.4\% \\
\geocloak (Full) & \textbf{4.0\%} \\
\bottomrule
\end{tabular}
}
\parbox{\columnwidth}{\footnotesize \textsuperscript{\dag}o3 + Seedream-4.5. \textsuperscript{\ddag}Qwen3.6-35B-A3B + Qwen-Image-Edit-2511.}
\end{table}

\subsection{Design Choices of \geocloak}
\label{section:design_choice}

\autoref{table:design_choice} compares four variants against the full system on Im2GPS3k: \geocloak w/o Semantic Editing, \geocloak w/o Perturbation, a closed-source alternative that pairs o3 with Seedream~4.5, and an open-source alternative that pairs Qwen3.6-35B-A3B with Qwen-Image-Edit-2511.
The first two variants retain the agentic control loop, including iterative clue analysis and feedback-driven skill configuration. 
They therefore isolate skill availability rather than the contribution of agentic orchestration.

\mypara{Semantic Editing vs.\ Adversarial Perturbation}
When restricted to semantic editing alone, \geocloak w/o Perturbation achieves strong protection with an overall city-level accuracy of 5.1\%, demonstrating that targeted semantic edits are highly effective within the agentic framework.
The full system further reduces this accuracy to 4.0\% by allowing the agent to also select adversarial perturbation.
In contrast, \geocloak w/o Semantic Editing performs considerably worse (15.4\%), which is expected.
Under our strict utility constraint (CLIP similarity>0.85), the adversarial perturbation budget is severely limited, making it difficult to suppress all geolocation clues through noise alone.
However, when combined in the full agent, adversarial perturbation and semantic editing are complementary: semantic edits first remove the most decisive discrete clues, after which a moderate perturbation is sufficient to suppress the remaining scene-level features.
The GeoBench example in~\autoref{figure:qualitative_adv_result} (\refappendix{appendix:visual_comparison}) demonstrates this complementary effect.
\geocloak first erases the semi-transparent text ``NATASHA CADONICI'' from the glass, after which a moderate perturbation successfully defeats all 8 previously successful attacks.

\mypara{Adaptive guidance with a single \clueagent outperforms multi-agent debate on hard cases}
When \clueagent fails to identify the correct geolocation, \editagent cannot effectively protect the image, as it has no reliable clues to target.
This is particularly problematic when \clueagent fails to identify even the correct country, a condition that triggers early termination under the assumption that no protection is needed, while in reality, attackers powered by stronger or different LVLM may still correctly geolocate the image, compromising transferability.
The adaptive mode in \geocloak is designed to address this: when \clueagent fails to identify the correct country, the ground truth is explicitly provided to guide \clueagent in identifying location-relevant clues, until no high-importance clues remain.
Among all non-exempt images in our evaluation, 12.1\% of cases trigger this adaptive mode.
We additionally explore a multi-agent alternative, where an additional o3 instance engages in a one-round debate with Gemini-2.5-Pro to derive a more accurate clue list.
In Im2GPS3k, only 2.9\% of attackers in the adaptive mode still successfully geolocate the protected image, substantially outperforming the multi-agent debate approach (11.8\%).
This result suggests that in these hard cases, where discriminative location clues are inherently difficult to identify from the image alone, providing ground-truth guidance more effectively focuses the search on relevant evidence than augmenting the agent with additional reasoning capacity.

\begin{figure}[!t]
\centering
\begin{subfigure}{0.45\columnwidth}
\includegraphics[width=\columnwidth]{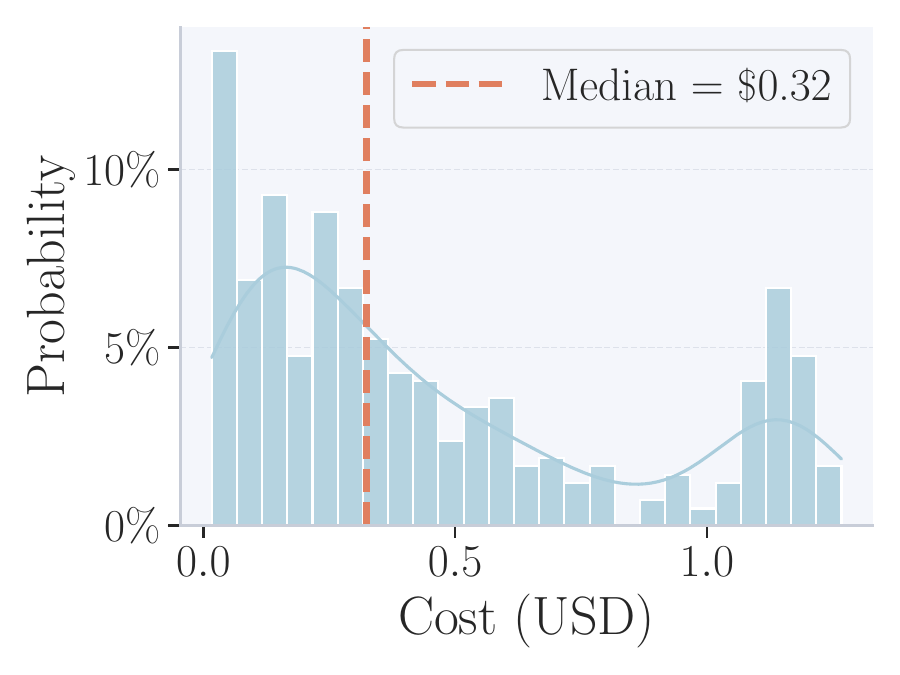}
\caption{Token Cost}
\label{figure:money_cost}
\end{subfigure}
\begin{subfigure}{0.45\columnwidth}
\includegraphics[width=\columnwidth]{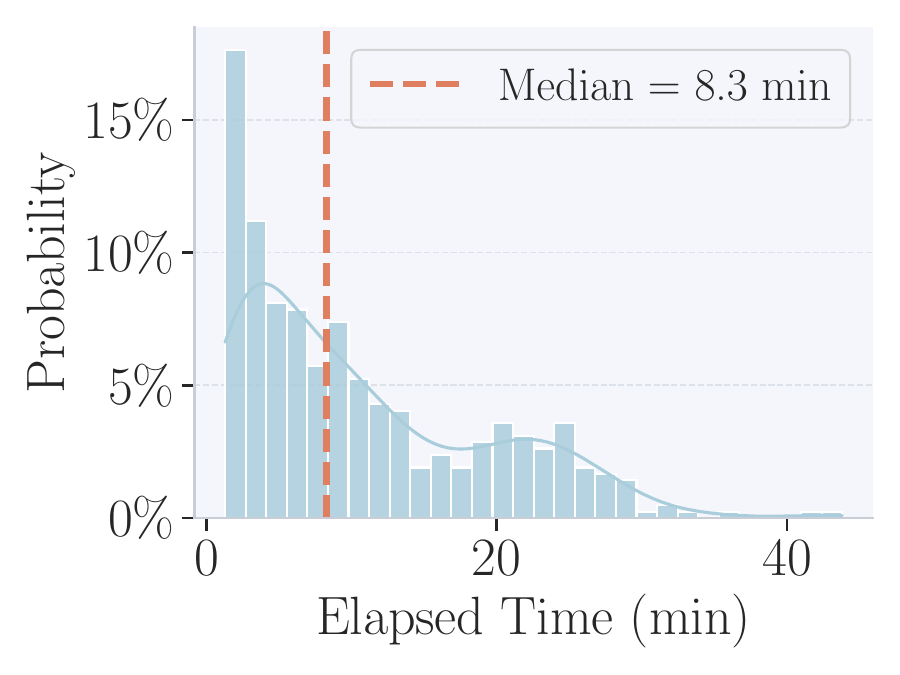}
\caption{Elapsed Time}
\label{figure:time_cost}
\end{subfigure}
\caption{Cost and elapsed time distribution of \geocloak.
API cost is computed from the total token usage of \geocloak based on the official Gemini pricing~\cite{gemini_pricing}.}
\label{figure:cost}
\end{figure}

\mypara{GeoCloak generalizes across proprietary and open-source backbones}
As reported in~\autoref{table:design_choice},
\geocloak (o3+Seedream 4.5) demonstrates that the framework generalizes to different LVLM backbones, reducing overall city-level accuracy from 24.1\% to 6.3\%.
The choice of backbone influences protection strength: \geocloak with Gemini-2.5-Pro+Gemini-2.5-Flash-Image achieves stronger overall protection (4.0\%) than with o3+Seedream 4.5 (6.3\%), though both substantially outperform all baselines (\autoref{table:main_evaluation}).
To understand this gap, we analyze the runtime trajectories of both configurations on Im2GPS3k.
o3 as \clueagent exhibits weaker geolocation clue identification, even in adaptive mode, where the ground-truth location is explicitly provided, it fails to identify highly location-relevant clues in several cases.
This causes the agent to terminate without applying any protection to these images.
Meanwhile, the Gemini backbone employs a combination of semantic editing and adversarial perturbation in 54.5\% of cases, compared to only 34.1\% for o3.
Its typical pattern is to first apply semantic editing, modifying small clues, to introduce semantic-level misdirection, followed by adversarial perturbation to suppress residual scene-level features.
Among cases where both skills are combined, the Gemini backbone achieves transferable protection, where the protected image fools all evaluation attackers, in 66.7\% of cases, compared to only 40.0\% for o3.
This suggests that the Gemini backbone not only uses skill combinations more frequently but also more effectively.
The fully open-source configuration, instantiated with Qwen3.6-35B-A3B and Qwen-Image-Edit-2511, further reduces overall accuracy from 24.1\% to 7.4\%, demonstrating that \geocloak can also operate without proprietary backbones.
Manual inspection reveals no obvious quality differences among images edited by Seedream-4.5, Qwen-Image-Edit-2511, and Gemini-2.5-Flash-Image.

\subsection{Cost and Efficiency Analysis}
\label{section:exp_cost}

\mypara{\geocloak achieves strong protection at acceptable cost}
As shown in~\autoref{figure:cost}, the median cost is \$0.32 per image, and the median elapsed time is 8.3 minutes.
Using the Gemini Batch API would further halve this cost, reducing the estimated median to approximately \$0.16 per image.
Image editing is the largest cost component and is explicitly included in these estimates.
\geocloak typically achieves city- and state-level protection within 2 turns (median = 2.0 for both), as the geolocation-decisive clues at these granularities tend to be more peripheral and discrete, such as a specific street sign or address number, making them easier to identify and suppress.
Country-level protection requires considerably more iterations (median = 5.0), consistent with our earlier finding that country-level clues are more distributed and require repeated recontextualization.
In practice, users can run \geocloak in advance to protect images they plan to share.

\subsection{Adaptive Attacks}
\label{section:adaptive_attacks}

\mypara{\geocloak has high robustness against diverse attackers}
Our agent has already been evaluated against a diverse set of attackers, ranging from simple zero-shot prompting to structured two-stage cot prompting (GeoMiner) and tool-augmented agentic inference (GeoVista), embedding-based retrieval method (GeoCLIP) and commercial tool (GeoSeer) and consistently achieves the strongest protection.

\mypara{Adaptive prompting provides limited gain at the cost of unprotected image performance}
We further evaluate \geocloak against an adaptive attacker that is explicitly aware that the image may have been modified.
Rather than using a standard zero-shot prompting, this attacker instructs the LVLM to identify and discard potentially manipulated or conflicting clues before inferring the location.
As shown in~\autoref{table:adaptive_prompting}, the adaptive attacker achieves a higher success rate than naive prompting after \geocloak defense (15.9\% vs.\ 11.1\% overall).
This indicates that conflict-aware prompting provides some additional attack capability.
Nevertheless, \geocloak reduces the adaptive attacker's performance from 41.6\% to 15.9\%, a substantial reduction that demonstrates the robustness of \geocloak's protection even against an informed attacker.
Notably, the adaptive approach provides little benefit on unprotected images and even slightly degrades the performance (41.6\% vs.\ 43.4\%), suggesting that indiscriminately treating clues as suspicious can mislead the attacker into discarding genuine evidence.
This means an attacker cannot blindly apply adaptive prompting without prior knowledge of whether the image has been protected, further limiting its practical threat.

\mypara{Adaptive fine-tuning only partially recovers geolocation performance}
We further consider an adaptive attacker with access to \geocloak-protected images and their location labels, who attempts to adapt geolocation models directly to the protected-image distribution.
We fine-tune both GeoCLIP, a retrieval-based geolocation model, and Qwen3.5-9B, a reasoning-based model adapted using LoRA, with an 80/20 train/test split of protected images.
Fine-tuning provides only partial recovery for both attacker families.
For vanilla GeoCLIP, applying \geocloak reduces city-level accuracy from 17.0\% on unprotected images to 7.4\% on protected images; fine-tuning on protected images raises it to only 9.6\%.
Similarly, vanilla Qwen3.5-9B~\cite{qwen35} achieves 24.5\% on unprotected images and 6.4\% on protected images, while LoRA fine-tuning recovers accuracy to 11.3\%, still less than half of its unprotected performance.
Thus, although adaptive fine-tuning weakens \geocloak to some extent, it does not restore the original geolocation capability, demonstrating that \geocloak remains effective against adaptive fine-tuning attacks across both retrieval-based and reasoning-based models.

\mypara{\geocloak remains robust against denoising-based adaptive attacks}
Because \geocloak uses adversarial perturbation as one of its protection skills, an adaptive attacker can apply image purification before inference to remove the perturbation.
Here we evaluate DiffPure~\cite{NGHXVA22}, a representative diffusion-based purification method.
However, it substantially degrades image utility, fine-grained details (e.g., text and human subjects) become blurred or unrecognizable, while perturbation artifacts remain visible.
This behavior occurs across its guided diffusion~\cite{DN21}, DDPM~\cite{HJA20}, and Stable Diffusion-based~\cite{RBLEO22} variants.
We also evaluate JPEG compression as a lightweight alternative with negligible utility loss.
As shown in~\autoref{table:adaptive_prompting}, compression provides little benefit, indicating that \geocloak remains robust to this adaptive attack.

\section{Discussion, Limitation, and Future Work}
\label{section:discussion}

\mypara{The Necessity of Our Agent Architecture}
\geocloak cannot be trivially reduced to either a fixed pipeline or a trivial agent composition.
Geolocation clues are image-specific and only discoverable through inference-time reasoning.
This makes static strategies insufficient to jointly preserve privacy and utility.
Moreover, \geocloak's effectiveness relies on adaptive reasoning, including adversarially grounded stopping criteria, dynamic strategy selection, adaptive optimization, and iterative feedback between clue discovery and editing.
For example, in~\autoref{figure:iterate_example}, two attempts to modify the shop sign fail the utility check.
Based on this feedback, \editagent dynamically adjusts its strategy by rolling back the failed edits and replacing the Moroccan-style sofa instead, shifting \clueagent's prediction from Morocco to Algeria.
Such failures are difficult to address with a fixed editing pipeline, where subsequent edits are predetermined rather than adjusted to intermediate feedback.

\begin{table}[!t]
\centering
\caption{City-level accuracy (\%) over five LVLMs under naive and adaptive attackers with zero-shot prompting, before and after \geocloak defense, and (for the naive attacker) after additional JPEG compression.}
\label{table:adaptive_prompting}
\scalebox{0.5}{
\begin{tabular}{llrrrrr}
\toprule
\textbf{Attacker} & \textbf{Setting} & \textbf{GeoBench} & \textbf{Im2GPS3k} & \textbf{Dataset-GSS} & \textbf{Dataset-PCW} & \textbf{Overall} \\
\midrule
  \multirow{3}{*}{Naive} & Initial & 69.1 & 26.2 & 34.9 & 36.1 & 43.4 \\
  & \geocloak & 18.9 & 4.8 & 9.3 & 8.2 & 11.1 \\
    & \geocloak (JPEG) & 20.0 & 4.9 & 9.6 & 8.2 & 11.5 \\
\midrule
  \multirow{2}{*}{Adaptive} & Initial & 66.6 & 26.5 & 32.8 & 33.0 & 41.6 \\
   & \geocloak & 24.0 & 8.0 & 15.2 & 11.9 & 15.9 \\
\bottomrule
\end{tabular}}
\end{table}

\mypara{Prompt Sensitivity}
Prompts templates used in \geocloak are provided in \refappendix{appendix:prompt_templates}.
We find that paraphrasing these prompts, without altering each agent (\autoref{section:system}), has negligible impact on protection performance.
This indicates that \geocloak is robust to prompt variations and does not require careful prompt engineering to achieve strong results.
While frameworks such as DSPy~\cite{KSMZSVHSJMMZP23} enable systematic prompt optimization for single-turn pipelines, they are not directly applicable to our multi-turn agentic setting, where prompts interact dynamically across agents and iterative feedback loops.
We therefore leave systematic prompt optimization for multi-turn agentic systems to future work.

\mypara{User-Centered Agentic System}
We argue that geolocation privacy protection should respect user intent, protecting only images whose owners do not intend to disclose their location.
Because ground-truth intent is unavailable in existing datasets, our evaluation (\autoref{section:exp_landmark}) approximates it using visual landmark analysis, while automated utility checks estimate acceptable modifications.
To move beyond these proxies, we introduce an interactive version of \geocloak that allows users to specify objects that must remain unchanged and accept, roll back, or stop edits at each step.
This gives image owners direct control over the information their images reveal.The system is publicly accessible, with an example in~\refappendix{appendix:interactive_geocloak}.
Future work will include real-world user feedback.

\mypara{Manual Inspection of \geocloak's Outputs}
Throughout system development, all authors manually inspect randomly sampled intermediate outputs from \geocloak's agent trajectories.
We specifically examine whether the agents' intended actions are successfully executed and whether subsequent agent decisions are consistent with the observed privacy and utility feedback.
By the final system iteration, the sampled trajectories generally exhibit successful execution of the proposed edits and coherent multi-turn refinement behavior.

\mypara{LVLM Capability}
\geocloak's effectiveness depends on the underlying LVLM's visual reasoning capability, as stronger models can identify more comprehensive clues and enable more targeted protection.
Our evaluation against 11 attacker configurations, including stronger models and different model families, shows substantial reductions in geolocation accuracy under the current stopping rule.
A final GT-guided audit after standard-mode termination could provide an additional check for residual clues, although it would still depend on \clueagent's ability to identify them.
Its additional protection benefit and processing overhead remain to be evaluated.
This dependence allows \geocloak to benefit from future model advances without architectural changes.
Our open-source Qwen configuration demonstrates that \geocloak does not require proprietary backbones, although its 7.4\% attack success rate remains higher than the 4.0\% of our primary Gemini configuration (\autoref{table:design_choice}).
Improving protection with smaller, resource-constrained open-source backbones remains important future work.

\section{Conclusion}
\label{section:conclusion}

We present \geocloak, an agentic framework for geolocation privacy protection.
Our approach is motivated by an observation visual-geocalocation asymmetry: geolocation-decisive clues tend to be visually peripheral, enabling targeted protection with minimal impact on image quality.
\geocloak exploits this asymmetry through a combination of targeted editing and adversarial perturbation, driven by an agentic system that adaptively analyzes each image and iteratively refines its protection strategies under utility-aware feedback.
Extensive evaluation demonstrates that our agent outperforms baselines in both effectiveness and utility.
We further release an interactive version of \geocloak that allows users to directly participate in the protection process.
We hope this work contributes to the development of practical, user-intent-aware geolocation privacy protection, and inspires further research into semantics-preserving approaches to visual privacy in the LVLM era.

\bibliographystyle{plainurl}
\bibliography{normal_generated_py3}

\appendix 

\section*{Open Science}

In this paper, we provide the code for interactive version of \geocloak to faciliate the real-world deployment.
An example screenshot is provided in \refappendix{appendix:interactive_geocloak}, and the (anonymous) artifact is publicly accessible at \url{https://anonymous.4open.science/r/GeoCloak-7AA4}.
All evaluation datasets used in this work are publicly available: Im2GPS3k~\cite{im2gps3k_dataset}, GeoBench~\cite{geobench_dataset}, Dataset-GSS~\cite{imageo_dataset}, and Dataset-PCW~\cite{imageo_dataset}.
In our evaluation, we access all evaluated LVLMs via their official APIs.

\section*{Ethical Considerations} 

\mypara{Our Primary Goal}
\geocloak is designed exclusively as a privacy protection tool, intended to help individuals prevent unintended location disclosure when sharing images online.
We do not anticipate significant dual-use concerns, as the system is designed to suppress geolocation information rather than enable tracking.

\mypara{Dataset and Image Usage}
All datasets used in our evaluation are publicly available benchmarks.
Images used in qualitative examples are either from these public benchmarks.

\mypara{Human Annotation Tasks}
Our annotation tasks were exempt from IRB review as they involved no sensitive data collection, no deception, and no risk to participants.
The tasks were benign in nature and did not involve any sensitive, disturbing, or personally identifiable content.
Annotators evaluated whether images contained intentionally photographed landmarks and rated the perceptual quality of protected benchmark images.
No annotator reported any discomfort during the study.

\mypara{Potential for Misuse}
While \geocloak is designed for privacy protection, the underlying image editing capabilities could theoretically be misused to manipulate images for deceptive purposes.
However, the system is specifically optimized for subtle, targeted edits that preserve visual semantics, making it less suitable for generating deceptive content compared to general-purpose image generation tools.

\begin{figure*}[!ht]
\centering
\includegraphics[width=\textwidth]{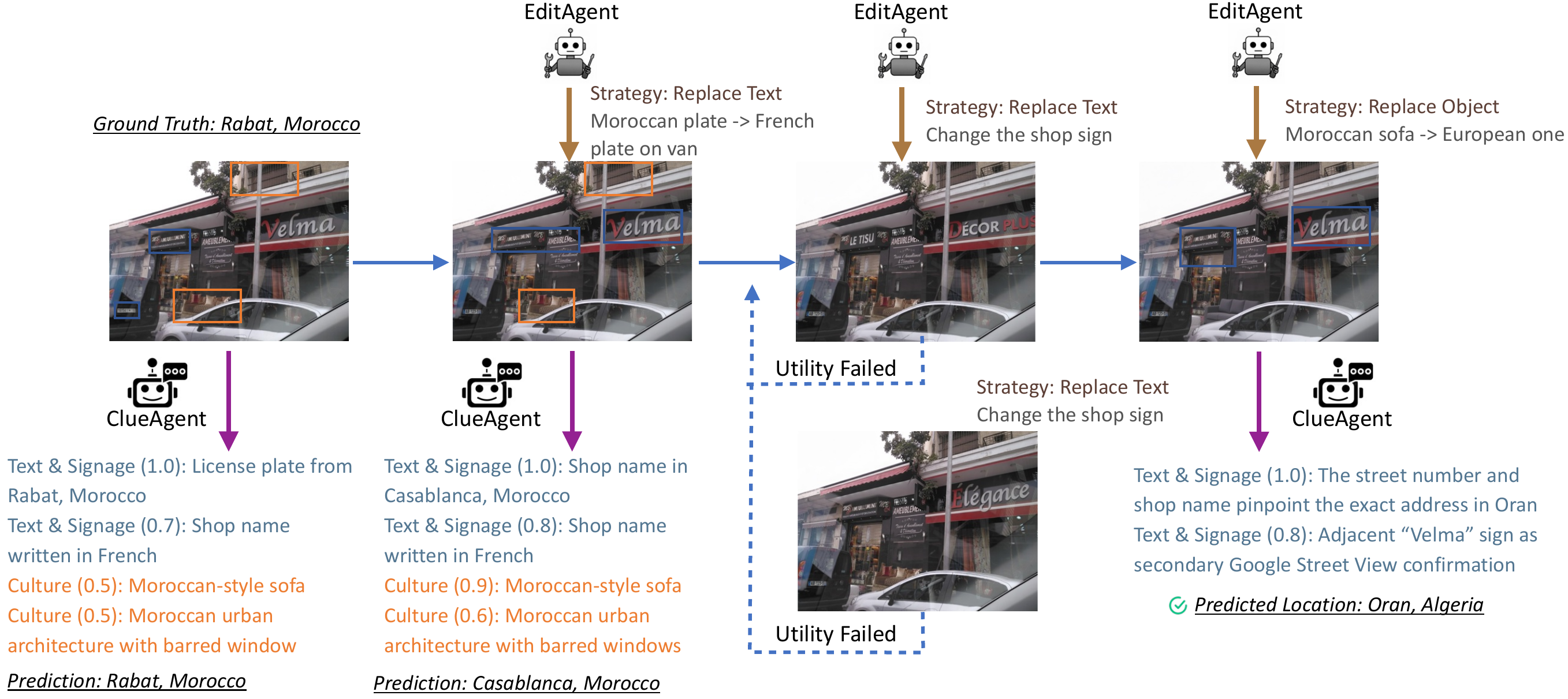}
\caption{A multi-turn \geocloak protection example.
Starting from Rabat, Morocco, \editagent first replaces the Moroccan license plate with a French one, shifting \clueagent's prediction to Casablanca, Morocco.
A subsequent attempt to change the shop sign fails the utility check twice and is rolled back.
\editagent then switches strategy to replace the Moroccan sofa with a European one, successfully redirecting the predicted location to Oran, Algeria, with minimal visual change across all turns.}
\label{figure:iterate_example}
\end{figure*}

\section{Examples of Multi-Turn \geocloak Protection Example}
\label{appendix:protection_example}

We exhibit an successful example of multi-turn \geocloak protection example in~\autoref{figure:iterate_example}.

\section{Prompt Templates}
\label{appendix:prompt_templates}

In this section, we provide the prompt templates for \clueagent (\autoref{lst:clueagent}), \editagent (\autoref{lst:editagent}), semantic editing skill (\autoref{lst:editing}), adversarial perturbation skill (\autoref{lst:adv}), and evaluation scoping (\autoref{lst:scope}).

\section{Examples of Semantic Editing Skill}
\label{appendix:editing}

\autoref{table:editing_strategies} exhibit examples of each strategy of the semantic editing skill.
In all cases, modifications are subtle and semantics-preserving, yet sufficient to redirect \clueagent's predicted location.

\begin{table*}[ht]
\centering
\footnotesize
\renewcommand{\arraystretch}{1.0}
\begin{tabular}{>{\centering\arraybackslash}m{0.09\textwidth}|>{\centering\arraybackslash}m{0.14\textwidth}|>{\centering\arraybackslash}m{0.2\textwidth}|>{\centering\arraybackslash}m{0.2\textwidth}|>{\centering\arraybackslash}m{0.25\textwidth}}
\toprule
\textbf{Strategy} & \textbf{Application Scenario} & \textbf{Original} & \textbf{Protected} & \textbf{How it protects} \\
\midrule
\begin{minipage}[c][2.2cm][c]{\linewidth}\centering\rotatebox[origin=c]{90}{\textbf{\shortstack{Replace\\Text}}}\end{minipage} & \begin{minipage}[c][2.2cm][c]{\linewidth}\RaggedRight Signs, street names, or any visible text reveal the location.\end{minipage} & \begin{minipage}[c][2.2cm][c]{\linewidth}\centering\includegraphics[width=\linewidth,height=2.2cm,keepaspectratio]{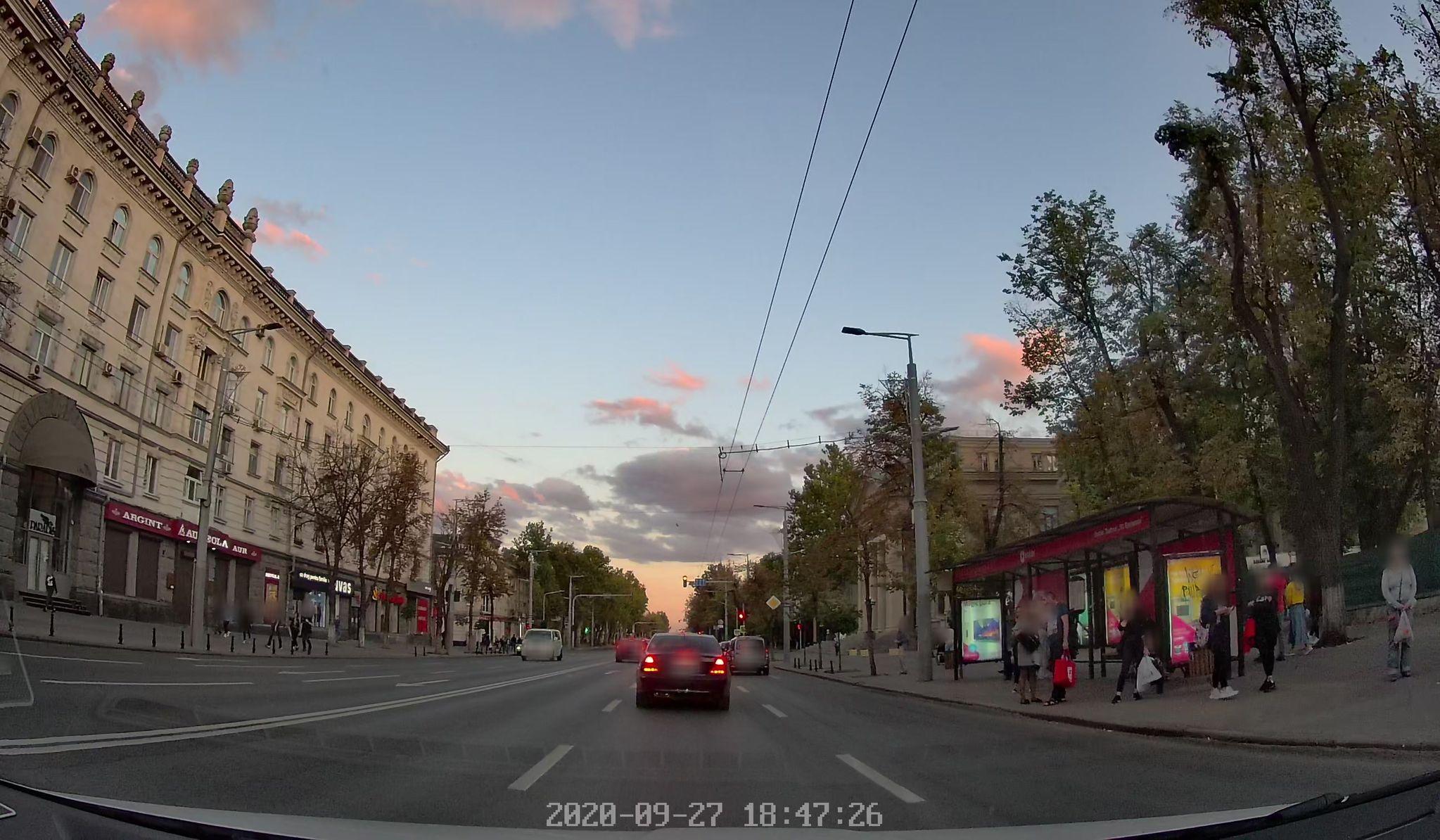}\end{minipage} & \begin{minipage}[c][2.2cm][c]{\linewidth}\centering\includegraphics[width=\linewidth,height=2.2cm,keepaspectratio]{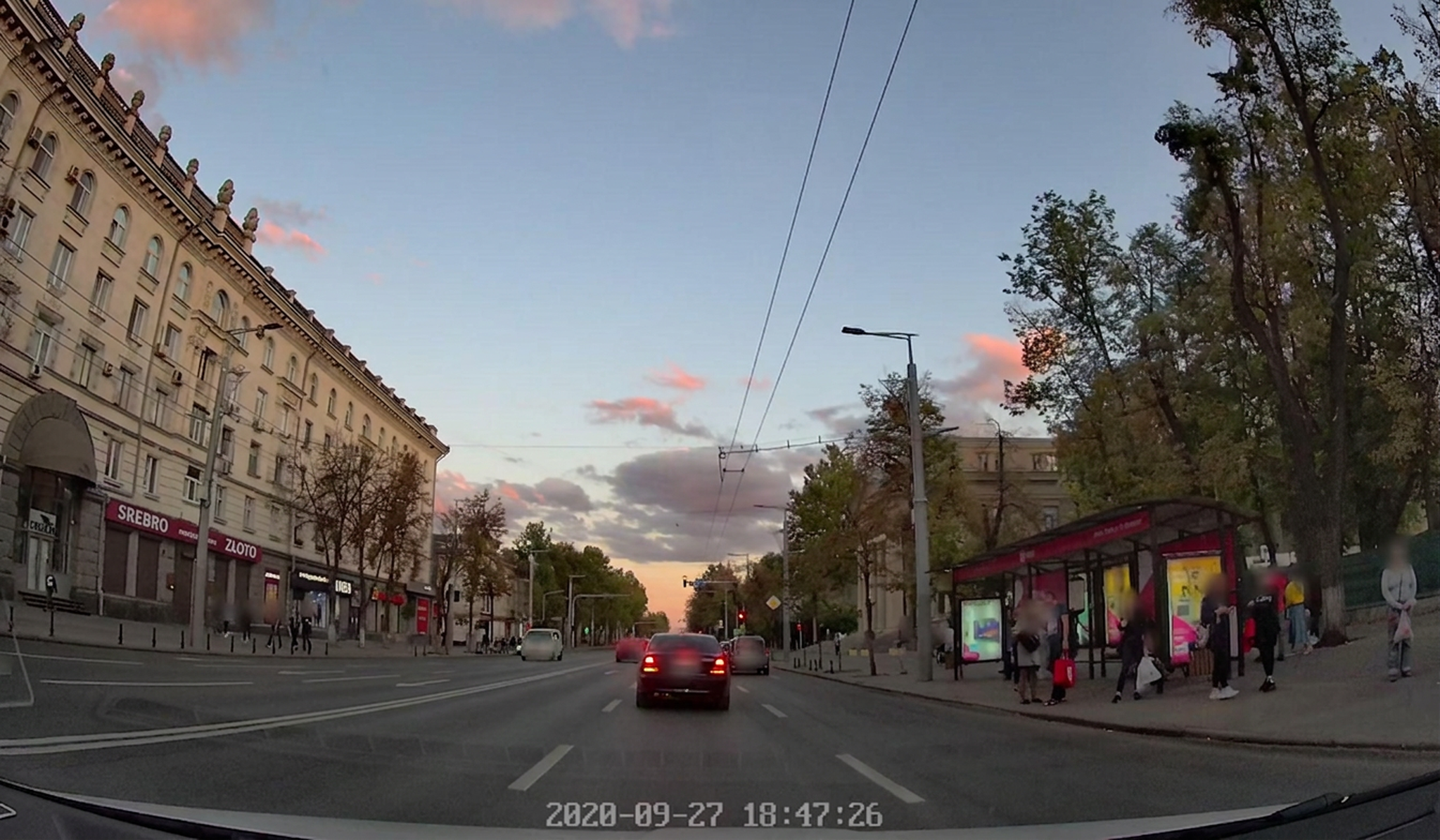}\end{minipage} & \begin{minipage}[c][2.2cm][c]{\linewidth}\RaggedRight Replacing Romanian sign text on the red shop signs with Polish words shifts the language clue and misleads the predicted location from Chi\c{s}in\u{a}u, Moldova to Warsaw, Poland.\end{minipage} \\
\hline
\begin{minipage}[c][2.2cm][c]{\linewidth}\centering\rotatebox[origin=c]{90}{\textbf{\shortstack{Replace\\Object}}}\end{minipage} & \begin{minipage}[c][2.2cm][c]{\linewidth}\RaggedRight Location-specific objects are present (e.g., post boxes and traffic signs).\end{minipage} & \begin{minipage}[c][2.2cm][c]{\linewidth}\centering\includegraphics[width=\linewidth,height=2.2cm,keepaspectratio]{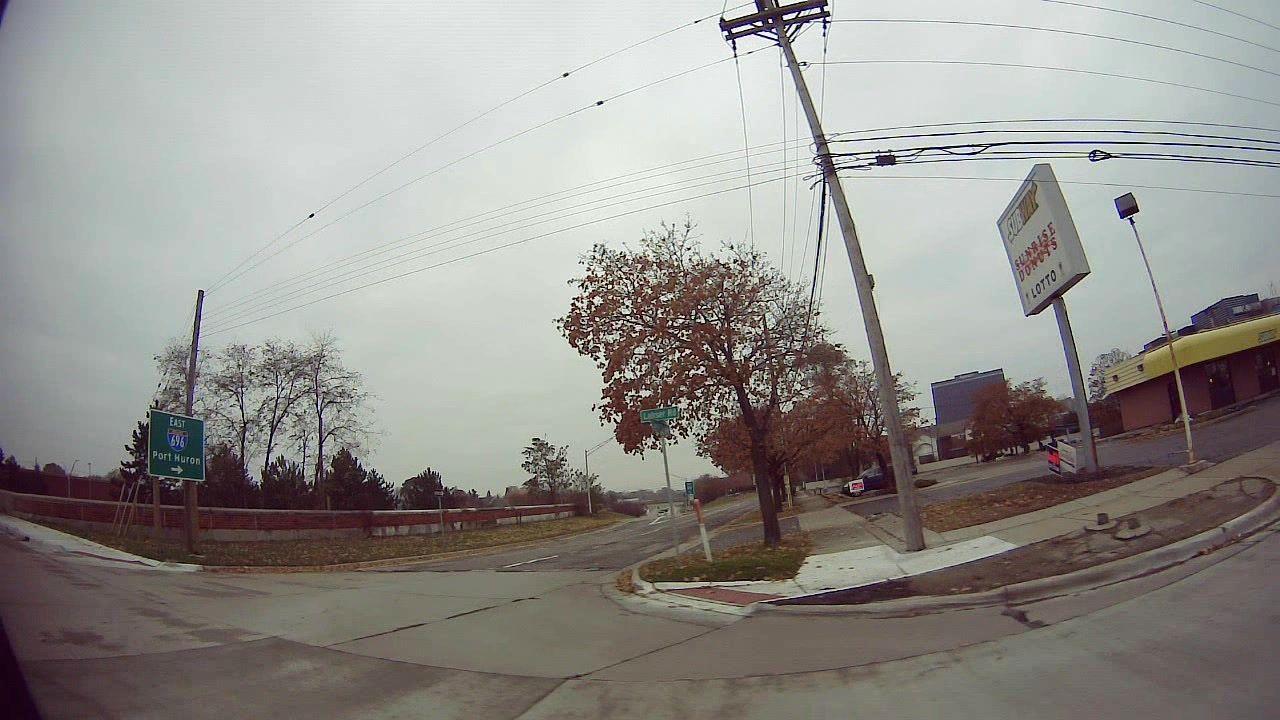}\end{minipage} & \begin{minipage}[c][2.2cm][c]{\linewidth}\centering\includegraphics[width=\linewidth,height=2.2cm,keepaspectratio]{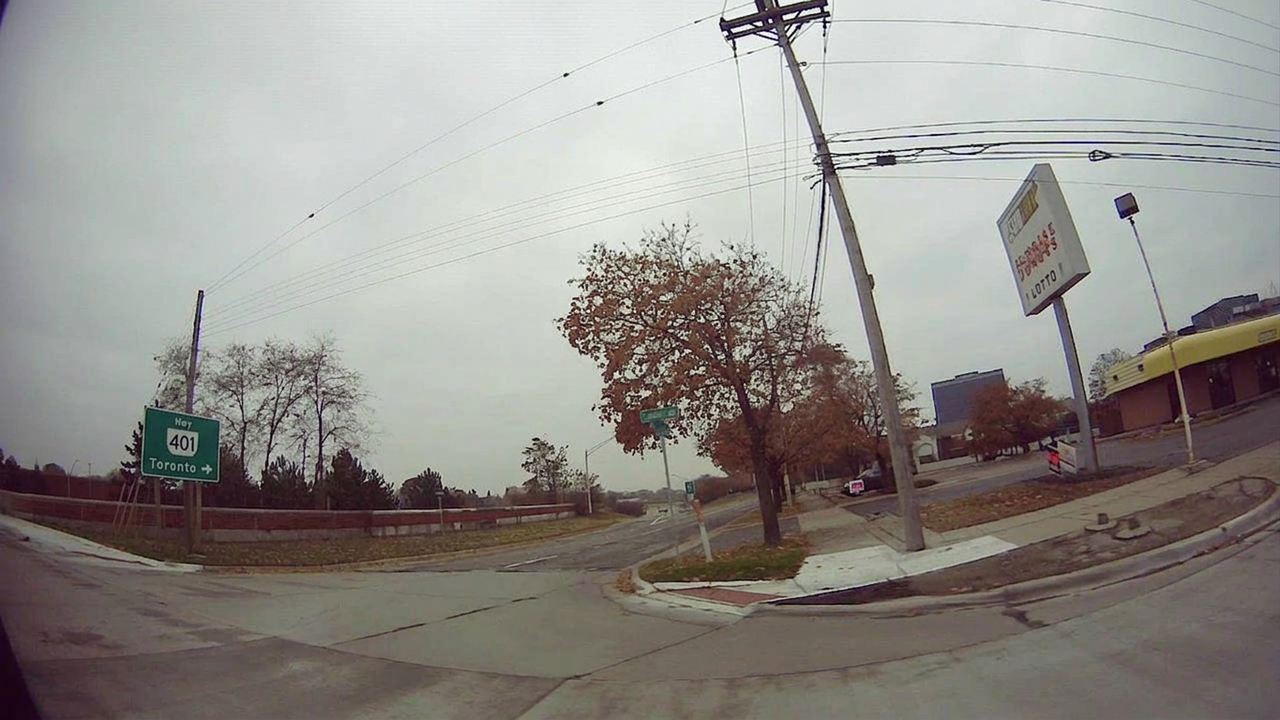}\end{minipage} & \begin{minipage}[c][2.2cm][c]{\linewidth}\RaggedRight Replacing the American highway sign  with a Canadian highway sign shifts the predicted location from Southfield, Michigan, USA, to Windsor, Ontario, Canada.\end{minipage} \\
\hline
\begin{minipage}[c][2.2cm][c]{\linewidth}\centering\rotatebox[origin=c]{90}{\textbf{\shortstack{Erase\\Object}}}\end{minipage} & \begin{minipage}[c][2.2cm][c]{\linewidth}\RaggedRight An object, sign, text, or logo is too specific to replace but small enough to remove cleanly.\end{minipage} & \begin{minipage}[c][2.2cm][c]{\linewidth}\centering\includegraphics[width=\linewidth,height=2.2cm,keepaspectratio]{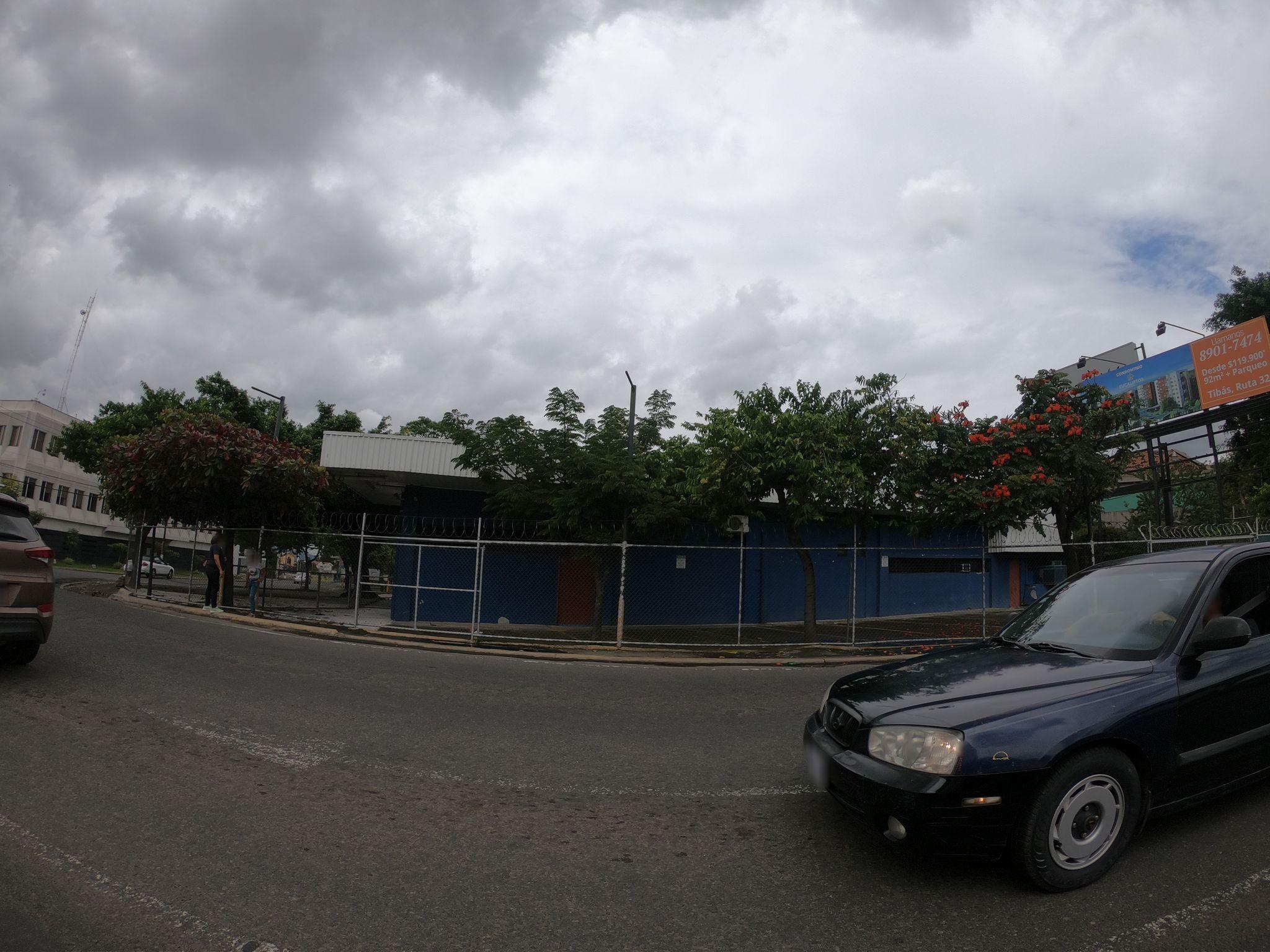}\end{minipage} & \begin{minipage}[c][2.2cm][c]{\linewidth}\centering\includegraphics[width=\linewidth,height=2.2cm,keepaspectratio]{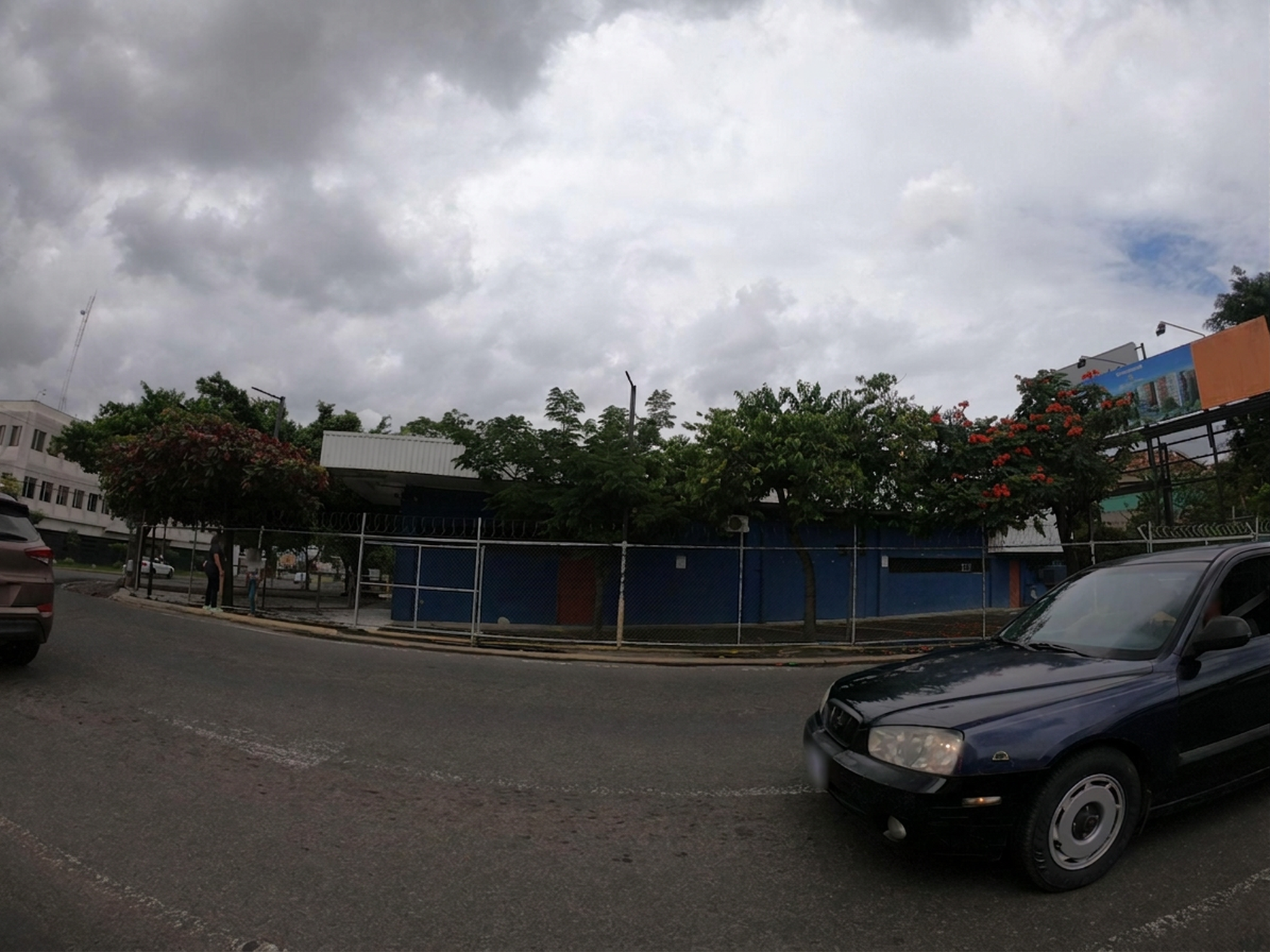}\end{minipage} & \begin{minipage}[c][2.2cm][c]{\linewidth}\RaggedRight Removing all text from the orange billboard shifts the predicted location from San José, Costa Rica, to San Salvador, El Salvador.\end{minipage} \\
\hline
\begin{minipage}[c][2.2cm][c]{\linewidth}\centering\rotatebox[origin=c]{90}{\textbf{\shortstack{Add\\Decoy}}}\end{minipage} & \begin{minipage}[c][2.2cm][c]{\linewidth}\RaggedRight Dominant clues are too large or structural to edit or perturb directly.\end{minipage} & \begin{minipage}[c][2.2cm][c]{\linewidth}\centering\includegraphics[width=\linewidth,height=2.2cm,keepaspectratio]{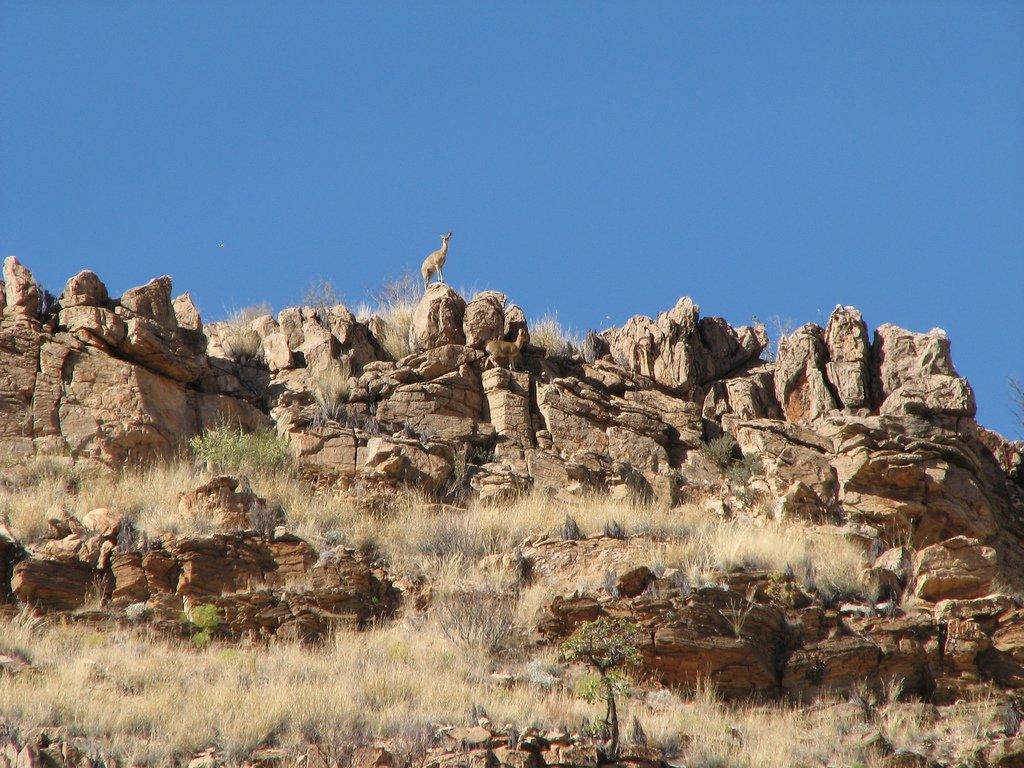}\end{minipage} & \begin{minipage}[c][2.2cm][c]{\linewidth}\centering\includegraphics[width=\linewidth,height=2.2cm,keepaspectratio]{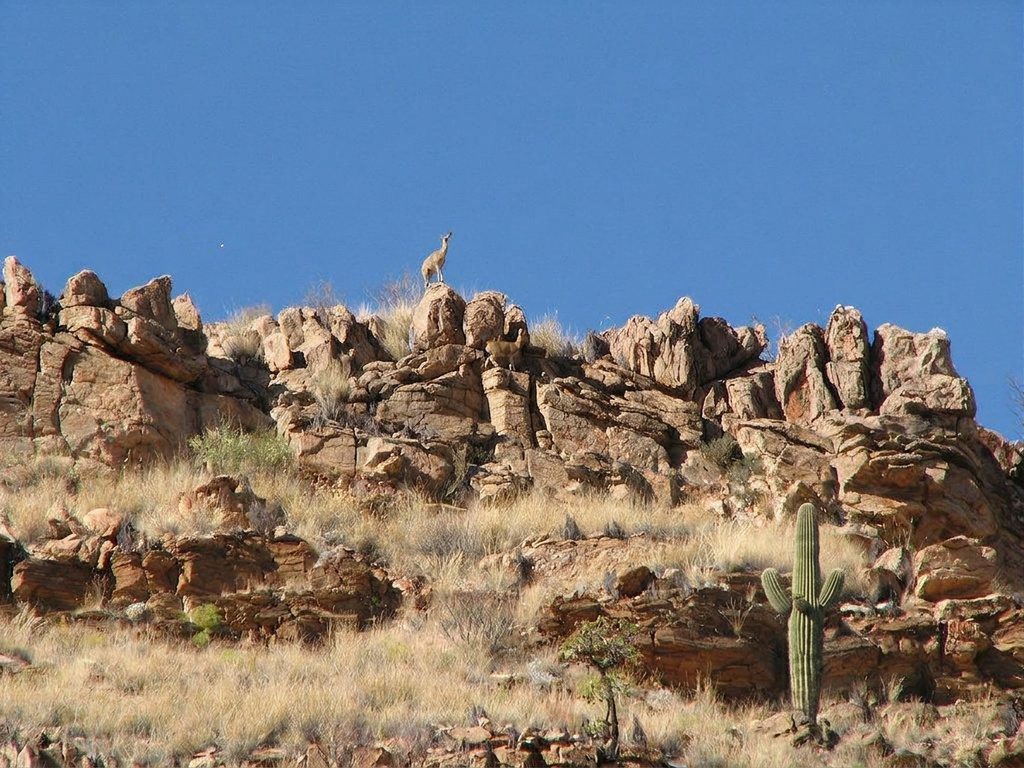}\end{minipage} & \begin{minipage}[c][2.2cm][c]{\linewidth}\RaggedRight Adding a small Saguaro cactus growing among the rocks shifts the predicted location from Khomas, Namibia to Tucson, Arizona, US.\end{minipage} \\
\hline
\begin{minipage}[c][2.2cm][c]{\linewidth}\centering\rotatebox[origin=c]{90}{\textbf{\shortstack{Style\\Transfer}}}\end{minipage} & \begin{minipage}[c][2.2cm][c]{\linewidth}\RaggedRight Apply only to small or background elements.\end{minipage} & \begin{minipage}[c][2.2cm][c]{\linewidth}\centering\includegraphics[width=\linewidth,height=2.2cm,keepaspectratio]{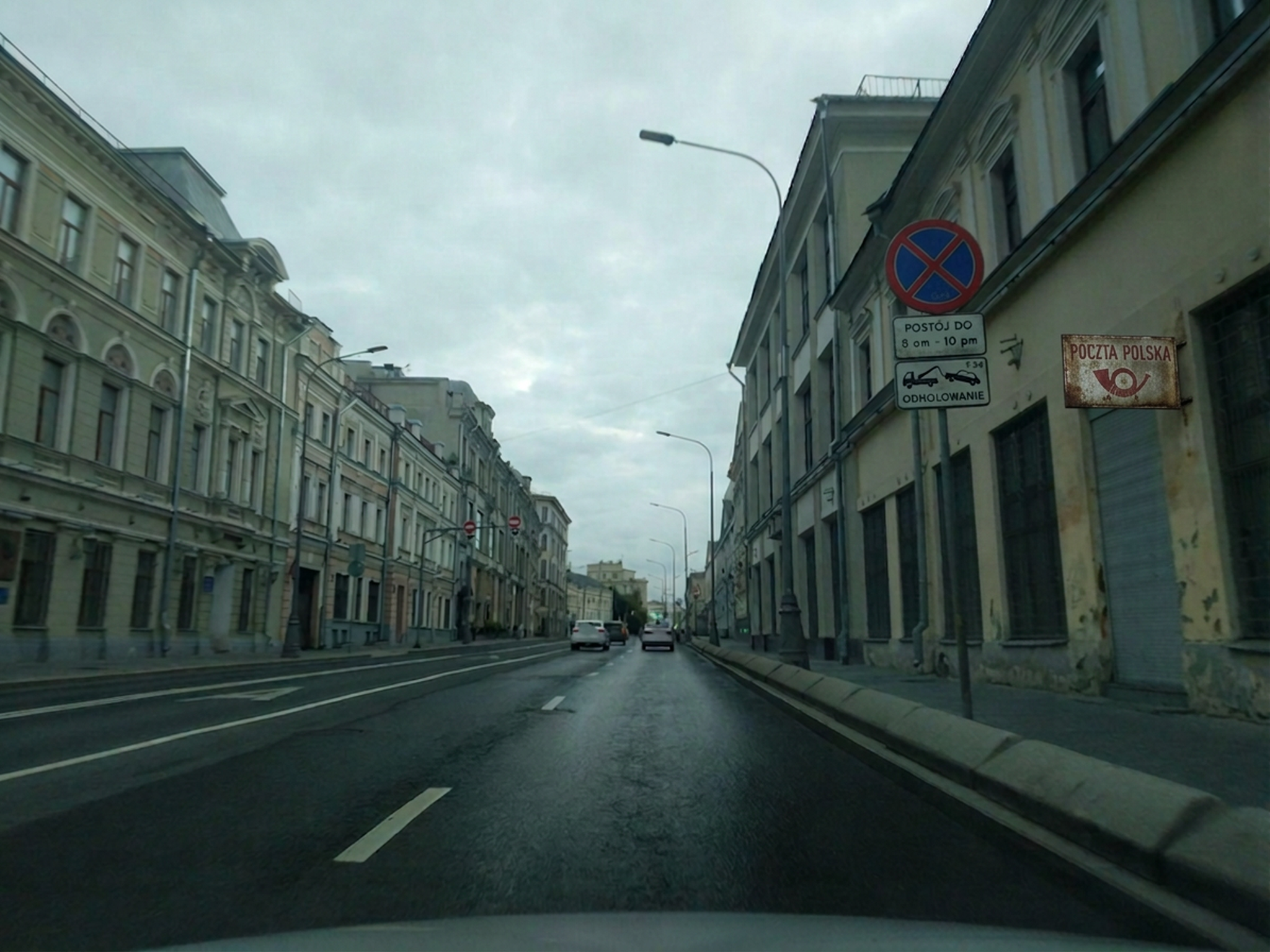}\end{minipage} & \begin{minipage}[c][2.2cm][c]{\linewidth}\centering\includegraphics[width=\linewidth,height=2.2cm,keepaspectratio]{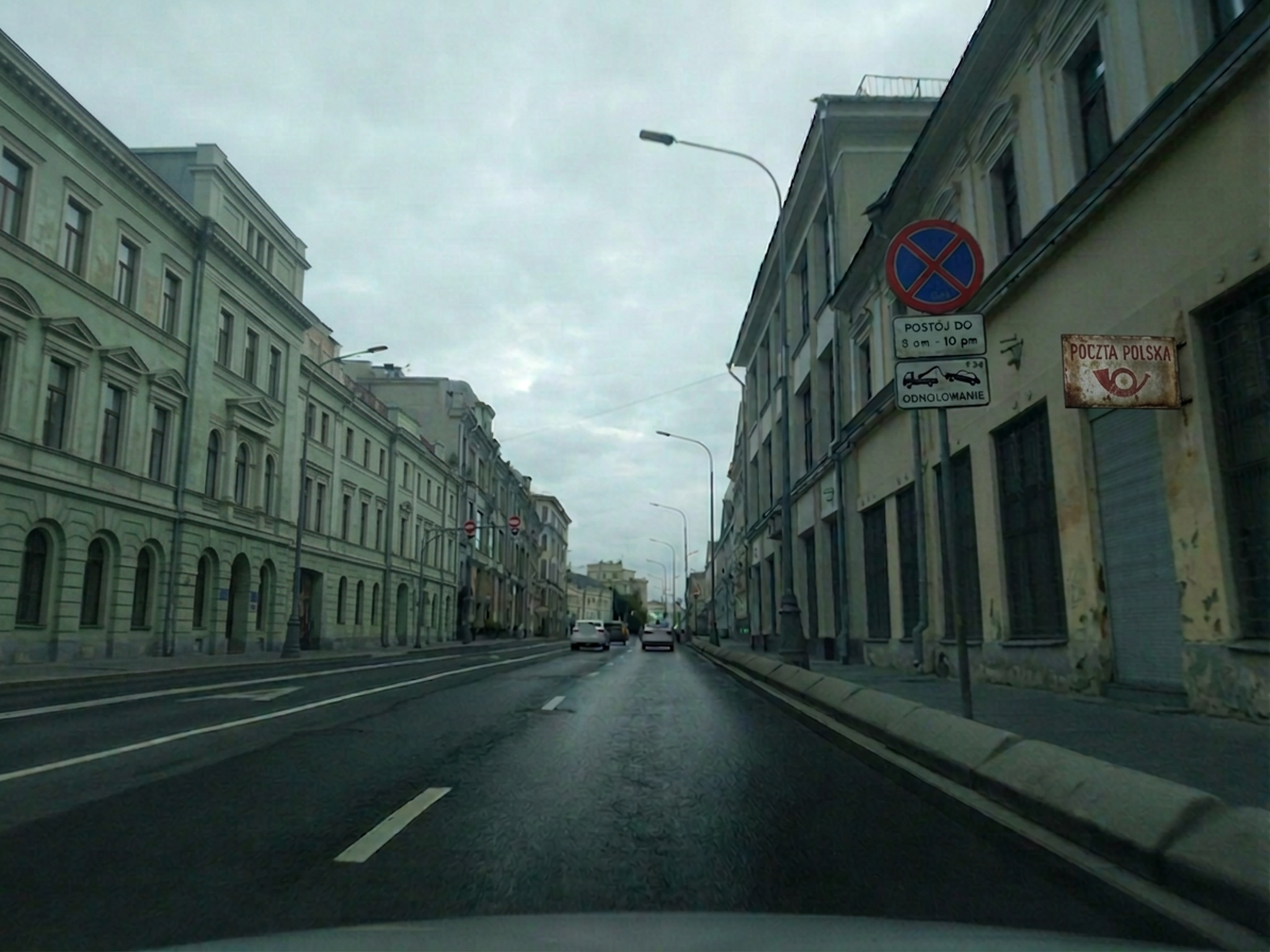}\end{minipage} & \begin{minipage}[c][2.2cm][c]{\linewidth}\RaggedRight Changing the architectural style of the greenish building on the left from Russian neoclassical to German neoclassical shifts the predicted location from Moscow, Russia to Warsaw, Poland.\end{minipage} \\
\bottomrule
\end{tabular}
\caption{Overview of the five Semantic Editing strategies in \editagent.
The \editagent generates plans for modifying the image based on the clues obtained from \clueagent.
In all cases, modifications are subtle and semantics-preserving, yet sufficient to redirect \clueagent's predicted location.}
\label{table:editing_strategies}
\end{table*}

\begin{table}[ht]
\centering
\caption{Geographic coverage of the four benchmark datasets.
}
\label{table:dataset_coverage}
\scalebox{0.65}{
\begin{tabular}{lrrr}
\toprule
\textbf{Dataset} & \textbf{\#Samples} & \textbf{\#Countries} & \textbf{\#Cities} \\
\midrule
IM2GPS3K        & 500 & 74 & 297 \\
Dataset-GSS        & 500 & 61 & 161 \\
Dataset-PCW        & 267 &  6 &  76 \\
GeoBench   & 512 & 64 & 117 \\
\midrule
\textbf{Overall} & \textbf{1,779} & \textbf{108} & \textbf{576} \\
\bottomrule
\end{tabular}}
\end{table}

\section{More Details of Evaluation Setup}
\label{appendix:eval_setup}

\subsection{Evaluation Datasets}
\label{appendix:datasets}

Im2GPS3k~\cite{VJH17} and YFCC4k~\cite{TSFENPBL16} were originally introduced to evaluate retrieval- and classification-based methods, and remain widely adopted as standard evaluation protocols in the LVLM era~\cite{JLLZWDHWWY24,LDDLZSZGL24,MCHDXR24,JPGZL25,LZLLZXX25}.
Newer benchmarks emerged to assess reasoning quality and broader capabilities~\cite{WLWHLR25,SYHTZCFGC25,LYHLDZJ25,CWWHL25,LZLLZXX25}.
In this paper, we evaluate \geocloak on four datasets that support level-wise geolocation evaluation to comprehensively assess protection effectiveness across diverse geographic and visual contexts.

\noindent \textit{Im2GPS3k}~\cite{VJH17} is a widely adopted benchmark in geolocation research, containing 3,000 geo-tagged images spanning diverse locations worldwide.

\noindent \textit{GeoBench}~\cite{WLWHLR25} consists of high-resolution photos with global coverage, specifically designed to challenge current LVLM with high localizability requirements.

\noindent \textit{Dataset-GSS}~\cite{LYHLDZJ25} is sourced from IMAGEO-Bench.
It is drawn from the NUS Global Streetscapes dataset~\cite{HQKYOIWZB24} across 688 cities worldwide.
The IMAGEO-Bench authors further filtered the original data, manually excluding images with insufficient geographic information, resulting in a diverse collection spanning varied geographies, environmental conditions, architectural styles, and cultural contexts.

\noindent \textit{Dataset-PCW}~\cite{LYHLDZJ25} is also sourced from IMAGEO-Bench, but does not draw from any publicly available sources, making it well-suited to evaluate geolocation protection on truly unseen images.

\noindent We use all available data from GeoBench and Dataset-PCW.
Because Im2GPS3k and Dataset-GSS are substantially larger, we randomly sample 500 images from each dataset.
As reported in~\autoref{table:dataset_coverage}, the resulting benchmark contains 1,779 images spanning 108 unique countries and 576 unique cities across all four datasets, ensuring broad geographic diversity in our evaluation.
This scale exceeds the 500-image DoxBench evaluation used in the GeoMiner study~\cite{LZLLZXX25} while providing substantially broader geographic coverage.
Across the available images, the full evaluation design comprises over 75,000 evaluation runs, covering 11 attacker configurations and four conditions (Initial, M-Attack, GeoShield, and \geocloak).

\subsection{Geolocation Inference Methods}
\label{appendix:inference_method}

\noindent \textit{Zero-Shot} is a simple single-turn prompt that asks \textit{``Where was this photo taken? Explain your reasoning based on visual evidence,''} along with a structured output format to ensure consistent responses.
Despite its simplicity, this prompt already elicits strong geolocation performance from LVLMs~\cite{LDDLZSZGL24,LZLLZXX25}, reflecting a realistic attack scenario where any user can infer location from a shared image without any prompt engineering expertise.

\noindent \textit{GeoMiner} is a two-stage collaborative attack framework from Dox-Bench that leverages two LVLM in a pipeline.
In the first stage, a \textit{Detector} model processes the image and produces a structured set of geolocation-relevant visual clues organized by category.
These clues are then passed to an \textit{Analyzer} model in the second stage, which incorporates them as explicit context to generate a more accurate location prediction.

\noindent \textit{GeoVista} is a web-augmented agentic framework that operates through an iterative thought-action-observation loop.
At each step, the model generates a reasoning thought and autonomously decides when and how to invoke tools, enabling a dynamic process of visual clue extraction and external information retrieval.
Two tools are available: \textit{image\_zoom\_in}, which crops and magnifies regions of interest specified by a bounding box, and \textit{search\_web}, which retrieves up to 10 relevant documents from the internet.

For GeoMiner and GeoVista, we follow the official GitHub implementations with default settings.

\subsection{SOTA Methods for Comparison}
\label{appendix:comparison_methods}

GeoShield and M-Attack are reproduced based on the released code~\cite{geoshield}, following all original hyperparameters: a CLIP ensemble of ViT-B/16, ViT-B/32, and CLIP-Laion backbones, perturbation budget $\epsilon$ = 8/255, step size $\alpha$ = 1/255, 100 PGD steps, input resolution 640$x$640, and RandomResizedCrop augmentation (scale 0.5–0.9) for transferability.
For GeoShield, geo-revealing regions are identified via GroundingDINO~\cite{LZRLZYJLYSZZ24}, and non-geographical descriptions are generated by GPT-5-mini.
For M-Attack, target images are sampled randomly from the same dataset at different locations.

\section{Fine-Grained Complexity and Failure-Mode Analysis}
\label{appendix:geo_identifiability}

We first investigate whether the number of geolocation clues provides a useful measure of image complexity.
We use Gemini-2.5-Pro with GeoMiner~\cite{LZLLZXX25} to extract structured clues from each original image and compare the number of extracted clues with the mean number of successful attackers.

\begin{table}[ht]
\centering
\caption{Mean number of successful attackers per image, stratified by the number of clues extracted by GeoMiner.}
\label{table:clue_complexity}
\scalebox{0.6}{%
\begin{tabular}{rrrrr}
\toprule
\textbf{\#Clues} & \textbf{Initial} & \textbf{\geocloak} & \textbf{GeoShield} & \textbf{M-Attack} \\
\midrule
2    & 3.83/11 & \textbf{0.00/11} & 1.67/11 & 1.33/11 \\
3    & 5.89/11 & \textbf{1.11/11} & 2.22/11 & 3.33/11 \\
4    & 6.39/11 & \textbf{1.35/11} & 3.36/11 & 4.00/11 \\
5    & 6.76/11 & \textbf{1.70/11} & 3.02/11 & 3.99/11 \\
6    & 6.58/11 & \textbf{1.74/11} & 3.17/11 & 3.82/11 \\
7    & 5.12/11 & \textbf{1.29/11} & 2.35/11 & 2.29/11 \\
8    & 7.40/11 & \textbf{0.40/11} & 6.00/11 & 5.80/11 \\
\midrule
\textbf{Overall} & \textbf{6.55/11} & \textbf{1.57/11} & \textbf{3.13/11} & \textbf{3.90/11} \\
\bottomrule
\end{tabular}}
\end{table}

As shown in~\autoref{table:clue_complexity}, the number of extracted clues has no monotonic relationship with initial attack success.
For example, images with five clues have 6.76 successful attackers on average, whereas those with seven clues have only 5.12.
This indicates that clue specificity and discriminative strength matter more than the raw number of clues, making clue count an unreliable proxy for geolocation difficulty.

We therefore use a direct empirical measure of image difficulty.
We define an image's \emph{initial geo-identifiability} as the number of attacker configurations that successfully geolocate it at the city level under no defense.
The score ranges from 1 to 11, corresponding to zero-shot and GeoMiner inference with five LVLMs plus GeoVista.
We analyze all non-exempt images across the four datasets and divide them into low-, medium-, and high-identifiability groups.

\begin{table}[ht]
\centering
\caption{Mean number of successful attackers per image, stratified by initial geo-identifiability.
Lower is better for all defense columns.}
\label{table:geo_identifiability}
\scalebox{0.6}{%
\begin{tabular}{lrrrr}
\toprule
\textbf{Geo-identifiability} & \textbf{Initial} & \textbf{\geocloak} & \textbf{GeoShield} & \textbf{M-Attack} \\
\midrule
Low (1--4/11)    & 2.18/11 & \textbf{0.35/11} & 0.39/11 & 0.57/11 \\
Medium (5--7/11) & 5.98/11 & \textbf{0.86/11} & 1.73/11 & 2.39/11 \\
High (8--11/11)  & 9.84/11 & \textbf{2.75/11} & 5.69/11 & 6.91/11 \\
\midrule
\textbf{Overall} & \textbf{6.55/11} & \textbf{1.57/11} & \textbf{3.13/11} & \textbf{3.90/11} \\
\bottomrule
\end{tabular}}
\end{table}

As shown in~\autoref{table:geo_identifiability}, \geocloak leaves the fewest successful attackers and yields the largest attacker reduction in every group.
The difference is most pronounced for highly geo-identifiable images: \geocloak reduces the mean from 9.84 to 2.75 successful attackers, compared with 5.69 for GeoShield and 6.91 for M-Attack.
To understand where \geocloak can struggle, we further examine low-identifiability images.
These images expose only weak or ambiguous geographic evidence, making it difficult for \clueagent to identify actionable clues and leaving less evidence for the protection agent to suppress.
Consequently, \geocloak's advantage is less pronounced.
Nevertheless, it still leaves the lowest mean number of successful attackers, outperforming both GeoShield and M-Attack.

\section{Evaluation Results Measured by Multiple Metrics}
\label{appendix:other_metrics}

As illustrated in~\autoref{table:main_eval_full_metric}, we report both level-wise accuracy and distance-based accuracy.
We observe that our agent still achieve better protection performance than two baselines.

\begin{table*}[ht]
\centering
\caption{Geolocation accuracy (\%) under no defense and each defense (aggregated across all datasets).
All metrics $\downarrow$ lower is better.}
\label{table:main_eval_full_metric}
\scalebox{0.6}{%
\begin{tabular}{lll | rrr | rrr}
\toprule
\multicolumn{3}{l}{} & \multicolumn{3}{c}{Level-Wise Acc.} & \multicolumn{3}{c}{Distance Acc.} \\
\cmidrule(lr){4-6}\cmidrule(lr){7-9}
Model & Method & Defense & City$\downarrow$ & State$\downarrow$ & Country$\downarrow$ & @1km$\downarrow$ & @25km$\downarrow$ & @100km$\downarrow$ \\
\midrule
\multirow{8}{*}{Gemini-2.5-Pro} & \multirow{4}{*}{Zero-Shot} & Initial & 48.3\% & 58.9\% & 87.9\% & 16.1\% & 55.2\% & 61.6\% \\
 &  & M-Attack & 32.3\% & 36.5\% & 55.2\% & 11.2\% & 34.8\% & 38.0\% \\
 &  & GeoShield & 26.0\% & 27.7\% & 44.4\% & 9.6\% & 26.7\% & 29.8\% \\
 &  & \geocloak & \textbf{9.5\%} & \textbf{10.9\%} & \textbf{18.0\%} & \textbf{2.8\%} & \textbf{9.8\%} & \textbf{11.9\%} \\
\cmidrule{2-9}
 & \multirow{4}{*}{GeoMiner} & Initial & 46.9\% & 59.1\% & 88.2\% & 30.2\% & 53.7\% & 62.1\% \\
 &  & M-Attack & 29.8\% & 31.4\% & 39.8\% & 18.7\% & 30.1\% & 31.2\% \\
 &  & GeoShield & 25.3\% & 27.1\% & 34.9\% & 15.9\% & 26.1\% & 27.6\% \\
 &  & \geocloak & \textbf{9.4\%} & \textbf{11.0\%} & \textbf{16.9\%} & \textbf{4.7\%} & \textbf{9.9\%} & \textbf{11.6\%} \\
\midrule
\multirow{8}{*}{Gemini-3.1-Pro} & \multirow{4}{*}{Zero-Shot} & Initial & 51.3\% & 60.1\% & 86.3\% & 16.3\% & 54.4\% & 59.7\% \\
 &  & M-Attack & 37.3\% & 38.9\% & 45.0\% & 12.8\% & 36.9\% & 37.8\% \\
 &  & GeoShield & 33.2\% & 34.6\% & 42.1\% & 11.3\% & 33.1\% & 34.4\% \\
 &  & \geocloak & \textbf{13.8\%} & \textbf{15.3\%} & \textbf{24.9\%} & \textbf{4.0\%} & \textbf{14.5\%} & \textbf{15.7\%} \\
\cmidrule{2-9}
 & \multirow{4}{*}{GeoMiner} & Initial & 48.9\% & 58.6\% & 79.6\% & 26.9\% & 43.2\% & 48.3\% \\
 &  & M-Attack & 33.9\% & 35.6\% & 39.7\% & 18.9\% & 26.7\% & 27.9\% \\
 &  & GeoShield & 29.9\% & 31.2\% & 37.6\% & 17.7\% & 23.7\% & 24.6\% \\
 &  & \geocloak & \textbf{13.3\%} & \textbf{15.2\%} & \textbf{23.6\%} & \textbf{4.2\%} & \textbf{8.8\%} & \textbf{10.0\%} \\
\midrule
\multirow{8}{*}{GPT-5.4} & \multirow{4}{*}{Zero-Shot} & Initial & 39.5\% & 50.0\% & 76.6\% & 12.5\% & 44.1\% & 50.6\% \\
 &  & M-Attack & 21.5\% & 22.4\% & 27.4\% & 6.9\% & 21.4\% & 22.1\% \\
 &  & GeoShield & 17.1\% & 17.8\% & 23.7\% & 6.0\% & 17.1\% & 17.7\% \\
 &  & \geocloak & \textbf{12.2\%} & \textbf{12.8\%} & \textbf{17.8\%} & \textbf{3.9\%} & \textbf{12.4\%} & \textbf{12.7\%} \\
\cmidrule{2-9}
 & \multirow{4}{*}{GeoMiner} & Initial & 37.3\% & 48.0\% & 76.5\% & 20.8\% & 41.0\% & 46.8\% \\
 &  & M-Attack & 21.5\% & 23.0\% & 28.5\% & 11.9\% & 21.7\% & 23.0\% \\
 &  & GeoShield & 16.4\% & 17.1\% & 23.3\% & 9.1\% & 16.9\% & 17.1\% \\
 &  & \geocloak & \textbf{7.3\%} & \textbf{8.2\%} & \textbf{13.2\%} & \textbf{2.5\%} & \textbf{7.7\%} & \textbf{8.2\%} \\
\midrule
\multirow{8}{*}{o3} & \multirow{4}{*}{Zero-Shot} & Initial & 46.9\% & 58.0\% & 84.6\% & 15.1\% & 53.2\% & 59.7\% \\
 &  & M-Attack & 25.2\% & 27.3\% & 34.5\% & 8.6\% & 25.2\% & 25.8\% \\
 &  & GeoShield & 17.7\% & 18.3\% & 26.3\% & 6.6\% & 17.7\% & 18.5\% \\
 &  & \geocloak & \textbf{13.3\%} & \textbf{14.3\%} & \textbf{20.6\%} & \textbf{4.1\%} & \textbf{13.3\%} & \textbf{14.4\%} \\
\cmidrule{2-9}
 & \multirow{4}{*}{GeoMiner} & Initial & 38.1\% & 47.3\% & 65.3\% & 21.8\% & 39.3\% & 43.9\% \\
 &  & M-Attack & 18.3\% & 19.3\% & 24.3\% & 11.4\% & 17.9\% & 19.0\% \\
 &  & GeoShield & 12.2\% & 13.0\% & 18.6\% & 8.1\% & 12.2\% & 13.0\% \\
 &  & \geocloak & \textbf{7.8\%} & \textbf{8.9\%} & \textbf{13.5\%} & \textbf{4.0\%} & \textbf{8.3\%} & \textbf{8.9\%} \\
\midrule
\multirow{8}{*}{Qwen3.5-Plus} & \multirow{4}{*}{Zero-Shot} & Initial & 31.1\% & 41.2\% & 71.7\% & 10.5\% & 35.7\% & 40.7\% \\
 &  & M-Attack & 15.3\% & 15.8\% & 21.9\% & 5.6\% & 15.3\% & 15.8\% \\
 &  & GeoShield & 11.8\% & 12.4\% & 17.1\% & 4.1\% & 11.2\% & 11.3\% \\
 &  & \geocloak & \textbf{6.5\%} & \textbf{6.9\%} & \textbf{10.9\%} & \textbf{1.6\%} & \textbf{5.8\%} & \textbf{6.0\%} \\
\cmidrule{2-9}
 & \multirow{4}{*}{GeoMiner} & Initial & 29.7\% & 39.7\% & 71.2\% & 14.2\% & 33.3\% & 39.8\% \\
 &  & M-Attack & 13.7\% & 14.5\% & 20.3\% & 4.9\% & 12.2\% & 13.2\% \\
 &  & GeoShield & 11.5\% & 12.4\% & 17.9\% & 4.7\% & 10.3\% & 11.4\% \\
 &  & \geocloak & \textbf{4.4\%} & \textbf{5.2\%} & \textbf{8.2\%} & \textbf{1.5\%} & \textbf{3.7\%} & \textbf{4.5\%} \\
\midrule
\multirow{4}{*}{GeoVista} & \multirow{4}{*}{GeoVista} & Initial & 35.2\% & 45.3\% & 71.9\% & 11.5\% & 39.3\% & 44.6\% \\
 &  & M-Attack & 21.4\% & 21.8\% & 25.3\% & 7.3\% & 21.1\% & 21.3\% \\
 &  & GeoShield & 15.6\% & 16.1\% & 19.6\% & 6.1\% & 15.3\% & 15.6\% \\
 &  & \geocloak & \textbf{11.1\%} & \textbf{12.1\%} & \textbf{14.9\%} & \textbf{3.3\%} & \textbf{10.9\%} & \textbf{11.4\%} \\
\bottomrule
\end{tabular}%
}
\end{table*}

\section{Visual Quality Comparison}
\label{appendix:visual_comparison}

\begin{figure}[ht]
\centering
\includegraphics[width=\linewidth]{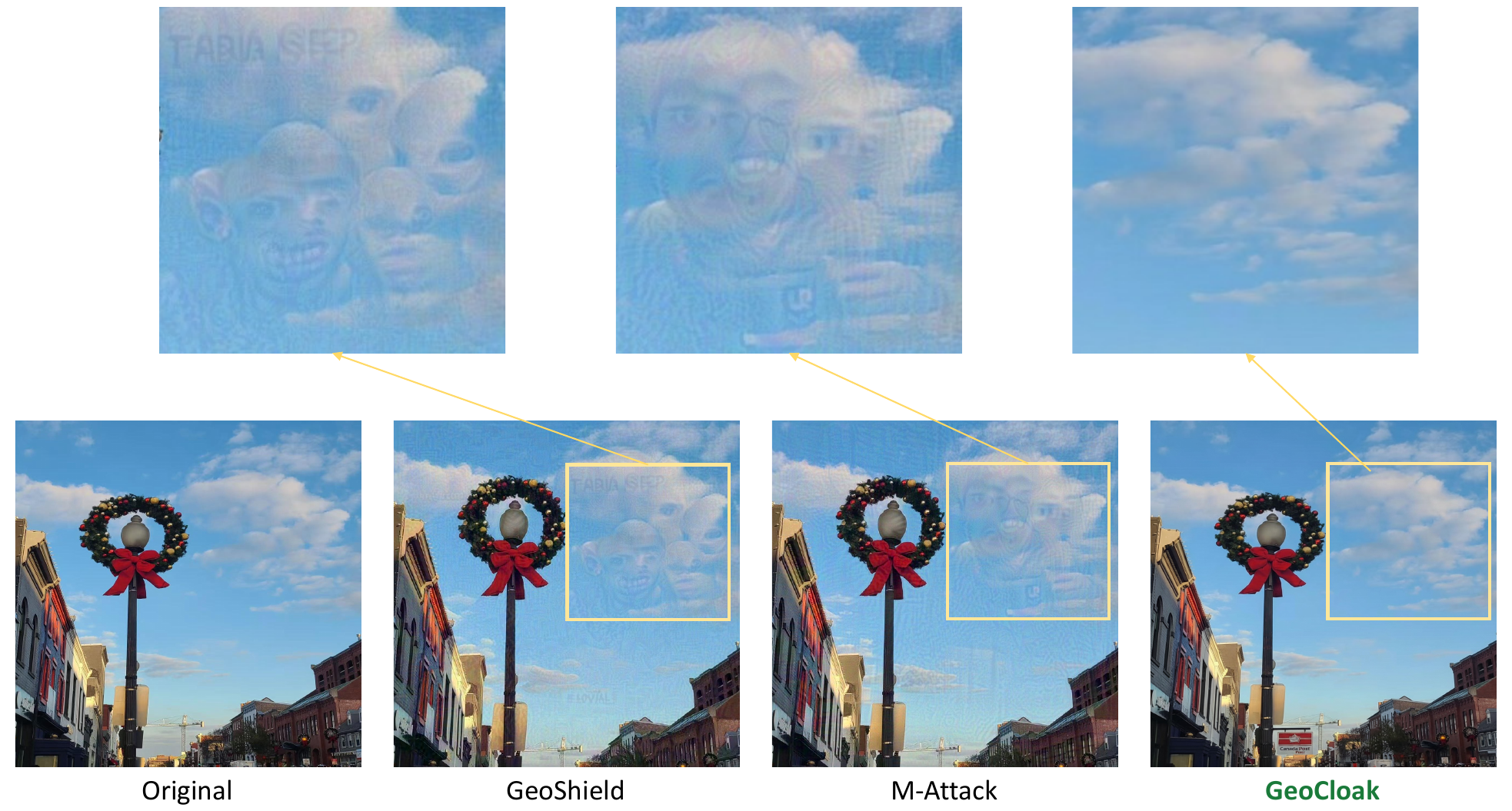}
\caption{Visual quality comparison with zoomed-in regions on a Im2GPS3k sample.
It serves as a supplementary illustration to \autoref{figure:qualitative_result}, providing additional visual comparison details
Zoomed-in crops (top) reveal that \geocloak introduces negligible utility degradation compared to GeoShield and M-Attack.
}
\label{figure:addition_zoom_in_1}
\end{figure}

\begin{figure}[!t]
\centering
\includegraphics[width=\linewidth]{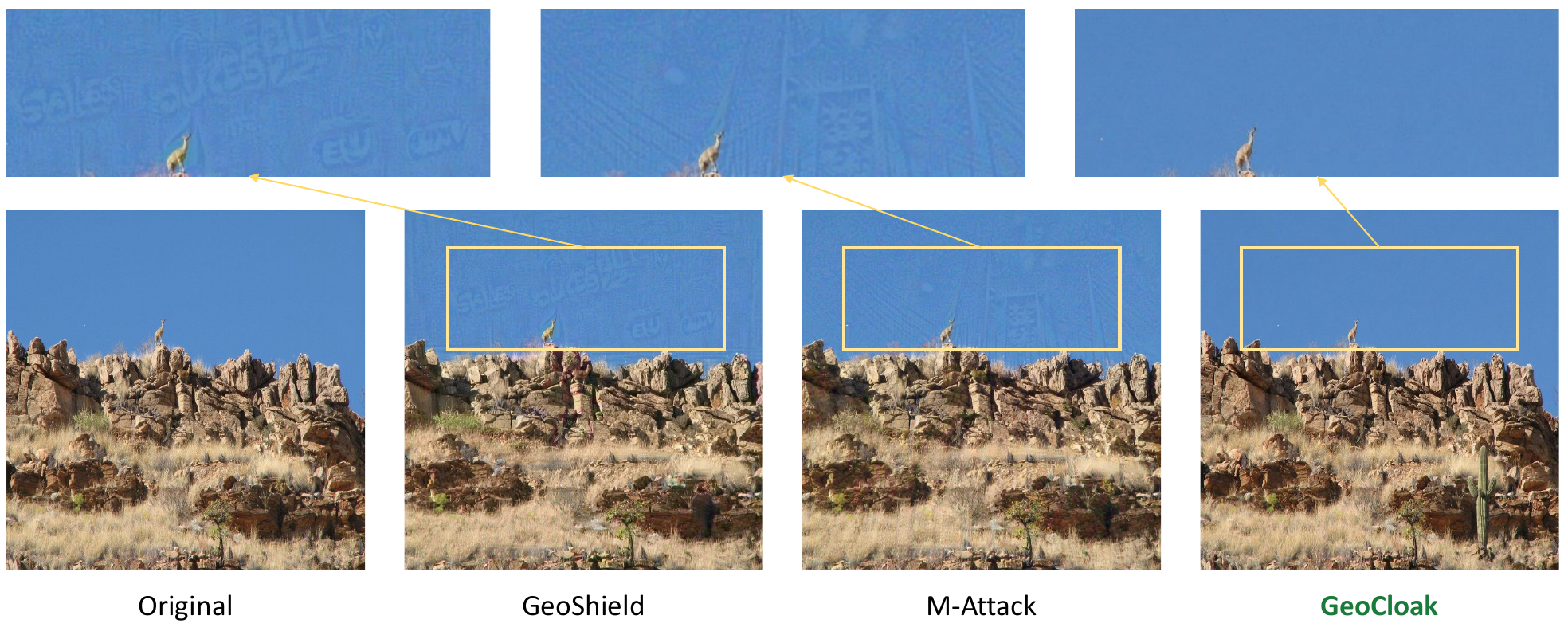}
\caption{Visual quality comparison with zoomed-in regions on a Dataset-PCW sample.
It serves as a supplementary illustration to \autoref{figure:qualitative_result}, providing additional visual comparison details
Zoomed-in crops (top) reveal that \geocloak introduces negligible utility degradation compared to GeoShield and M-Attack.
}
\label{figure:addition_zoom_in_2}
\end{figure}

\autoref{figure:addition_zoom_in_1} and \autoref{figure:addition_zoom_in_2} supplement \autoref{figure:qualitative_result} with zoom-in comparisons across three defense methods on images from Im2GPS3k and Dataset-PCW.
As can be clearly seen, \geocloak introduces negligible utility degradation compared to GeoShield and M-Attack, both of which produce prominent visual artifacts such as floating text and fragmented human heads.

\begin{figure}[ht]
\centering
\includegraphics[width=\linewidth]{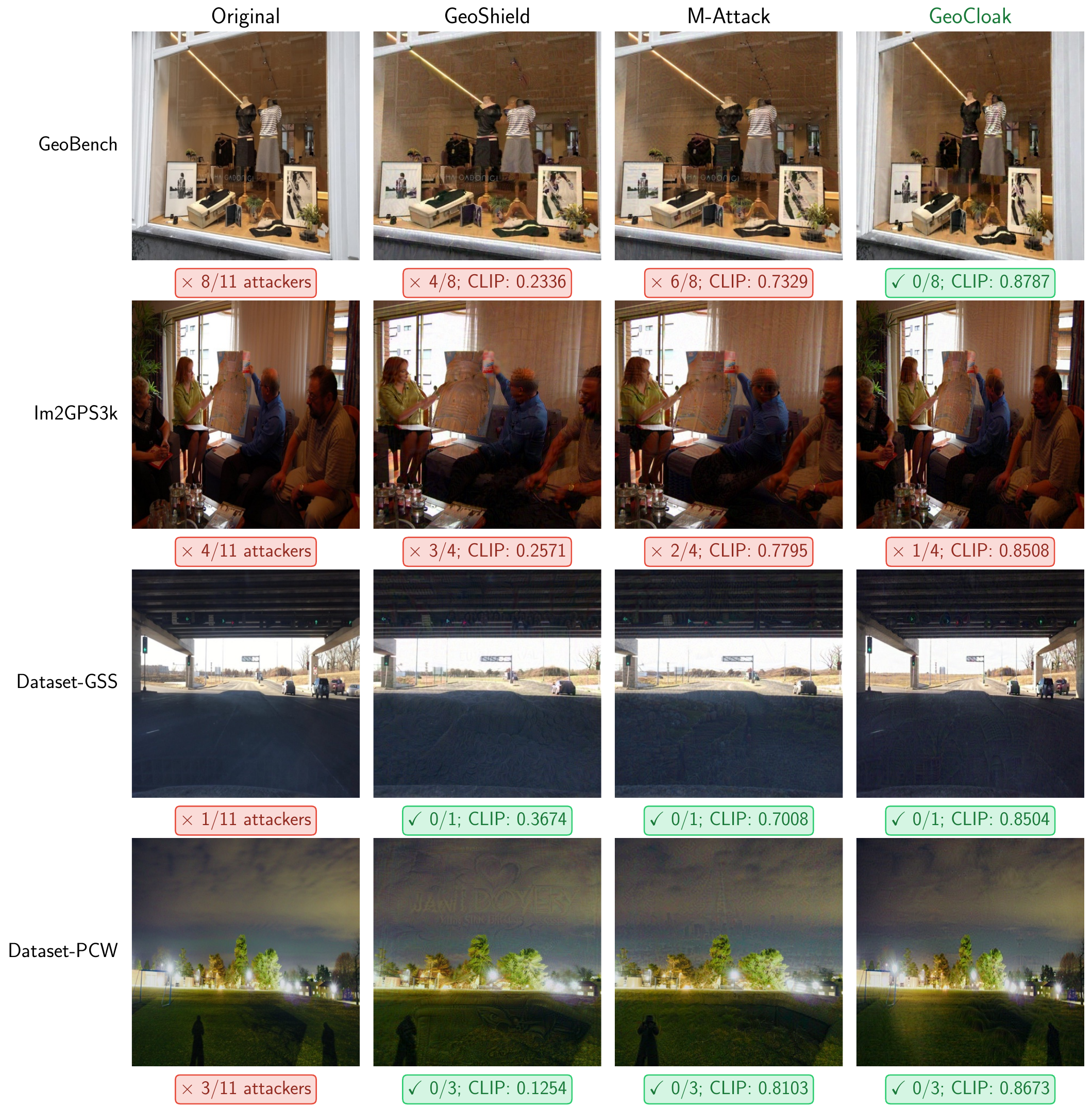}
\caption{Adversarial perturbation quality comparison.
Each row shows a case where adversarial perturbation is the key skill in \geocloak's successful protection turn.
Numbers indicate attackers that successfully geolocate the image and CLIP similarity.
\geocloak introduces substantially fewer visual artifacts than GeoShield and M-Attack while achieving equal or superior protection.
}
\label{figure:qualitative_adv_result}
\end{figure}

\autoref{figure:qualitative_adv_result} supplements \autoref{appendix:visual_comparison} with additional cases where adversarial perturbation serves as the key skill in \geocloak's successful protection.
Each row presents a case in which all three defense methods successfully apply adversarial perturbation from different datasets.
As best viewed with zoom, GeoShield and M-Attack both exhibit perceptible visual artifacts, whereas \geocloak maintains high visual fidelity.

\section{Human Perception Analysis}
\label{appendix:human_perception}

We randomly sample 100 source images, equally distributed in four datasets.
Each source image is paired with its \geocloak, GeoShield, and M-Attack output, producing 300 rating tasks.
Two independent annotators evaluate the same trials.
In each trial, the original image is presented alongside one protected image.
The defense identities are hidden.
Annotators score each protected image on a five-point Likert scale for naturalness, semantic preservation relative to the original image, and overall quality, with overall quality designated as the primary metric.
The evaluation interface provides the following guidelines as follows:
\begin{itemize}
    \item \emph{naturalness} assesses whether the protected image appears natural and realistic without obvious artifacts.
    \item \emph{semantic preservation} assesses whether the main objects, scene, and meaning of the original image are retained.
    \item \emph{overall quality} assesses the protected image holistically by considering both naturalness and content preservation.
\end{itemize}
All three metrics use the same five-point scale, ranging from 1 (severe and unacceptable problems) to 5 (a natural, complete image with almost no visible problems).

For analysis, we first average the two annotators' scores for each source image--defense pair, preserving the source image as the sampling unit rather than treating repeated ratings as independent observations.
We then estimate 95\% confidence intervals by bootstrapping the 100 source images with 10,000 resamples.
The complete results are reported in~\autoref{table:human_perception_full}.

\begin{table}[ht]
\centering
\caption{Complete human perception results on a five-point Likert scale.
Each cell reports the mean and image-level bootstrap 95\% confidence interval after averaging the two annotators' ratings for each source image and defense.
Higher is better ($\uparrow$).}
\label{table:human_perception_full}
\scalebox{0.6}{
\begin{tabular}{lccc}
\toprule
Defense & Naturalness$\uparrow$ & Semantic Preservation$\uparrow$ & Overall Quality$\uparrow$ \\
\midrule
GeoShield & 2.23 [2.06, 2.41] & 3.68 [3.59, 3.76] & 2.53 [2.38, 2.68] \\
M-Attack & 2.31 [2.15, 2.47] & 3.77 [3.67, 3.86] & 2.63 [2.49, 2.78] \\
\geocloak & \textbf{4.57 [4.44, 4.69]} & \textbf{4.45 [4.34, 4.56]} & \textbf{4.43 [4.30, 4.55]} \\
\bottomrule
\end{tabular}}
\end{table}

\section{Invalid Responses in Evaluation}
\label{appendix:invalid_response}

\begin{table}[ht]
\centering
\caption{API error and safety refusal rates (\%) per model, inference method, and defense method.
\textit{API Error}: request timeout or image-size limit (infrastructure failure).
\textit{Safety Refusal}: model explicitly declined to geolocalize (\textit{e.g.}, ``I cannot help with that'').
}
\label{table:invalid_response}
\scalebox{0.55}{
\begin{tabular}{lll | rr | rr | rr | rr}
\toprule
\multicolumn{3}{l}{} & \multicolumn{2}{c}{Initial} & \multicolumn{2}{c}{M-Attack} & \multicolumn{2}{c}{GeoShield} & \multicolumn{2}{c}{\geocloak} \\
\cmidrule(lr){4-5}\cmidrule(lr){6-7}\cmidrule(lr){8-9}\cmidrule(lr){10-11}
Model & Method & & API & Ref. & API & Ref. & API & Ref. & API & Ref. \\
\midrule
\multirow{2}{*}{Gemini-2.5-Pro} & Zero-Shot & & 0.0\% & 0.0\% & 0.0\% & 0.0\% & 0.0\% & 0.0\% & 0.0\% & 0.0\% \\
 & GeoMiner & & 0.0\% & 0.2\% & 0.0\% & 0.0\% & 0.0\% & 0.0\% & 0.0\% & 0.0\% \\
\midrule
\multirow{2}{*}{Gemini-3.1-Pro} & Zero-Shot & & 0.0\% & 0.0\% & 0.0\% & 0.0\% & 0.0\% & 0.0\% & 0.0\% & 0.0\% \\
 & GeoMiner & & 0.2\% & 0.0\% & 0.0\% & 0.0\% & 0.0\% & 0.0\% & 0.0\% & 0.0\% \\
\midrule
\multirow{2}{*}{GPT-5.4} & Zero-Shot & & 0.0\% & 0.0\% & 0.0\% & 0.0\% & 0.0\% & 0.0\% & 0.0\% & 0.0\% \\
 & GeoMiner & & 0.0\% & 0.0\% & 0.0\% & 0.0\% & 0.0\% & 0.0\% & 0.0\% & 0.0\% \\
\midrule
\multirow{2}{*}{o3} & Zero-Shot & & 0.0\% & 0.0\% & 0.0\% & 0.0\% & 0.0\% & 0.0\% & 0.0\% & 0.0\% \\
 & GeoMiner & & 0.2\% & 28.1\% & 0.0\% & 15.2\% & 0.0\% & 21.6\% & 0.0\% & 18.1\% \\
\midrule
\multirow{2}{*}{Qwen3.5-Plus} & Zero-Shot & & 6.2\% & 0.0\% & 0.7\% & 0.0\% & 0.8\% & 0.0\% & 2.0\% & 0.0\% \\
 & GeoMiner & & 9.6\% & 0.0\% & 0.8\% & 0.0\% & 0.8\% & 0.0\% & 2.1\% & 0.0\% \\
\midrule
\multirow{1}{*}{GeoVista} & GeoVista & & 0.0\% & 0.0\% & 0.0\% & 0.0\% & 0.0\% & 0.0\% & 0.0\% & 0.0\% \\
\bottomrule
\end{tabular}%
}
\end{table}

As shown in~\autoref{table:invalid_response}, a fraction of model evaluations produce invalid responses.
First, API-level failures (e.g., request timeouts and image-size limits) affect Qwen3.5-Plus evaluations, reflecting infrastructure constraints of the model provider.
Note that we leverage the official provider platform.
Second, o3 exhibits a high safety-refusal rate (28.1\% with no defense) under the GeoMiner inference method, where the model may explicitly identify the step-by-step geolocation prompt as a potential privacy risk and decline to respond.
We acknowledge that these sources of error are difficult to mitigate on our side.
Samples that encounter such errors during the initial evaluation are therefore excluded from post-defense evaluation and recorded as localization failures across all metrics.
Importantly, invalid responses are consistently treated as failures across all defense methods, ensuring no systematic bias toward any particular one.
They are also negligible for other models (0.0\% invalid response rate), so our conclusions remain unaffected.

\section{Interactive \geocloak}
\label{appendix:interactive_geocloak}

\begin{figure*}[!t]
\centering
\includegraphics[width=\linewidth]{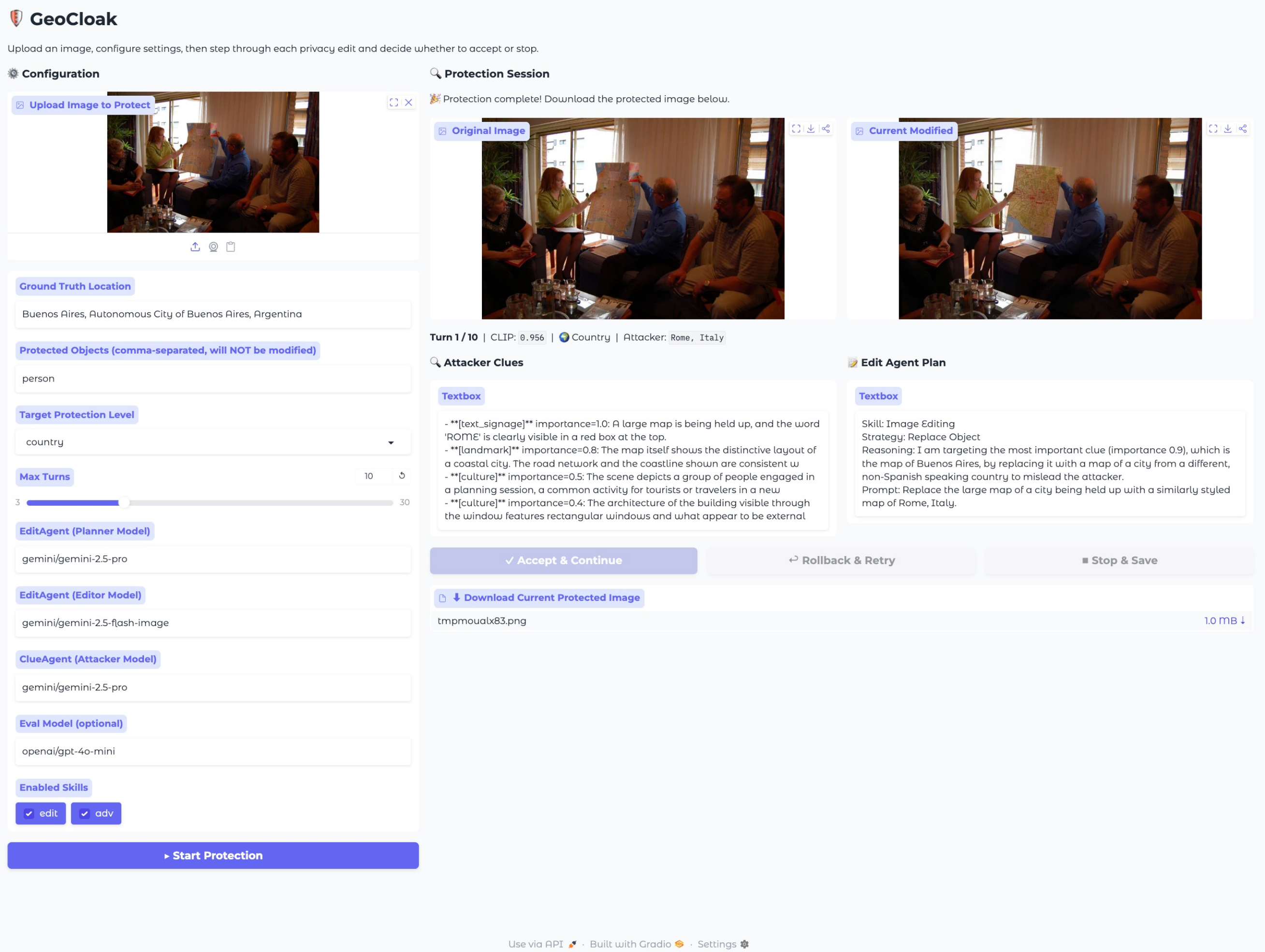}
\caption{
Interactive version of \geocloak.
}
\label{figure:interactive_example}
\end{figure*}

\autoref{figure:interactive_example} exhibit the interactive version of \geocloak.
Interactive version of \geocloak.
Users upload the image they wish to protect, specify objects that must remain unmodified, and set a target protection level (city, state, or country); it then iteratively proposes and applies privacy-preserving edits, pausing after each turn so the user can accept, roll back, or stop.
This design replaces automated utility checks with direct human judgment, giving image owners transparent, turn-by-turn control over what their shared images ultimately reveal.

\begin{lstlisting}[caption={\textbf{ClueAgent Prompt}}, captionpos=t, label=lst:clueagent]
ACT AS A GEOLOCALIZATION EXPERT AGENT.
Your task is to identify the location of this image.

Analyze this image step-by-step:
1. Look for landmarks, text/signage, cultural elements, and spatial patterns.
2. Determine the most likely location.
3. Explain your reasoning.
4. List the specific VISUAL CLUES that helped you identify it.

Return your analysis in JSON format:
{
    "predicted_location": "City, State (if applicable), Country",
    "reasoning": "Your step-by-step reasoning process",
    "clues": [
        {
            "type": "landmark|text_signage|culture|spatial",
            "description": "what the clue is and why it suggests this location",
            "region": [x1, y1, x2, y2],
            "importance": 0.0-1.0
        }
    ]
}

Clue types:
- landmark: Famous buildings, monuments, natural landmarks, recognizable structures
- text_signage: Street signs, shop names, billboards, language on signs
- culture: Clothing styles, vehicle types (left/right-hand drive), local customs, architectural style
- spatial: Road layout, urban planning patterns, intersection design, city structure

Importance scoring:
- 0.9-1.0: Uniquely identifies location (e.g., Eiffel Tower, specific street sign with address)
- 0.7-0.8: Narrows to city/region (e.g., language on signs, distinctive architecture)
- 0.4-0.6: Narrows to country (e.g., driving side, general cultural style)
- 0.1-0.3: Weak/ambiguous clue (e.g., common objects, generic features)

Note: "region" should be bounding box in pixel coordinates [x1, y1, x2, y2], or null if not precisely localizable.

IMPORTANT: Some clues in the image may be contradictory or misleading (e.g., a foreign-brand shop sign in a local street, a tourist souvenir from another country, or an imported vehicle). When clues conflict, critically evaluate each one, weigh higher-confidence, location-specific clues (e.g., unique architecture, local-language street signs, distinctive road markings) over generic or easily transplanted clues (e.g., global brand logos, mass-produced vehicles). Explicitly note any contradictory clues in your reasoning and explain why you prioritize certain clues over others.
\end{lstlisting}

\begin{lstlisting}[caption={\textbf{EditAgent Prompt}}, captionpos=t, label=lst:editagent]
You are an Image Privacy Protection Agent. Your task is to write Python code that modifies images to obfuscate location-revealing clues.

---

## Your Workflow

Each turn the orchestrator sends you a structured observation with four XML sections:

```
<code_output>     stdout from your last code execution                      </code_output>
<utility>        whether your modification was accepted or rolled back      </utility>
<attacker_result> attacker's current prediction, protection level, clues  </attacker_result>
```

**How to read each section:**
- `<utility> status: FAILED` - your image was rolled back, `image` is restored to the previous state; try a different approach
- `<utility> status: PASSED` - your modification was committed; `image` is now that modified version
- `<attacker_result>` - shows `predicted_location`, current `protection_achieved` vs target, and ranked clues to tackle

Based on the observation, write ONE Python code block that further modifies `image` to remove/obfuscate the remaining clues.

---

## Available Variables & Functions

- `image`: Current image (float32 numpy [0,1], persists across turns, always start from this)
- `original_image`: Original unmodified image (read-only, for reference only)
- `image_height`, `image_width`: Image dimensions (int)
- `clues`: List of attacker clues from the most recent analysis (auto-updated each turn)
  - Each clue: {'type': str, 'description': str, 'region': [x1,y1,x2,y2], 'importance': float}
- `edit_image(image, prompt)`: Semantic image editing, returns float32 numpy [0,1]
- `ensemble`: Pre-loaded CLIP ensemble (EnsembleExtractor, 4 backbones, do NOT recreate)
- `get_crops(tensor)`: Returns (local_crop, global_crop) for CLIP feature extraction
- `bbox_to_crop(tensor, bbox)`: Crops tensor to [x1,y1,x2,y2] bounding box

Standard libraries available: `torch`, `numpy (np)`, `cv2`, `PIL.Image`, `io`, `base64`

---

## Rules

- ALWAYS store your result: `image = modified_image`
- Do NOT import modules; all functions are pre-injected into scope
- Write exactly ONE ```python``` code block per turn
- After writing your code block, STOP and wait for feedback

---

## Skills

{{skill_num}} skills are available. Choose based on the nature of the remaining clues:

| Skill | When to Use |
|-------|-------------|
| **Image Editing** | Semantically modify discrete clues: change text/signs, swap objects, erase landmarks, shift style, add decoys. Best when clues are specific and editable. Consider **Add Decoy** when clues are large/structural and adversarial perturbation has repeatedly failed; plant a misleading foreign element instead of altering existing content. |
| **Adversarial Perturbation** | Gradient-based noise targeting CLIP features. Best for large landmarks or scene-level features that cannot be semantically edited. |

**Retry rule:** After a utility failure or attack failure, adjust based on the active skill's retry logic and try again. Switch to a different skill only when you have exhausted the retry options for the current skill.

Skill specification(s) are provided below.

---


{{SKILLS}}



---
\end{lstlisting}

\begin{lstlisting}[caption={\textbf{Editing Skill Prompt}}, captionpos=t, label=lst:editing]
## Image Editing Skill

Read the strategies and guidelines below, then in your response output **two things in order**:

1. A JSON config block declaring your plan:
```json
{{"skill": "Image Editing", "strategy": "Replace Text", "prompt": "the exact prompt you will pass to edit_image", "reasoning": "one sentence: which clue you are targeting and why this strategy fits"}}
```
- `"skill"`: always `"Image Editing"`
- `"strategy"`: one of `"Replace Text"`, `"Replace Object"`, `"Erase Object"`, `"Add Decoy"`, `"Style Transfer"`
- `"prompt"`: the exact string you will pass to `edit_image`
- `"reasoning"`: which clue (description + importance) you are targeting and why

2. A `python` code block calling `edit_image` with that prompt.

Both must appear in the same response turn.

### General Principles

- **Blend naturally**: all modifications must be consistent with the scene's lighting, perspective, scale, and style; nothing should look artificially inserted or out of place
- **Stay culturally close**: replacements and additions must come from a visually similar culture/region; do not introduce elements from a visually distant culture (e.g. do not add East Asian signage to a European street scene)
- **Start small**: prioritize clues that are both high-importance and small in region first; they give the most privacy gain with the least visual disruption
- **Preserve the main subject**: never replace, recolor, or occlude the image's primary subject as a whole (e.g. do not swap out an entire building, vehicle, or person, and do not change the dominant color scheme of a main object). Fine-grained details *on* the main subject, such as text, logos, or small markings, are fine to edit
- **Persist and escalate**: After each edit, check both `check_utility` and `attacker.attack` results. If the attacker still identifies the location: (1) **retry within the same strategy**, adjust the prompt, target a different element, or pick a different replacement target (up to 2 retries); (2) if still failing, **switch to a different strategy** targeting the same or next clue; (3) only switch skill after exhausting multiple strategies

### Editing Strategies

| Strategy | Use when |
|---|---|
| **Replace Text** | signs, storefronts, street names, license plates, or any visible text reveals the location |
| **Replace Object** | location-specific objects are present (tram models, post boxes, traffic signs, vehicle types) |
| **Erase Object** | an object, sign, text, or logo is too specific to replace but small enough to remove cleanly |
| **Add Decoy** | dominant clues are hard to replace, erase, style-transfer, or perturb directly (e.g. large central landmarks, iconic facades, wide street scenes), add a misleading element nearby to recontextualize them instead |
| **Style Transfer** | **apply only to peripheral, small, or background elements** (e.g. a side wall, distant building, or background vegetation); never restyle the main subject or dominant foreground elements, as broad style changes are visually conspicuous |

---

**Replace Text**
- Use when: signs, storefronts, street names, license plates, or any visible text reveals the location
- Strategy: change to a different language or writing system that suggests another country
- Example prompt: `"Change the German text 'B\"ackerei' on the shop sign to Czech text 'Pek\'arna'"`

**Replace Object**
- Use when: location-specific objects are present (tram models, post boxes, traffic signs, vehicle types)
- Strategy: swap with a visually similar object from a geographically nearby but distinct country
- Example prompt: `"Replace the Austrian tram with a German-style tram, keeping the same color and size"`

**Erase Object**
- Use when: a sign, logo, or label is too specific to replace but small enough to remove
- Strategy: remove entirely and let the background fill in naturally
- Example prompt: `"Remove the street sign on the left wall completely, inpaint with the surrounding wall texture"`

**Add Decoy**
- Use when: the image has empty space (walls, ground, sky) where a misleading element can be added
- Strategy: add an element that suggests a different country; place it so that it **recontextualizes nearby existing clues**; the goal is to make the attacker re-interpret other scene elements as also belonging to the target country, not just notice an isolated foreign object
- **Physical plausibility is required**: the added element must be attached to or resting on a real surface, e.g. a sign mounted on a wall, a sticker on a pole, text painted on the ground. Never place elements floating in mid-air or in physically impossible positions
- Example prompt: `"Add a small Japanese street sign with hiragana text mounted on the wall in the lower-right corner, matching the wall texture and lighting"`

**Style Transfer**
- Use when: architecture, vegetation, or overall scene style is the primary clue
- Strategy: shift visual style toward a geographically similar but distinct region
- Example prompt: `"Shift the style of the background buildings on the far left from Thai temple to Indonesian architecture, keeping the main foreground structure unchanged"`

---

### Good Location Swap Pairs
(visually similar but geographically distinct)

- Austria - Germany - Switzerland
- Taiwan - Hong Kong - Japan
- France - Belgium - Italy
- Czech - Poland - Slovakia
- UK - Ireland - Australia
- Thailand - Indonesia - Malaysia

---

### API

- `edit_image(image, prompt)`, returns numpy float32 array in [0, 1], original resolution, no mask required
- `numpy`, `plt` are available

**DO NOT import any modules or skills**: `edit_image` and all other functions are already pre-injected into the execution scope. Never write `import` or `from ... import` statements. There is no `editor` object, no `ImageEditing` class. Use `edit_image(image, prompt)` directly.
\end{lstlisting}

\begin{lstlisting}[caption={\textbf{Adversarial Perturbation Skill Prompt}}, captionpos=t, label=lst:adv]
## Adversarial Perturbation Skill

Read the options below, then in your response output **two things in order**:

1. A JSON config block declaring your choices:
```json
{{"skill": "Adversarial Perturbation", "target": "region_focus", "loss": "dual_loss", "non_geo_description": "...", "text_weight": 1.0, "steps": 100, "reasoning": "one sentence: which clue(s) you are targeting and why this configuration fits"}}
```
- `"skill"`: always `"Adversarial Perturbation"`
- `"target"`: `"region_focus"` (when clues have bbox) or `"full_image"`
- `"loss"`: `"dual_loss"` (when scene has clear visual content) or `"feature_only"`
- `"non_geo_description"`: required when `loss` is `"dual_loss"`, omit key otherwise
- `"text_weight"`: *(dual_loss only)* how strongly to pull toward `non_geo_description`; start with `5.0`
- `"steps"`: start with `50`
- `"reasoning"`: which clue(s) you are targeting and why this target/loss combination fits

2. A `python` code block implementing the attack.
```python
code snippet
```

Both must appear in the same response turn.

---

Use `torch` and the pre-loaded `ensemble` (an `EnsembleExtractor` instance already available in scope) to implement gradient-based noise.

---

### Fixed Setup

```python
import torch
import random

# Pre-injected. Do NOT recreate: ensemble, get_crops, bbox_to_crop
device = "cuda" if torch.cuda.is_available() else "cpu"
epsilon = 8 / 255   # privacy budget

image_tensor = torch.from_numpy(image).permute(2, 0, 1).float().to(device)
delta = torch.zeros_like(image_tensor, requires_grad=True)
```

---

### Autonomous Decisions

**A. Steps**: how many optimization steps?
- Start with 50; see Retry Logic for when to increase.

**B. Reference region**
- `region_focus` *(when clues have bbox)*: randomly sample one region per step from the attacker clues as the push-away reference. The full image is also included in the pool so global features are disrupted too.
- `full_image`: Use the full image as reference.

**C. Loss composition**
- `feature_only`: maximize cosine distance between adv crops and reference.
- `dual_loss` *(when scene has clear visual content)*: feature distance + pull toward a `non_geo_description` (see D).

**D. Generating `non_geo_description`** *(dual_loss only)*

Max 15 words. Derived from clue `description` fields: keep visual content (objects, colors, scene type), strip all geo-identifying language (place names, landmark names, culturally specific references, or any combination of elements that uniquely identifies a place).

Example: clue: *"Vanda orchids, Singapore's national flower, in a botanical garden"*
-> `"Colorful tropical orchid flowers with purple and orange blooms in a garden"`

---

### Attack Loop

```python
non_geo_description = "..."  # [dual_loss only]
text_weight = 5.0            # [dual_loss only] set from your JSON config

# clues: list of dicts from attacker.attack() result, each with 'region' [x1,y1,x2,y2]
# [region_focus] build bbox_list once, outside the loop
bbox_list = [[c['region'][0], c['region'][1], c['region'][2], c['region'][3]]
             for c in clues if c.get('region')]
bbox_list.append([0, 0, image_tensor.shape[2], image_tensor.shape[1]])  # always include full image

# [dual_loss only] precompute text features once; they don't change across steps
# REMOVE these two lines entirely if using feature_only
with torch.no_grad():
    txt_feats = ensemble.get_text_features(non_geo_description)

for step in range(steps):
    adv = image_tensor + delta
    local_crop, global_crop = get_crops(adv)

    # [region_focus] single random bbox per step
    ref_crop = bbox_to_crop(image_tensor.detach(), random.choice(bbox_list))
    # [full_image] ref_crop = image_tensor.detach()

    # Extract features for both crops; gradients flow through to delta
    local_feats  = ensemble.get_image_features(local_crop)   # Dict[int, (1,D)]
    global_feats = ensemble.get_image_features(global_crop)  # Dict[int, (1,D)]

    loss = torch.tensor(0.0, device=device)

    # [region_focus / full_image] single ref
    with torch.no_grad():
        ref_feats = ensemble.get_image_features(ref_crop)
    for i in range(len(ensemble)):
        for adv_feat in [local_feats[i], global_feats[i]]:
            loss = loss - torch.sum(adv_feat * ref_feats[i])
            loss = loss + text_weight * torch.sum(adv_feat * txt_feats[i])  # [dual_loss only] REMOVE if feature_only

    grad = torch.autograd.grad(loss, delta, create_graph=False)[0]
    with torch.no_grad():
        delta.data = delta.data + (epsilon / 10) * torch.sign(grad)
        delta.data = torch.clamp(delta.data, -epsilon, epsilon)
        delta.data = torch.clamp(image_tensor + delta.data, 0, 1) - image_tensor

modified_image = torch.clamp(image_tensor + delta, 0, 1).detach().cpu().permute(1, 2, 0).numpy()
```

---

### Retry Logic

**`non_geo_description` as a tuning dial** *(dual_loss only)*:
- More **descriptive** (rich visual detail, closely matching original scene) -> stronger utility anchor, weaker privacy effect
- More **generic** (broad scene type only) -> weaker utility anchor, stronger privacy effect
- Adjust in either direction depending on which constraint is failing.

**If utility check fails**:
- Add `dual_loss` if using `feature_only` - text pull constrains perturbation direction, reducing visual degradation
- If already using `dual_loss`: increase `text_weight` (e.g. `5` -> `20`) to strengthen the semantic anchor; or make `non_geo_description` more descriptive (closer to the original visual content)
- If still failing: decrease epsilon (e.g. `6/255`)

**If utility OK but attack still succeeds**:
- Switch `full_image` -> `region_focus` (concentrate disruption on high-importance clue regions)
- Make `non_geo_description` more generic (broader scene type, fewer visual details); or decrease `text_weight` (e.g. `5.0` -> `2.0`) to let the pull-away loss dominate more freely
- If using `dual_loss`: try switching to `feature_only`, removes the text anchor, allowing the perturbation to move more freely away from location-specific features (monitor utility closely)
- Increase steps (e.g. 100 -> 200)

---

### Coding Rules

- NEVER use in-place ops in forward pass (`x += y`); use `x = x + y`
- NEVER reassign `delta`; always modify `delta.data`
- All `get_image_features` expect `(C, H, W)` float `[0,1]`; do NOT pre-normalize
- NEVER call `attacker.attack()` inside the optimization loop; `clues` is already available from a prior call before entering this skill; calling it per-step wastes ~20s x steps and provides no gradient benefit
- check_utility is handled by the orchestrator automatically
\end{lstlisting}

\begin{lstlisting}[caption={\textbf{Evaluating Scope: Landmark Intention Analysis}}, captionpos=t, label=lst:scope]
**Task:** Analyze the object highlighted in the **RED BOX**.

        **Physical Data:**
        *   **Relative Size:** {metrics['R_area']:.2\%} of total image.
        *   **Centrality:** {metrics['D_center']:.2f} (0.00 is dead center).
        *   **Saliency Gap:** {metrics['SDR']:.2f} (Box Brightness vs Avg).

        **Decision Logic:**
        1.  **IDENTIFY**: Is it a Widely-Known Landmark (Tier A), Local Spot (Tier B), or Generic (Tier C)?
            - Tier A: Recognized beyond a local community, includes globally famous landmarks, nationally iconic structures, UNESCO World Heritage Sites, and major tourist attractions whose location is already public knowledge
            - Tier B: Recognizable only within a local or regional community (neighborhood temple, city-specific monument, regional landmark unknown outside its area)
            - Tier C: Generic structure with no identifying features (random building, anonymous wall), not identifiable
        2.  **PHYSICAL CHECK**:
            * Object needs PROMINENCE (High Size/Saliency/Center) to be 'Intentional'.
        3.  **CONTEXT**: Subject vs Background?

        **Output Format:**
        Provide PURE JSON. For "reasoning": if Tier A or B, include the specific landmark name and its geolocation (city, country) if you can identify it.
        {{
            "object_name": "...",
            "tier": "A/B/C",
            "verdict": "Intentional" or "Incidental",
            "reasoning": "..."
        }}
\end{lstlisting}

\end{document}